\documentclass[conference]{IEEEtran}

\usepackage{times}
\usepackage[numbers]{natbib}
\usepackage{multicol}
\usepackage[bookmarks=true]{hyperref}
\usepackage{graphicx}
\usepackage{amsmath}
\usepackage{amssymb}
\usepackage{booktabs}
\usepackage[table]{xcolor}
\usepackage{colortbl}
\usepackage{pifont}
\newcommand{\cmark}{\ding{51}}%
\newcommand{\xmark}{\ding{55}}%
\usepackage{xspace}
\let\labelindent\relax
\usepackage{enumitem}
\usepackage{mdframed}
\usepackage{subcaption}
\usepackage{capt-of}
\usepackage{algorithm}
\usepackage{algpseudocode}  
\usepackage{listings}
\providecommand{\NewStructureName}[1]{}
\providecommand{\AssignStructureRole}[2]{}
\providecommand{\NewTaggingSocket}[2]{}
\providecommand{\NewTaggingSocketPlug}[3]{}
\providecommand{\AssignTaggingSocketPlug}[2]{}
\providecommand{\UseStructureName}[1]{}
\providecommand{\UseTaggingSocket}[1]{}
\providecommand{\tagstructbegin}[1]{}

\usepackage{tcolorbox}
\tcbuselibrary{breakable,skins,listings}
\usepackage{tabularx}
\usepackage{stfloats}

\usepackage{hyperref}

\definecolor{StanfordRed}{RGB}{140,21,21}
\definecolor{StanfordDarkRed}{RGB}{111,17,17}
\definecolor{StanfordLightGray}{RGB}{245,243,241}

\hypersetup{
    colorlinks=true,      
    linkcolor=StanfordRed,        
    citecolor=StanfordRed,    
    urlcolor=StanfordRed         
}

\algrenewcommand\algorithmicindent{1.1em}

\newtcolorbox{promptbox}[1][]{%
  enhanced,
  breakable,
  colback=StanfordLightGray,          
  colframe=StanfordRed,               
  boxrule=0.8pt,
  arc=6pt,
  left=8pt,right=8pt,top=8pt,bottom=8pt,
  fonttitle=\bfseries\color{white},
  colbacktitle=StanfordRed,           
  coltitle=white,
  #1
}

\newtcblisting{promptlisting}[1][]{%
  enhanced,
  breakable,
  colback=white,
  colframe=StanfordRed,
  colbacktitle=StanfordRed,
  fonttitle=\bfseries\color{white},
  coltitle=white,
  boxrule=0.8pt,
  arc=6pt,
  left=8pt,right=8pt,top=8pt,bottom=8pt,
  listing only,
  listing options={
    basicstyle=\ttfamily\footnotesize,
    breaklines=true,
    columns=fullflexible,
    keepspaces=true,
    mathescape
  },
  #1
}

\newcommand{\secref}[1]{Section~\ref{#1}}
\renewcommand{\eqref}[1]{Eq.~\ref{#1}}
\newcommand{\figref}[1]{Figure~\ref{#1}}

\newcommand{\tabref}[1]{Table~\ref{#1}}

\newcommand{\algoref}[1]{Algorithm~\ref{#1}}

\newcommand{\appendref}[1]{Appendix~\ref{#1}}

\newcolumntype{Y}{>{\centering\arraybackslash}X}

\newcommand{\eg}{\emph{e.g.}}

\def\S{\mathcal{S}}
\def\A{\mathcal{A}}
\def\H{\mathcal{H}}

\def\O{\mathcal{O}}  
\def\P{\mathcal{P}}  
\def\E{\mathcal{E}}  

\DeclareMathOperator*{\Exp}{\mathbb{E}}

\def\method{\texttt{PhysMem}\xspace}                    

\def\scientificloop{scientific memory loop\xspace}

\def\vlm{VLM\xspace}

\def\taskParts{Parts Organization\xspace}
\def\taskBall{Ball Navigation\xspace}
\def\taskStack{Balanced Stacking\xspace}
\def\taskBrick{Brick Insertion\xspace}

\def\prinAvoid{\textsc{Avoid}\xspace}
\def\prinPrefer{\textsc{Prefer}\xspace}
\def\prinSequence{\textsc{Sequence}\xspace}

\def\episodicMem{episodic memory\xspace}
\def\workingMem{working memory\xspace}
\def\longtermMem{long-term memory\xspace}

\def\fighero{fig:hero}
\def\figpipeline{fig:pipeline}
\def\figloop{fig:scientific-loop}
\def\figtasks{fig:tasks}
\def\figcurves{fig:learning-curves}

\def\figscaling{fig:principle-scaling}
\def\figdifficulty{fig:difficulty-scaling}

\def\tabood{tab:ood-transfer}

\def\tabablation{tab:ablation}

\newenvironment{compactitem}{
  \begin{itemize}[noitemsep,topsep=0pt,parsep=0pt,partopsep=0pt,leftmargin=*]
}{
  \end{itemize}
}

\newenvironment{compactenum}{
  \begin{enumerate}[noitemsep,topsep=0pt,parsep=0pt,partopsep=0pt,leftmargin=*]
}{
  \end{enumerate}
}

\newcommand{\paperquoteRight}[2]{%
  \begin{center}
    \vspace{0.3em}
    \begin{minipage}{0.92\linewidth}
      \raggedleft\itshape
      #1\par
      \vspace{-0.15em}
      \hfill\rule{0.86\linewidth}{0.4pt}\par
      \vspace{-0.2em}
      \raggedleft\normalfont #2
    \end{minipage}
  \end{center}
}

\begin{document}

\title{\textcolor{StanfordRed}{\textit{\textbf{Phys}}}\textcolor{black}{\textit{\textbf{Mem}}}: Scaling Test-Time \textbf{Mem}ory for Embodied \textcolor{StanfordRed}{\textbf{Phys}}ical Reasoning}
\author{
  \vspace{-0.5em}
  \IEEEauthorblockN{Haoyang Li\textsuperscript{$\spadesuit$} \quad Yang You\textsuperscript{$\clubsuit$} \quad Hao Su\textsuperscript{$\spadesuit$} \quad Leonidas Guibas\textsuperscript{$\clubsuit$}}\\
  \vspace{-0.3em}
  \IEEEauthorblockN{\textsuperscript{$\clubsuit$}Stanford University \quad \textsuperscript{$\spadesuit$}UC San Diego}
  \vspace{-2.9em}
}

\maketitle


\begin{abstract}
Reliable object manipulation requires understanding physical properties that vary across objects and environments. Vision-language model (VLM) planners can reason about friction and stability in general terms, but they often cannot predict how a specific ball will roll on a particular surface or which stone will provide a stable foundation without direct experience. We present \method, a memory framework that lets a VLM robot planner learn physical principles from interaction at test time, without updating model parameters. The system records experiences, drafts candidate hypotheses, and verifies them through targeted interaction before committing validated knowledge to guide future decisions. Verification before application is the central design choice: the system tests each hypothesis against new observations instead of applying retrieved experience directly, so that prior experience does not harden into a stale rule once physical conditions change. We evaluate \textit{\method} on three real-world manipulation tasks and simulation benchmarks across four VLM backbones. On a controlled brick insertion task, principled abstraction achieves 76\% success against 23\% for direct experience retrieval, and real-world experiments show consistent improvement over 30-minute deployment sessions. Website: \href{https://phys-mem.github.io/}{https://phys-mem.github.io/}

\end{abstract}

\IEEEpeerreviewmaketitle


\section{Introduction}
\label{sec:introduction}
\paperquoteRight{For the things we have to learn \\ before we can do them, we learn by doing them.}{Aristotle, \textit{Nicomachean Ethics}}

Vision-language models (VLMs) can describe physical concepts such as friction, balance, and momentum~\citep{driess2023palme,brohan2023rt2,gemini2025robotics}. When deployed as robot planners, however, they often struggle to predict how these principles apply to a specific situation. A VLM planner may understand friction in the abstract yet misjudge how far a ball will roll on a given surface, or recognize stability as a concept yet fail to identify which irregular stone will provide a stable foundation.
This gap between declarative knowledge and physical grounding is well documented~\citep{zhang2026vlm4vla,kawaharazuka2025vla}, and its consequences compound in planning, where a single misjudgment about contact or dynamics can invalidate an entire action sequence.
This paper investigates whether a VLM robot planner can acquire useful physical understanding during deployment, through its own interaction, without updating model parameters.
To isolate reasoning from execution, we separate high-level planning (VLM decisions) from low-level control (motion execution), ensuring that observed improvements reflect better physical understanding rather than better motor policies.

We focus on settings where the correct strategy is hard to infer from vision alone: tasks where the relevant physical parameters are not visually apparent and cannot be inferred from pre-training data.
Specifically, we consider the following 3 tasks: 1) \textit{\taskParts} requires learning spatial relationships between irregular shapes and available gaps; 2)
\textit{\taskBall} requires discovering contact dynamics through trial, since a soccer ball's behavior under obstacles cannot be predicted from appearance; 3)
\textit{\taskStack} requires stability judgments about stones whose mass distribution and surface friction only reveal themselves upon contact.
Initial attempts on these tasks often fail because the planner lacks the specific physical intuition that only interaction can provide.

\begin{figure}[t]
    \centering
    \includegraphics[width=\columnwidth]{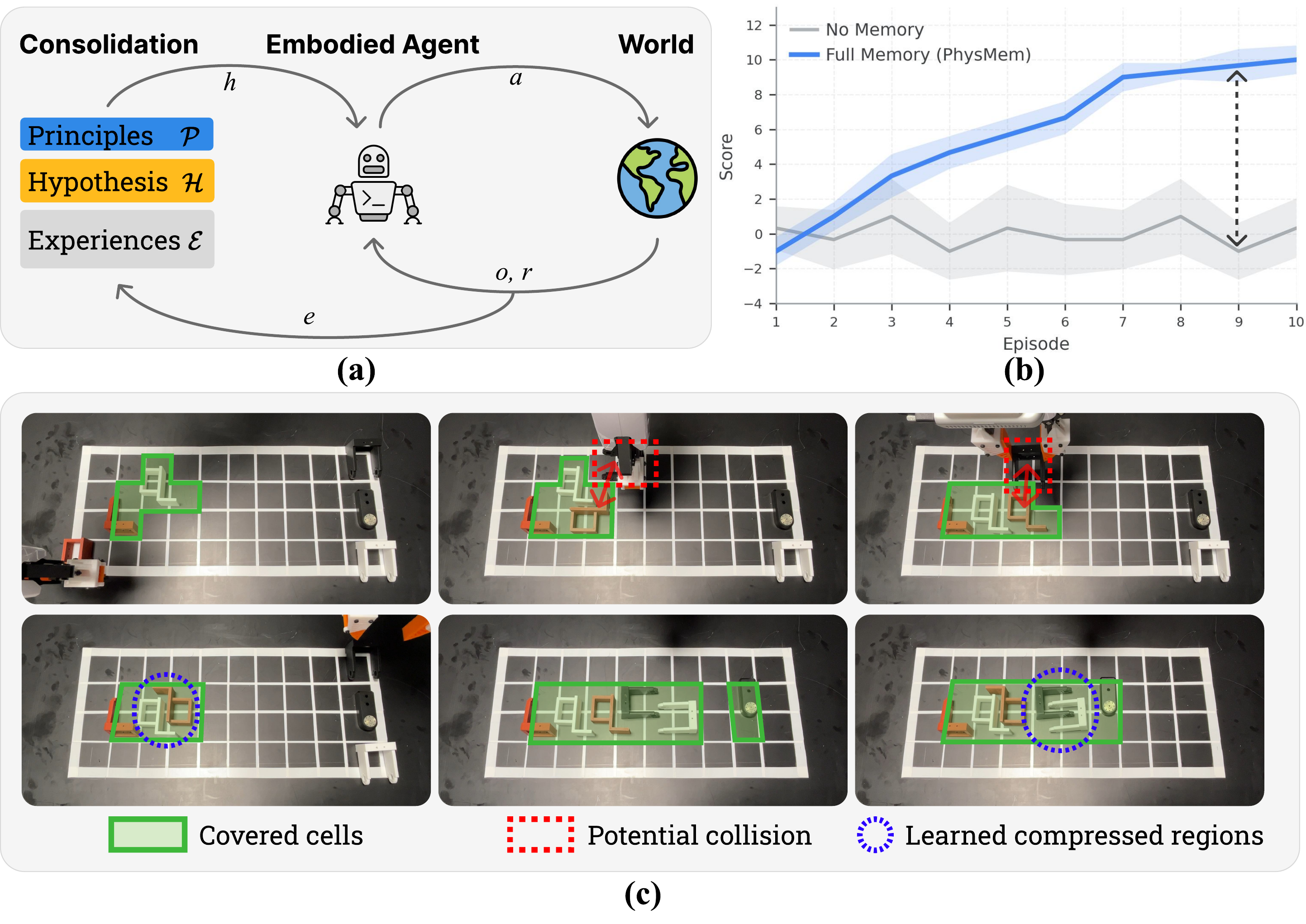}
    \caption{%
        \textbf{\method learns physical principles through interaction.}
        \textbf{(a)}~Continually Learn via the memory consolidation system maintains principles~$\P$, hypotheses~$\H$, and experiences~$\E$, which guide the embodied agent's actions~$a$; world feedback (observations~$o$, rewards~$r$) generates new experiences~$e$ that refine knowledge.
        \textbf{(b)}~Test-time learning on \textit{\taskParts}: \method (blue) improves continuously while no-memory baseline (gray) remains flat.
        \textbf{(c)}~Qualitative results on \textit{\taskParts}: green boxes show covered cells, red dashed boxes indicate potential collisions avoided, and blue circles highlight learned space-saving strategies.
    }
    \label{\fighero}
    \vspace{-1.5em}
\end{figure}

\begin{figure*}[t]
    \centering
    \includegraphics[width=0.98\textwidth]{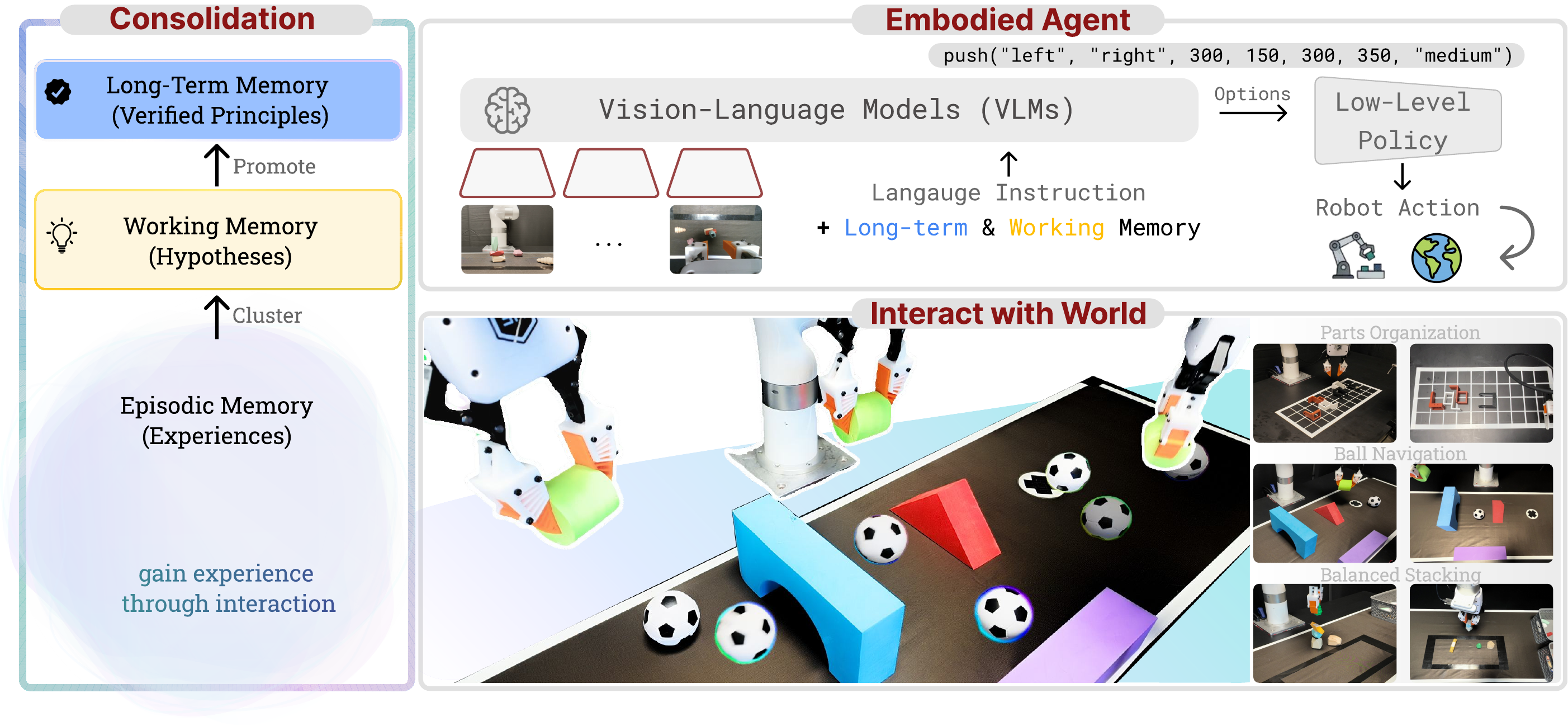}
    \caption{%
        \textbf{System overview of \method.}
        \textbf{Left (Consolidation):} A three-tier memory system stores raw experiences in \episodicMem, clusters them into testable hypotheses in \workingMem, and promotes verified knowledge to \longtermMem as principles. The consolidation process continuously refines memory through interaction.
        \textbf{Top-right (Embodied Agent):} A Vision-Language Model receives language instructions along with retrieved principles and active hypotheses from memory, then outputs high-level decisions that are executed by a low-level policy.
        \textbf{Bottom-right (World Interaction):} The agent interacts with challenging physical tasks (\taskParts, \taskBall, and \taskStack), and outcomes feed back into the memory system as new experiences.
    }
    \label{\figpipeline}
    \vspace{-1.5em}
\end{figure*}

A natural approach is to add memory, storing experiences and retrieving them when the scene appears similar~\citep{blundell2016episodic,pritzel2017neural,shi2025memoryvla,sridhar2025memer}.
However, embodied situations rarely repeat exactly, and retrieval-augmented approaches~\citep{lewis2020retrieval,gao2023rag,li2025mapvla} apply past experience without verifying whether it still holds.
Our experiments make this concrete: on a controlled brick insertion benchmark, retrieving raw experience reaches only 23\% success while principled abstraction reaches 76\%.
The issue is that unverified memory hardens into a fixed rule, and a small change in friction or object shape then turns a useful heuristic into a repeated error.
Avoiding this failure mode calls for verified abstraction, not just better recall.
People do not memorize every contact event. Instead, they form compact principles and revise them when evidence contradicts expectation~\citep{popper1959logic}.

We introduce \method, a test-time memory framework that enables VLM robot planners to learn physical principles through a \scientificloop.
The system records experiences and detects surprises, defined as outcomes that violate currently held principles. It clusters related experiences to generate candidate hypotheses, uses action-level attribution to isolate planning decisions from execution noise, and verifies hypotheses through targeted interaction before promoting them to long-term storage.
We further introduce Memory folding, which compresses supporting episodes into promoted principles, keeping context tractable over extended deployment~\citep{liu2021lifelong}.
The resulting principle set is human-readable and can be inspected, edited, or transferred to new settings.

Across real-world runs lasting over 30 minutes, \method enables clear learning curves: for \taskParts, performance improves from -1 to 9.7 with memory while remaining near 0 without; for \taskBall, the gap is 14.7 versus 0.7.
In transfer experiments, prior principles provide strong starting points when physics are similar, but test-time adaptation (ours) becomes essential when dynamics change, improving success from 10\% to 40\% on novel ball types.
Simulation experiments across four \vlm backbones confirm that our principled abstraction outperforms direct retrieval by 53\%, with performance stabilizing around 16--64 principles after an initial high-variance phase.
These results suggest that structured test-time learning can meaningfully improve \vlm planners on physical tasks, producing principles that are both effective for decision-making and interpretable for human inspection.

\noindent\textbf{Contributions.} We make three contributions.
(i)~We introduce \method, a test-time memory framework that uses targeted interaction to verify typed hypotheses about a deployment context before they condition a frozen VLM planner.
(ii)~We evaluate \method on three real-world manipulation tasks and a brick-insertion simulator across four VLM backbones; principled abstraction reaches 76\% on medium-difficulty insertion, a +23-point gain over the no-memory baseline (53\%) and +15 points over a free-form Reflexion-style baseline (61\%), while direct experience retrieval drops to 23\%, below no memory.
(iii)~We show that the gain scales with base-VLM capability, rising from +5\% on Gemini-ER-1.5 to +23\% on Gemini-3-Flash, which fits the picture of a system surfacing what the model already knows rather than installing new physics.

\vspace{-1.2em}
\section{Related Work}
\label{sec:related_work}


\subsection{Vision-Language-Action Models for Robot Planning}
Vision-language models have enabled natural language task specification and common-sense reasoning for robot planning~\citep{ahn2022saycan,driess2023palme,huang2022inner,liang2023code}.
The Robotics Transformer series~\citep{brohan2022rt1,brohan2023rt2} demonstrated that web-scale pretraining transfers to robotic control, motivating large-scale cross-embodiment efforts~\citep{open_x_embodiment2023,team2024octo,kim2024openvla}.
Recent work has pushed toward more capable generalist policies through flow matching~\citep{black2024pi0}, embodied reasoning~\citep{gemini2025robotics,bjorck2025groot,helix2025figure}, and chain-of-thought mechanisms~\citep{zawalski2024ecot,zhao2025cotvla}.
Comprehensive analyses~\citep{kawaharazuka2025vla,li2024robovlms,zhang2026vlm4vla} reveal that VLM general capabilities poorly predict downstream VLA performance, highlighting a persistent domain gap between declarative knowledge and physical grounding.
These approaches generalize well from pre-trained knowledge but cannot adapt to novel physical properties encountered at test time, which is the gap \method addresses by grounding VLM reasoning in interaction experience.


\subsection{Adaptation Paradigms for Deployed Agents}
\label{sec:related:paradigms}

How should a deployed VLM/VLA planner improve as it interacts with the world? The literature offers four broad answers, summarized in \tabref{tab:paradigm-taxonomy}. The most ambitious option updates the model itself, through online reinforcement learning~\citep{luo2024rlvla,xu2025tarl,shi2024yell,pi2025pi06}, meta-learning~\citep{finn2017maml}, test-time training~\citep{sun2020ttt}, or imitation finetuning of a memory-augmented backbone, as in MemER~\citep{sridhar2025memer}, MemoryVLA~\citep{shi2025memoryvla,li2024memonav}, MEM~\citep{torne2026mem}, and SAM2Act~\citep{fang2025sam2act}. A second family keeps the base frozen and feeds it context through retrieval~\citep{lewis2020retrieval,gao2023rag,li2025mapvla} or natural-language reflection~\citep{shinn2023reflexion,madaan2023selfrefine}. The piece they leave out is a check on whether a remembered experience still applies in the current scene; our experiments make this concrete: selective retrieval matches the no-memory baseline (53\% on medium), and free-form reflection plateaus at 61\%.

\method sits in a fourth corner of this space: a frozen base paired with structured typed memory and a verification step before commitment. For that reason we benchmark against training-free proxies of MemER's and MemoryVLA's retrieval mechanisms rather than their full systems, since the parameter-update side of those methods is not directly comparable in a frozen-base setting.

\begin{table*}[t]
    \centering
    \small
    \caption{\textbf{Adaptation paradigms for deployed VLM/VLA agents.} \method sits in a separate row: frozen base, structured memory, and a verification step before commitment.}
    \label{tab:paradigm-taxonomy}
    \setlength{\tabcolsep}{6pt}
    \renewcommand{\arraystretch}{1.15}
    \begin{tabular}{l l c l c c}
        \toprule
        Paradigm & Examples & Param.\ update & Memory content & Interp. & Inf.\ cost \\
        \midrule
        Online RL & RECAP, $\pi_{0.6}$ & \cmark\ Gradient & Implicit weights & \xmark & sec \\
        Trained memory VLA & MemER, MemoryVLA, MEM & \cmark\ Imit.\ FT & Visual / latent / text & \xmark & ms \\
        Reflection (in-episode) & Reflexion, Self-Refine & \xmark & Free-form text & \cmark & sec \\
        Retrieval & RAG, MAP-VLA & \xmark & Raw experiences & $\triangle$ & ms \\
        \midrule
        \textbf{Verified abstraction (ours)} & \textbf{\method} & \xmark & \textbf{Verified principles} & \cmark & ms \\
        \bottomrule
    \end{tabular}
\end{table*}


\subsection{Test-Time Learning in Robotics}
Test-time adaptation enables robots to adjust to new conditions without full retraining.
Meta-learning approaches~\citep{finn2017maml,finn2017oneshot} learn initializations for rapid few-shot adaptation, while domain randomization~\citep{tobin2017domain,andrychowicz2020dactyl} addresses sim-to-real transfer.
Sequence modeling formulations~\citep{chen2021decision,laskin2022incontext} enable in-context learning without gradient updates.
Recent work explores online adaptation for VLAs through reinforcement learning~\citep{luo2024rlvla,xu2025tarl}, language corrections~\citep{shi2024yell}, and learning from deployment experience~\citep{pi2025pi06}.
Test-time training methods~\citep{sun2020ttt,kim2025feedtta,yoo2025wormi} update models on unlabeled test data through self-supervision.
These methods adjust the policy implicitly rather than learning explicit principles.
\method instead produces human-readable hypotheses that can be inspected, edited, or transferred to new settings.


\subsection{Memory in Embodied Agents}
Memory systems enhance robot planning through experience accumulation~\citep{blundell2016episodic,pritzel2017neural,fang2019scene}.
Recent work integrates memory directly into VLAs through hierarchical retrieval~\citep{sridhar2025memer}, cognitive-inspired dual memory banks~\citep{shi2025memoryvla,li2024memonav}, multi-scale embodied memory~\citep{torne2026mem}, and visual foundation models~\citep{fang2025sam2act}.
World models~\citep{hafner2023dreamerv3,bruce2024genie,agarwal2025cosmos,team2025sima2} learn to imagine future scenarios, enabling planning without real-world interaction.
Reflection-based approaches~\citep{shinn2023reflexion,madaan2023selfrefine} use LLMs to learn from failures, while retrieval-augmented methods~\citep{lewis2020retrieval,gao2023rag,li2025mapvla} provide relevant past experiences.
Lifelong learning methods~\citep{liu2021lifelong,meng2025legion,zhang2024extract} enable continuous accumulation without catastrophic forgetting.
Reflection that triggers only on failures misses what successes have to teach, and retrieval without a check applies experience even when the world has moved on.
\method's \scientificloop does both: it draws hypotheses from successes and failures, tests each through targeted interaction, and keeps only the survivors as grounded principles.

\section{Method}
\label{sec:method}

We present \textit{\method}, a test-time memory framework that lets a VLM robot planner learn physical principles through interaction. The mechanism is a \scientificloop that drafts hypotheses from experience, checks each against targeted experiments, and commits only the survivors to guide future decisions.


\subsection{Problem Formulation}
\label{sec:method:formulation}

We formulate physical manipulation as a sequential decision problem
following the options framework~\citep{sutton1999options}.
Let $\S$ denote the state space and $\O$ the observation space
(visual observations).
At each decision point $t$, the agent receives observation $o_t \in \O$
and task description $\tau$.
Following Sutton et al., we define an \emph{option} $\omega = \langle \mathcal{I}, \pi, \beta \rangle$
as a temporally-extended action consisting of an initiation set $\mathcal{I} \subseteq \S$,
an intra-option policy $\pi: \S \times \A \to [0,1]$, and a termination
condition $\beta: \S \to [0,1]$.

In our framework, a VLM-based high-level policy $\pi_\theta^H$ selects
options based on observations, task context, and learned principles:
\begin{equation}
    \omega_t = \pi_\theta^H(o_t, \tau, \P_t)
    \label{eq:highlevel}
\end{equation}
where $\P_t \subseteq \P$ denotes the set of active principles retrieved
from memory at time $t$.
The selected option $\omega_t$ is executed by a low-level policy
$\pi^L$ (which may be a motion planner, VLA, or other controller)
until termination.
The challenge is that physical properties (friction, dynamics, spatial
relationships) vary across objects and environments, and cannot be
fully captured by pre-trained VLM knowledge $\theta$ alone.

Our goal is to learn a principle set $\P^*$ that grounds the VLM's
reasoning in interaction experience, such that:
\begin{equation}
    \Exp\left[\sum_{t=0}^{T} r_t \mid \pi_\theta^H(\cdot, \cdot, \P^*)\right] >
    \Exp\left[\sum_{t=0}^{T} r_t \mid \pi_\theta^H(\cdot, \cdot, \varnothing)\right]
    \label{eq:objective}
\end{equation}
where $r_t$ is the reward at step $t$.
Critically, $\P^*$ must be learned \emph{at test time} without
modifying the VLM parameters $\theta$.

\begin{algorithm}[t]
\caption{Scientific Memory Loop}
\label{alg:scientific-loop}
\begin{algorithmic}[1]
\small
\Require Experience buffer $\E$, Hypothesis store $\H$, Principle store $\P$
\Require Reflection model $f_\phi$, thresholds $\tau_p, \tau_r, n_{\min}$
\Statex
\Function{RecordExperience}{$o, \omega, r, c, \mathbf{s}, \P_{\text{active}}$}
    \State $e \gets (o, \omega, r, c, \mathbf{s})$
    \State $\rho \gets \text{Resonance}(e, \P_{\text{active}})$ \Comment{Eq.~\ref{eq:resonance}}
    \State $\E \gets \E \cup \{e\}$
    \If{$\rho < 1$} \Comment{Surprising experience}
        \State \Call{TriggerConsolidation}{$e$}
    \Else
        \State \Call{ReinforceActivePrinciples}{$\P_{\text{active}}$}
    \EndIf
\EndFunction
\Statex
\Function{Consolidate}{$\E$}
    \State $\{\mathcal{C}_k\} \gets \text{ClusterBySymbolicState}(\E)$
    \For{each cluster $\mathcal{C}_k$ with $|\mathcal{C}_k| \geq n_{\min}$}
        \State $\H_k \gets f_\phi(\mathcal{C}_k, \P, \H)$ \Comment{Eq.~\ref{eq:hypothesis}}
        \State $\H \gets \H \cup \H_k$
    \EndFor
\EndFunction
\Statex
\Function{AttributeAndPromote}{$\E, \H, \P$}
    \For{each hypothesis $h \in \H$}
        \State Update $\text{conf}(h)$ via action-level attribution \Comment{Eq.~\ref{eq:attribution}}
        \If{$\text{conf}(h) \geq \tau_p$ \textbf{and} $|\E_{\text{support}}| \geq 3$}
            \State $\P \gets \P \cup \{h\}$ \Comment{Promote to principle}
            \State $\E \gets \E \setminus \E_{\text{folded}}$ \Comment{Memory folding}
            \State $\H \gets \H \setminus \{h\}$
        \ElsIf{$\text{conf}(h) \leq \tau_r$ \textbf{and} $|\E_{\text{contradict}}| \geq 2$}
            \State $\H \gets \H \setminus \{h\}$ \Comment{Refute hypothesis}
        \EndIf
    \EndFor
\EndFunction
\end{algorithmic}
\end{algorithm}


\subsection{System Overview}
\label{sec:method:overview}

\figref{\figpipeline} illustrates the \method architecture.
The system consists of three components:
(1) a VLM-based \textbf{Planner} that generates high-level decisions
given observations and principles,
(2) a \textbf{Memory} system organized in three tiers (episodic,
working, and long-term), and
(3) an \textbf{Executor} that carries out planned actions via
low-level policies.

At each episode, the planner queries memory for relevant principles,
makes decisions, and observes outcomes.
What makes the loop work is that memory is not a static database; it
evolves through a \scientificloop that continuously refines raw
experiences into verified principles.
Where retrieval-augmented approaches apply retrieved knowledge as-is, \method tests each hypothesis before commitment, which avoids the dogmatism problem of stale experience hurting performance.

\subsection{Scientific Memory Loop}
\label{sec:method:loop}

The \scientificloop is the central mechanism of \method. It moves raw experience through verified principles in four phases: experience collection, hypothesis generation, attribution, and principle promotion.
\algoref{alg:scientific-loop} provides the pseudocode.

\paragraph{Experience Collection with Resonance Checking}
We store experiences from both successful and unsuccessful interactions.
Each experience $e = (o, \omega, r, c, \mathbf{s})$ records the
observation $o$, selected option $\omega$, outcome $r \in \{0, 1\}$,
context $c$ (task description, subtask), and symbolic state
$\mathbf{s}$ (discrete features like action type, object properties).

A central mechanism is resonance checking, which measures how well
an experience matches active principles.
Let $\P_{\text{active}} \subseteq \P$ be the principles applied
during decision-making.
We compute a resonance score:
\begin{equation}
    \rho(e, \P_{\text{active}}) = \frac{|\{p \in \P_{\text{active}} : \text{consistent}(e, p)\}|}{|\P_{\text{active}}|}
    \label{eq:resonance}
\end{equation}
where $\text{consistent}(e, p)$ checks whether the experience outcome
aligns with the principle's prediction.
When $\rho < 1$ (a ``surprise''), the experience is prioritized for
consolidation; when $\rho = 1$, the experience reinforces existing
principles without triggering new hypothesis generation.
This surprise-driven filtering focuses learning on novel situations.

\paragraph{Hypothesis Generation}
Experiences are periodically clustered by symbolic similarity.
For each cluster $\mathcal{C}_k$ with sufficient experiences
($|\mathcal{C}_k| \geq n_{\min}$), we use a reflection model
$f_\phi$ (VLM in real-world, LLM in simulation) to generate
hypotheses about patterns:
\begin{equation}
    \H_k = f_\phi(\mathcal{C}_k, \P, \H_{\text{existing}})
    \label{eq:hypothesis}
\end{equation}
where $\P$ and $\H_{\text{existing}}$ provide context to avoid
generating duplicate knowledge.
Each hypothesis $h \in \H_k$ takes a typed form:
\begin{compactitem}
    \item \prinAvoid: ``Don't do $X$ when $Y$'' (from failures)
    \item \prinPrefer: ``Do $X$ when $Y$'' (from successes)
    \item \prinSequence: ``Do $X$ before $Y$'' (temporal constraints)
\end{compactitem}

\paragraph{Action-Level Attribution}
Hypotheses are judged by action-level outcomes, not episode-level success.
For hypothesis $h$ about action type $a^*$, we update confidence using
only experiences where that specific action was attempted:
\begin{equation}
    \text{conf}(h) \gets \text{conf}(h) + \alpha \cdot
    \frac{|\{e \in \E_h : a_e = a^*,\, r_e = 1\}|}
         {|\{e \in \E_h : a_e = a^*\}|}
    \label{eq:attribution}
\end{equation}
where $\E_h$ denotes experiences relevant to $h$ and $\alpha$ is the learning rate.
This isolates specific action effects from confounding factors.

\paragraph{Verification and Principle Promotion}
Hypotheses achieving high confidence ($\text{conf}(h) \geq \tau_p$,
typically 0.8) with sufficient supporting evidence
($|\E_{\text{support}}| \geq 3$) are promoted to principles.
Upon promotion, source experiences are ``folded'' into the principle,
compressing episodic memory while preserving learned knowledge:
\begin{equation}
    \P \gets \P \cup \{h\}, \quad
    \E \gets \E \setminus \E_{\text{folded}}
    \label{eq:promotion}
\end{equation}
Hypotheses with low confidence ($\text{conf}(h) \leq \tau_r$) and
contradicting evidence are refuted and removed.


\begin{figure}[t]
    \centering
    \includegraphics[width=0.95\columnwidth]{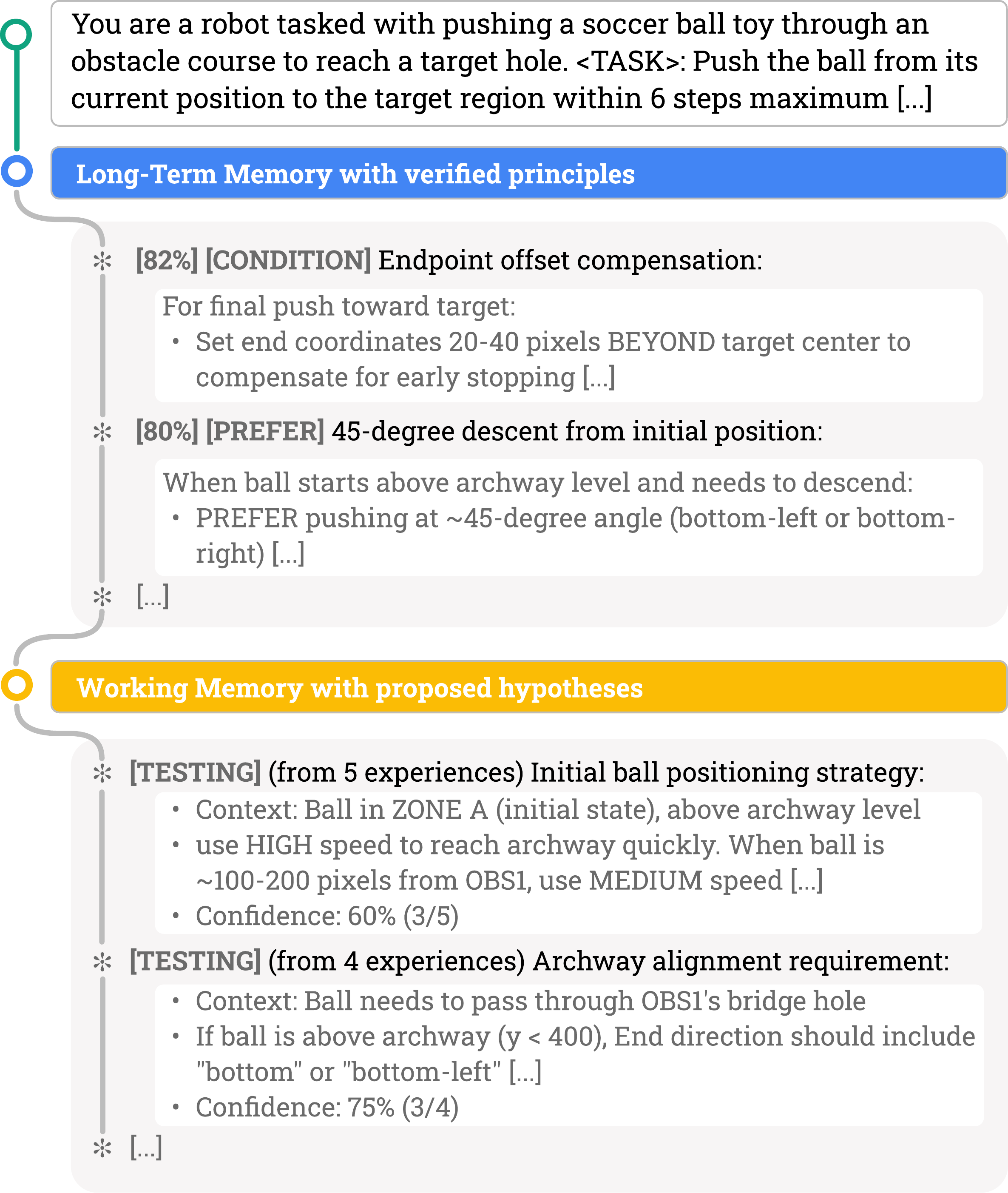}
    \caption{%
        \textbf{Memory injection into VLM prompts.}
        Verified principles (blue) and active hypotheses (yellow) are
        inserted into the planner's context with confidence scores and
        typed constraints (\prinPrefer, \prinAvoid, \prinSequence).
    }
    \label{\figloop}
    \vspace{-1.5em}
\end{figure}

\subsection{Memory Architecture}
\label{sec:method:architecture}

\method organizes memory in three tiers:
(1)~\textbf{Episodic memory} stores raw experiences with symbolic
state for efficient filtering, bounded by capacity $N_{\max}$;
(2)~\textbf{Working memory} holds hypotheses under test, each with
confidence scores from supporting/contradicting evidence;
(3)~\textbf{Long-term memory} contains verified principles that
guide decisions, with importance scores that decay over time
($\gamma = 0.995$) to forget outdated knowledge.


\subsection{Retrieval and Application}
\label{sec:method:retrieval}

At inference time, \method retrieves relevant principles via symbolic
filtering (matching action type and object properties) followed by
semantic ranking.
The top-$k$ principles and active hypotheses are injected into the
VLM prompt as shown in \figref{\figloop}.
When resonance is low, the system prioritizes fresh learning over
applying potentially misleading prior knowledge.
Full implementation details are in \appendref{app:method-details}.


\section{Experimental Setup}
\label{sec:setup}

\begin{figure}[t]
    \centering
    \includegraphics[width=\columnwidth]{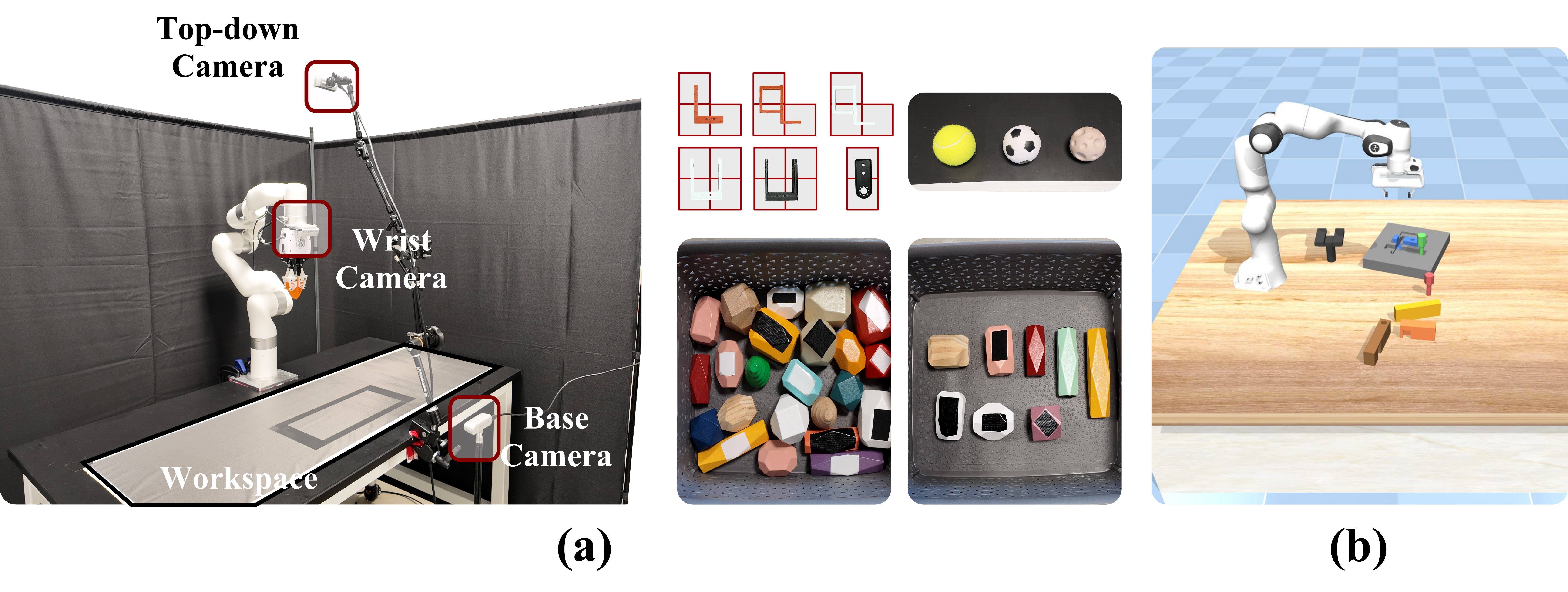}
    \caption{%
        \textbf{Experimental environments.}
        (a)~Left: Real-world platform with xArm6 robot, fin-ray soft grippers,
        and multi-view RealSense cameras in an enclosed workspace.Right: The partial props used in the experiments.
        (b)~Reflect-VLM simulation~\citep{feng2025reflective} with Franka
        Panda robot for large-scale experiments.
    }
    \label{fig:environment}
    \vspace{-1.5em}
\end{figure}

We evaluate \method in two complementary settings: real-world manipulation
tasks that require physical understanding VLMs lack from pretraining, and
simulation experiments that enable large-scale analysis across VLMs and
difficulty levels. \figref{fig:environment} shows both environments.

Our experimental design deliberately separates high-level reasoning from low-level control.
The VLM outputs discrete decisions (which part to place, which direction to push, which stone to stack), while motion planning handles execution.
This separation isolates the variable under study: if performance improves, the improvement must stem from better reasoning since the executor remains fixed.
We evaluate both task success and plan quality, since success rate alone can be misleading, where a planner might achieve high success through conservative strategies that do not demonstrate physical understanding.

\subsection{Real-World Tasks}

\begin{figure}[t]
    \centering
    \includegraphics[width=\columnwidth]{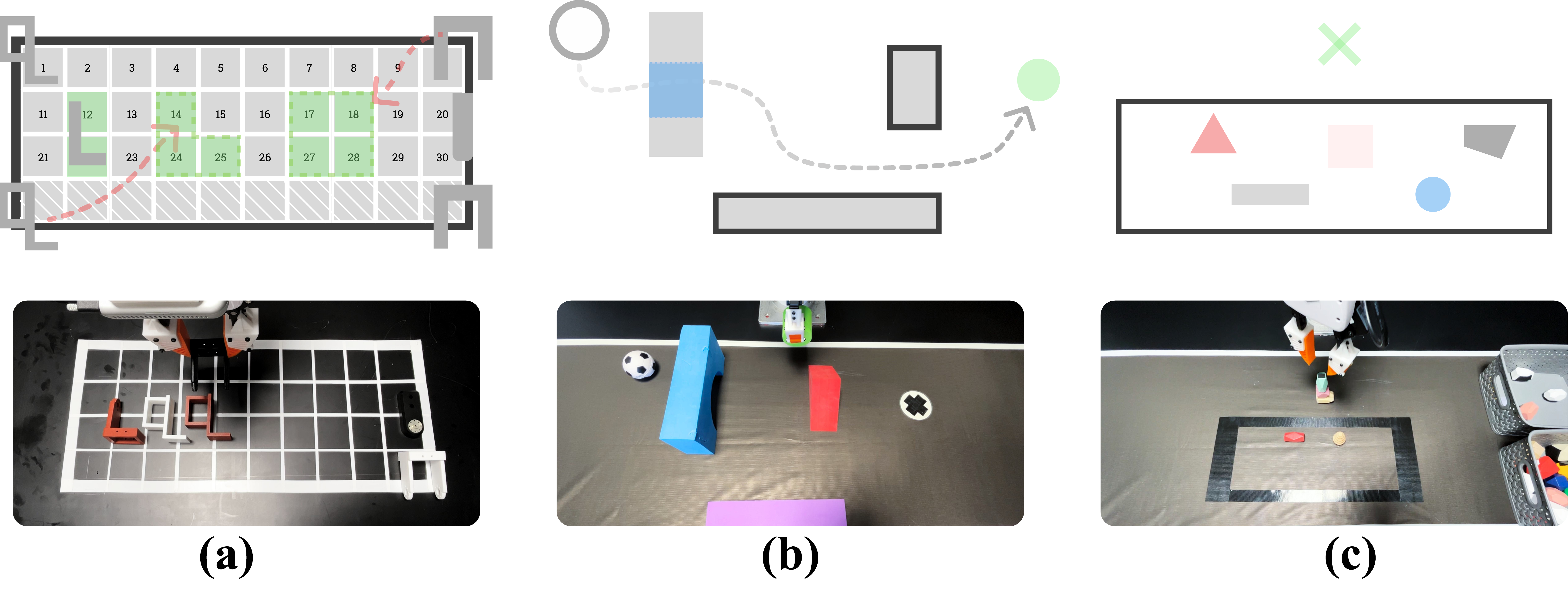}
    \caption{%
        \textbf{Real-world tasks.}
        Top: symbolic representations; bottom: actual setups.
        (a)~Grid layout and placement trajectories for \taskParts.
        (b)~Obstacle course and ball trajectory for \taskBall.
        (c)~Stone arrangement and stacking position for \taskStack.
    }
    \label{fig:task-overview}
    \vspace{-1.5em}
\end{figure}

Our real-world platform uses an xArm6 with the TPU-printed fin-ray
soft grippers and Intel RealSense D435 cameras (top-down at
1280$\times$720, wrist-mounted at 640$\times$480). For VLM planning,
we use Gemini-3.0-Flash~\citep{google2025gemini3flash} with thinking mode.
Hypothesis generation uses Qwen3-VL~\citep{qwen3vl2025}, running
asynchronously every 15 seconds. Each task runs for 10--20 episodes.
We design three tasks where correct reasoning requires physical parameters that vision alone cannot provide (\figref{fig:task-overview}):

\paragraph{\textbf{\taskParts}}
Place 6 irregularly-shaped parts onto a 3$\times$10 grid, minimizing total cells occupied.
Each part occupies 2--4 cells; the VLM outputs placement indices.
A good plan exploits interlocking geometries to pack efficiently.
The challenge: part shapes allow overlapping in 3D when aligned correctly, but these spatial relationships only emerge through placement attempts.

\paragraph{\textbf{\taskBall}}
Push a soccer ball through obstacles to reach the target within 6 steps.
The VLM specifies push direction, coordinates, and speed; scoring rewards progress toward the goal.
A good plan accounts for rolling distance and obstacle rebounds.
The challenge: surface friction and ball elasticity vary across the workspace, requiring trial-and-error calibration.

\paragraph{\textbf{\taskStack}}
Build a stable tower from 5 balance stones with varying sizes, textures, and weight distributions.
The VLM selects stacking order; scoring rewards height and penalizes collapses.
A good plan sequences stones by stability contribution.
The challenge: friction and weight distribution are invisible, and only contact reveals which pairings hold. Full task specifications appear in \appendref{app:realworld}.

\subsection{Simulation Benchmark}

We complement real-world evaluation with the Reflect-VLM brick insertion
benchmark~\citep{feng2025reflective}, a MuJoCo-based environment with a
Franka Panda robot. The VLM must learn correct insertion ordering:
placing the wrong brick first blocks subsequent insertions. Difficulty
scales with brick count: easy (2--3), medium (4--5), hard (6--8).
This controlled setting enables experiments at scale (500+ episodes)
across multiple VLMs. Details appear in \appendref{app:simulation}.


\section{Experiments}
\label{sec:experiments}

Our experiments address five questions:
\begin{compactenum}
    \item \textbf{Memory Paradigms}: How does \method compare against
          existing memory paradigms (selective retrieval, token-memory,
          and reflection) for deployed VLM agents?
          (\secref{sec:exp:paradigms})
    \item \textbf{Physics Understanding}: Can \method improve the VLM's
          understanding of task-specific physical properties through
          interaction? (\secref{sec:exp:qualitative})
    \item \textbf{Decision Quality}: Does improved physics understanding
          lead to better embodied decision-making over time?
          (\secref{sec:exp:evolution})
    \item \textbf{Memory Transfer}: Can experience-derived memory provide
          benefits in out-of-distribution scenarios, and when does it fail?
          (\secref{sec:exp:transfer})
    \item \textbf{Architecture Design}: Which components of \method are
          essential, and why do we need memory compression and forgetting?
          (\secref{sec:exp:ablation})
\end{compactenum}


\subsection{Comparison Against Memory Paradigms}
\label{sec:exp:paradigms}

We compare \method against five baselines, one for each main memory paradigm used by deployed VLM agents, all run under the same frozen-base setup (Gemini-3-Flash, identical task budget, top-$K{=}5$ memory injection where applicable). MemER~\citep{sridhar2025memer} and MemoryVLA~\citep{shi2025memoryvla} require parameter updates (imitation finetuning or end-to-end VLA training); running their full systems would take us out of the frozen-base setting, so we benchmark training-free proxies that isolate their retrieval mechanism (implementation details in \appendref{app:method-details}). The five baselines: \textbf{No Memory}; \textbf{Direct Retrieval} (top-$K$ raw experiences by symbolic-state similarity, no abstraction); \textbf{MemER-style Retrieval} (VLM-judged comparative ranking, top-$K{=}5$, no verification); \textbf{MemoryVLA-style Token Memory} (full episode history accumulated as context); and \textbf{Reflexion-style}~\citep{shinn2023reflexion} (free-form reflection on successes and failures, retrieval by semantic similarity).

\begin{table}[t]
    \centering
    \small
    \caption{\textbf{Main results: \method vs.\ memory paradigms on \taskBrick} (100 episodes per cell, 3 seeds). MemER-style retrieval lands at the No-Memory baseline on medium (53\%), and only \method's verified principles exceed No Memory on hard. Best per-difficulty in bold.}
    \label{tab:paradigm-main}
    \setlength{\tabcolsep}{4pt}
    \renewcommand{\arraystretch}{1.15}
    \begin{tabular}{l | c c c | c}
        \toprule
        & \multicolumn{3}{c|}{Success Rate} & \\
        \cmidrule(lr){2-4}
        Method & Easy & Med. & Hard & Tokens \\
        \midrule
        No Memory                       & 83\% & 53\% & 28\% & 0.40$\times$ \\
        Direct Retrieval                & 48\% & 23\% & 8\%  & 0.55$\times$ \\
        MemER-style Retrieval           & 85\% & 53\% & 19\% & 1.0$\times$ \\
        MemoryVLA-style Token Memory    & 69\% & 35\% & 13\% & 2.5$\times$ \\
        Reflexion-style                 & 86\% & 61\% & 31\% & 1.2$\times$ \\
        \midrule
        \textbf{\method (Full)}         & \textbf{89\%} & \textbf{76\%} & \textbf{39\%} & 1.0$\times$ \\
        \bottomrule
    \end{tabular}
    \vspace{-0.5em}
\end{table}

Three patterns come out of \tabref{tab:paradigm-main}, and together they pin the gain to abstraction and verification rather than memory access itself. Selective retrieval is no better than no memory: MemER-style retrieval lands at 53\% on medium, exactly the No-Memory baseline, recovering Direct Retrieval's 30-point loss but adding nothing on top. Unverified memory then hurts once situations stop repeating: on hard, all three unverified variants drop below the 28\% No-Memory floor (Direct Retrieval 8\%, MemoryVLA-style 13\%, MemER-style 19\%), and only \method's verified principles (39\%) come out ahead. More context is not more help either: MemoryVLA-style trails MemER-style by 18 points at 2.5$\times$ the token cost, and \method's lead over MemER-style is largest on the harder tasks (4\%/23\%/20\% on easy/medium/hard), the opposite of what a better-retrieval interpretation would predict.


\subsection{Learned Physical Principles}
\label{sec:exp:qualitative}

\begin{figure*}[t]
    \centering
    \includegraphics[width=\textwidth]{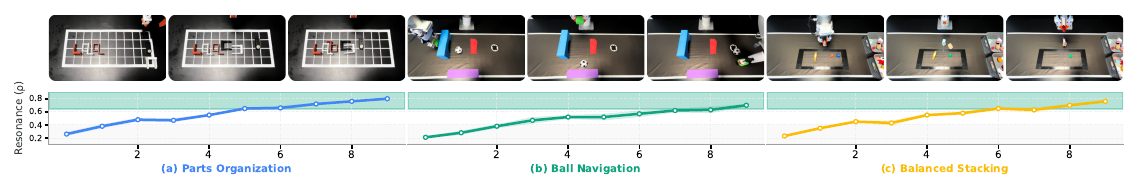}
    \caption{%
        \textbf{Resonance score evolution across tasks.}
        Each panel shows RGB keyframes alongside the resonance curve.
        Resonance $\rho$ measures prediction-outcome alignment: low values
        indicate surprising outcomes that trigger learning; high values
        (green region) indicate principled reasoning.
        \textbf{Left}: \taskParts progresses from scattered placement to
        efficient packing.
        \textbf{Middle}: \taskBall learns ball dynamics through trial.
        \textbf{Right}: \taskStack discovers stable stone combinations.
        All tasks transition from reactive ($\rho < 0.5$) to rational
        ($\rho > 0.7$) behavior within 10 episodes.
    }
    \label{\figtasks}
    \vspace{-2.0em}
\end{figure*}

Beyond task scores, we examine whether \method produces reasoning that becomes better grounded in the physics of the scene. \figref{\figtasks} shows the resonance score $\rho$, which measures how well predictions align with observed outcomes. High resonance means the planner anticipated task dynamics correctly; low resonance means the outcome surprised the active principles. All three tasks follow a consistent pattern: $\rho$ climbs from $\approx 0.2$ in early episodes to $\approx 0.9$ by episode 10, crossing $\rho = 0.7$ around episode 5--6, after which predictions match outcomes more often than not.

\begin{figure*}[t]
    \centering
    \includegraphics[width=0.98\textwidth]{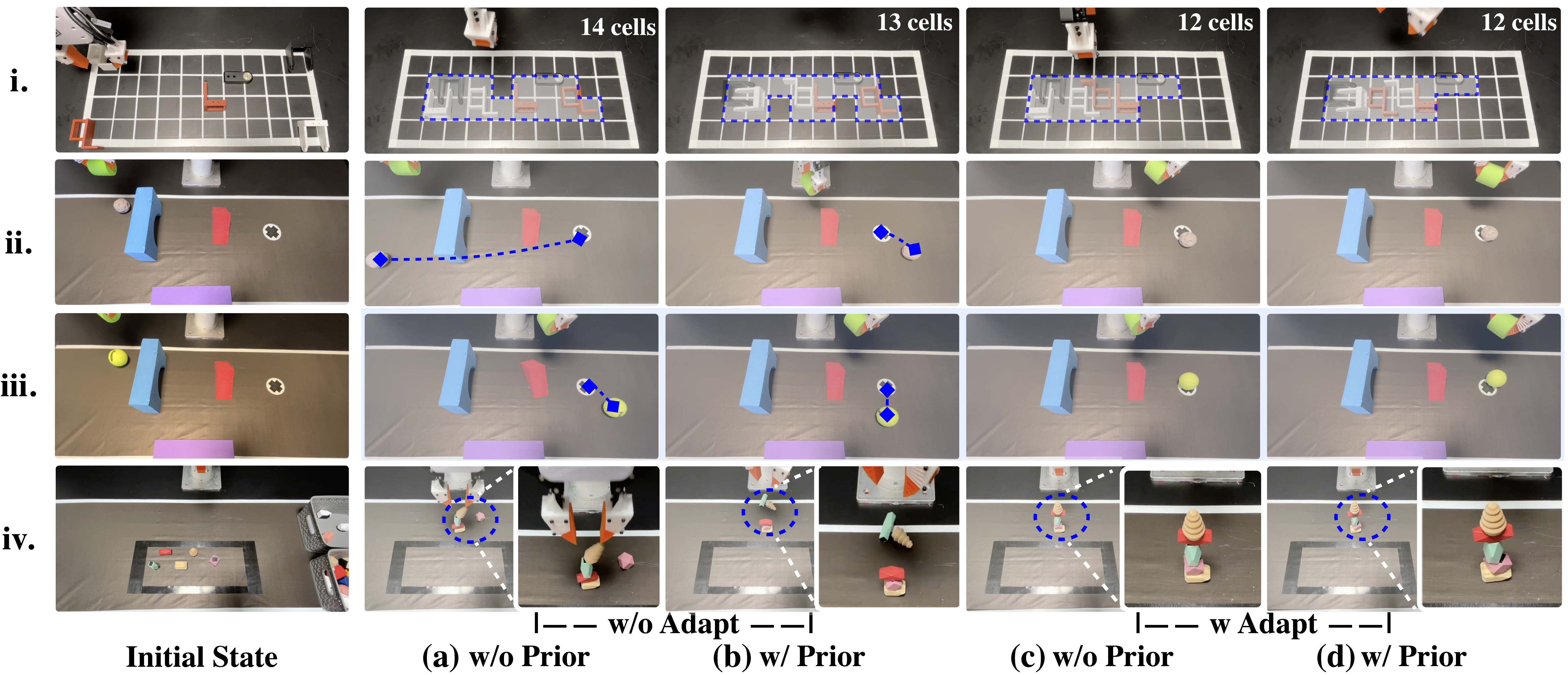}
    \caption{%
        \textbf{End-state keyframes across tasks and OOD variants.}
        A $2{\times}2$ ablation of prior knowledge and on-the-fly adaptation, shown as final states for four task instances. Rows: (i)~\taskParts on its OOD configuration (covered cells, lower is better); (ii)~\taskBall with the in-distribution soccer ball; (iii)~\taskBall with an OOD tennis ball; (iv)~\taskStack. Columns: (a)~no prior, no adaptation; (b)~prior only; (c)~adaptation only; (d)~full \method. Across all four rows, the full setting reaches the most efficient or most stable end state; either component alone leaves a visible gap that \secref{sec:exp:transfer} quantifies.
    }
    \label{fig:transfer-keyframes}
    \vspace{-1.1em}
\end{figure*}

This metric captures reasoning quality that task scores alone cannot reveal.
Raw success rates do not distinguish a planner that understands physics from one that finds safe but suboptimal strategies (always pushing slowly, avoiding complex placements).
High resonance asks the planner's internal model to actually predict the outcome rather than just produce an acceptable result.
Episodes in the high-resonance regime ($\rho > 0.7$) score $2.3\times$ higher than episodes 1--3.
The steady rise in $\rho$ indicates that \method accumulates verified physical understanding instead of overfitting to a narrow solution.

The principles themselves are human-readable: spatial constraints for
\taskParts (``avoid overlapping internal regions of q-shaped parts''),
dynamics rules for \taskBall, and stability heuristics for \taskStack
(``select the largest high-friction stone as base''). Persistent failures
occur when learned principles conflict or do not apply. Full principle
inventories appear in \appendref{app:principles-examples}.

\subsection{Test-Time Evolution}
\label{sec:exp:evolution}
\begin{figure}[t]
    \centering
    \includegraphics[width=0.5\textwidth]{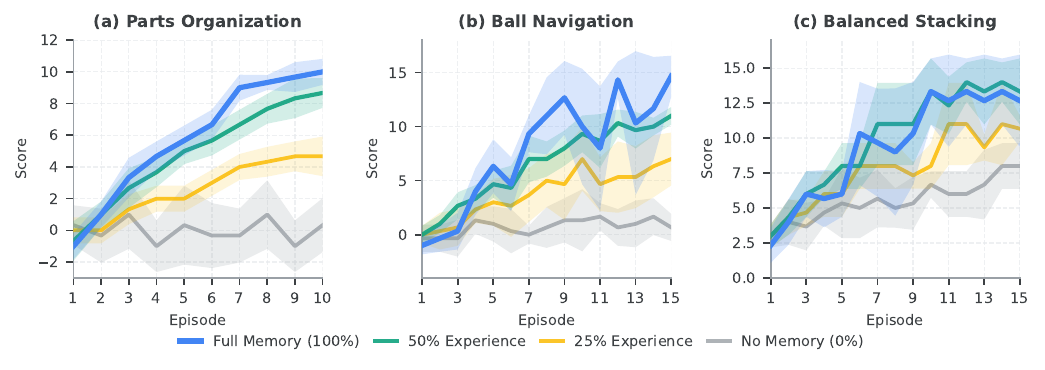}
    \caption{%
        \textbf{Test-time evolution across experience utilization levels.}
        Each panel shows learning curves with shaded standard deviation
        bands (3 runs per condition).
        \textbf{Blue}: Full memory (100\%);
        \textbf{Green}: 50\% experience;
        \textbf{Yellow}: 25\% experience;
        \textbf{Gray}: No memory (0\%).
        Without memory, performance remains flat across all tasks.
        Complex dynamics (\taskBall) benefit most from full experience,
        while simpler tasks (\taskStack) show diminishing returns.
    }
    \label{\figcurves}
    \vspace{-1.5em}
\end{figure}

To evaluate how \method actually improves through interaction and how much experience is needed, we evaluate performance under different experience utilization levels (0\%, 25\%, 50\%, 100\%), with 3 runs per condition. The test-time learning curve is shown in \figref{\figcurves}.

Without memory (0\%), performance remains flat; with full memory, \taskParts
improves from -1$\to$9.7 and \taskBall shows an even larger gap (14.7 vs.\ 0.7).
Task complexity determines experience requirements: \taskBall benefits from
full experience (50\% vs.\ 100\%: 11.0 vs.\ 14.7), while \taskStack shows
diminishing returns (50\% nearly matches 100\%). The memory system efficiently
extracts generalizable principles when task structure permits.


\subsection{Memory Transfer}
\label{sec:exp:transfer}

Can learned principles transfer to out-of-distribution (OOD) scenarios?
We ablate prior knowledge (principles from in-distribution deployment)
and test-time adaptation (via \scientificloop). Each condition runs 10 trials. We report results in \tabref{\tabood} and details in \appendref{app:ood-settings}.

\definecolor{cellred1}{RGB}{255,235,235}
\definecolor{cellred2}{RGB}{255,210,210}
\definecolor{cellred3}{RGB}{255,180,180}
\definecolor{cellred4}{RGB}{248,150,150}
\definecolor{cellgreen1}{RGB}{235,250,235}
\definecolor{cellgreen2}{RGB}{210,242,210}
\definecolor{cellgreen3}{RGB}{180,230,180}
\definecolor{cellgreen4}{RGB}{150,218,150}

\begin{table}[b]
    \centering
    \vspace{-1.0em}
    \small
    \caption{%
        \textbf{Memory transfer to OOD task variants.}
        We ablate prior knowledge (Prior) and test-time adaptation (Adapt).
        Prior knowledge helps when physics are similar (\taskParts, \taskStack).
        For \taskBall with new ball types, adaptation is essential since
        dynamics differ substantially. Best results in \textbf{bold}.
        \colorbox{cellgreen3}{\strut Green}: better; \colorbox{cellred3}{\strut Red}: worse.
    }
    \label{\tabood}
    \setlength{\tabcolsep}{1.3pt}
    \begin{tabular}{@{}cc|cccccc@{}}
        \toprule
        & & \multicolumn{2}{c}{\taskParts} & \multicolumn{2}{c}{\taskBall} & \multicolumn{2}{c}{\taskStack} \\
        \cmidrule(lr){3-4} \cmidrule(lr){5-6} \cmidrule(lr){7-8}
        Prior & Adapt & Score & Succ. & Score & Succ. & Score & Succ. \\
        \midrule
        \xmark & \xmark & \cellcolor{cellred4}$-0.6$ & \cellcolor{cellred4}0/10 & \cellcolor{cellred3}1.6 & \cellcolor{cellred3}1/10 & \cellcolor{cellred4}6.7 & \cellcolor{cellred4}4/10 \\
        \xmark & \cmark & \cellcolor{cellred2}3.3 & \cellcolor{cellred3}1/10 & \cellcolor{cellgreen1}5.5 & \cellcolor{cellred2}2/10 & \cellcolor{cellred2}8.3 & \cellcolor{cellgreen1}7/10 \\
        \cmark & \xmark & \cellcolor{cellgreen1}6.9 & \cellcolor{cellred2}3/10 & \cellcolor{cellred2}2.9 & \cellcolor{cellred3}1/10 & \cellcolor{cellgreen1}9.2 & \cellcolor{cellgreen2}8/10 \\
        \cmark & \cmark & \cellcolor{cellgreen3}\textbf{8.3} & \cellcolor{cellgreen3}\textbf{4/10} & \cellcolor{cellgreen3}\textbf{7.1} & \cellcolor{cellgreen3}\textbf{4/10} & \cellcolor{cellgreen4}\textbf{12.3} & \cellcolor{cellgreen4}\textbf{9/10} \\
        \bottomrule
    \end{tabular}
    \vspace{-1.5em}
\end{table}

The results reveal when transfer succeeds and fails. For \taskStack, prior
knowledge alone achieves 80\% success because stability principles transfer
well to new stones. For \taskParts, prior knowledge improves score from
-0.6$\to$6.9 by avoiding common packing errors. In contrast, \taskBall
with new ball types shows that prior knowledge matches zero-shot performance: the new balls have different dynamics, so prior
principles do not transfer. Adding adaptation improves the score of \taskBall to 7.1 and 40\% success as the system learns new friction and elasticity
properties. The full \method achieves the best performance across all tasks,
confirming that prior knowledge and test-time adaptation are complementary.


The following experiments use the simulation benchmark for large-scale
analysis across VLMs and difficulty levels.
In simulation, we report success rates because the controlled environment enables fair comparison across hundreds of episodes, and the brick insertion task has a clear binary success criterion (all bricks correctly placed).
Real-world tasks use plan quality metrics to capture nuanced physical reasoning.


\subsection{Scaling Across VLMs}
\label{sec:exp:difficulty}

\begin{figure}[t]
    \centering
    \includegraphics[width=\columnwidth]{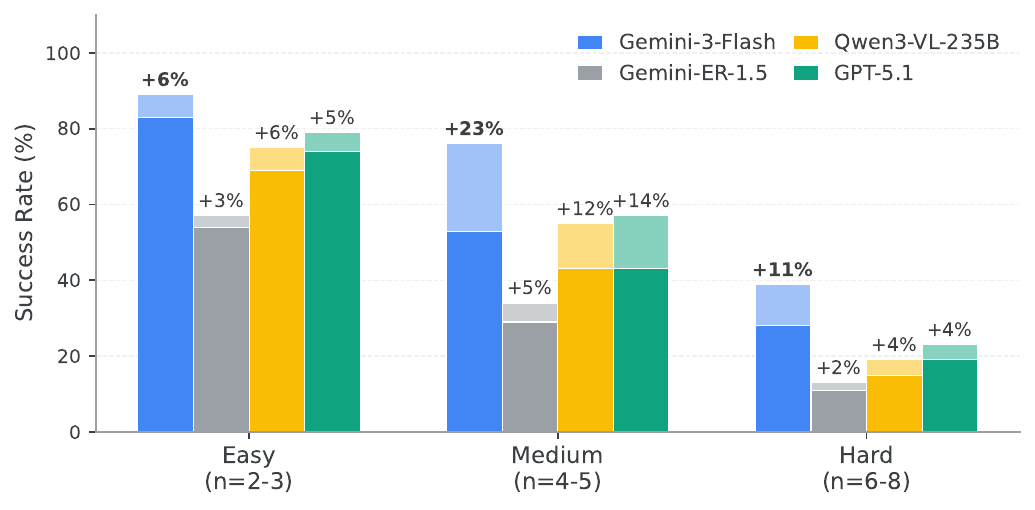}
    \caption{%
        \textbf{Success rates across VLMs and difficulty levels.}
        Bars show baseline performance (w/o \method); annotations
        indicate improvement with test-time learning.
        Gemini-3-Flash dominates across all difficulties and shows the
        largest improvement. Weaker models benefit less from test-time learning, suggesting that memory amplifies existing capabilities rather than compensating for fundamental limitations.
    }
    \label{\figdifficulty}
    \vspace{-1.5em}
\end{figure}

How does test-time learning interact with VLM capability and task difficulty?
We evaluate four VLMs with thinking mode: Gemini-3-Flash~\citep{google2025gemini3flash},
Gemini-ER-1.5~\citep{google2025geminirobotics}, GPT-5.1~\citep{openai2025gpt5},
and Qwen3-VL-235B~\citep{qwen3vl2025}. Each condition runs 100 episodes.
\figref{\figdifficulty} shows results. VLM capability sets the baseline. Gemini-3-Flash leads across difficulties, reaching 53\% on medium against 29--43\% for the other models. The gain from test-time learning then scales with that baseline: on medium, Gemini-3-Flash improves by $+23\%$ (53\% to 76\%), GPT-5.1 by $+14\%$, Qwen3-VL by $+12\%$, and Gemini-ER-1.5 by $+5\%$. The pattern fits a system that surfaces what the model already knows, since the planner needs enough capability in the first place to draft and verify useful hypotheses. All difficulty levels benefit, with medium showing the largest gain ($+23\%$); hard tasks still improve ($+11\%$) because accumulated failures yield \prinAvoid hypotheses that head off repeated mistakes.

\subsection{Principle Scaling}
\label{sec:exp:scaling}

\begin{figure}[t]
    \centering
    \includegraphics[width=0.95\columnwidth]{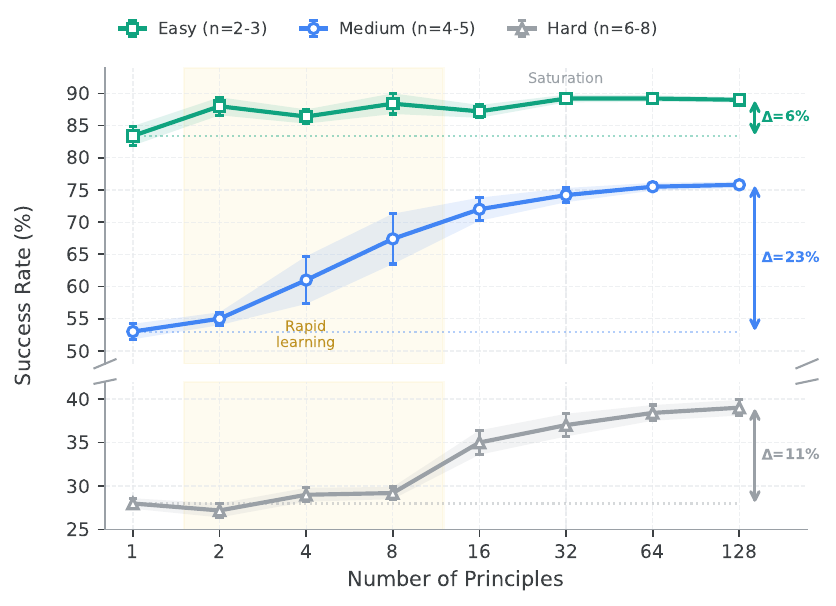}
    \caption{%
        \textbf{Principle scaling across difficulty levels.}
        Medium tasks (blue) show rapid learning between 2 and 8 principles before stabilizing.
        Easy tasks (green) saturate quickly with moderate improvement.
        Hard tasks (gray) show substantial improvement potentially through
        accumulated failure cases. Shaded region marks the ``rapid learning'' phase.
    }
    \label{\figscaling}
    \vspace{-0.5em}
\end{figure}

How does principle count affect success rate? We run 500 episodes per
difficulty using Gemini-3-Flash and measure performance as principles
accumulate (1$\to$128). \figref{\figscaling} reveals distinct patterns.

Medium tasks show the most pronounced scaling: performance rises from
55\%$\to$67\% between 2 and 8 principles, then stabilizes (76\% at 64+).
Easy tasks saturate quickly (83\%$\to$89\%). Hard tasks improve by $+11\%$ even though many remain unsolved; $68\%$ of the principles learned at hard difficulty are avoidance constraints (versus $41\%$ on medium), so the planner gets better largely by ruling moves out, not by solving the task end-to-end.


\subsection{Ablations}
\label{sec:exp:ablation}

Which components of \method are essential, and how does their importance
vary with task difficulty? We ablate key modules across all difficulty
levels (100 episodes each, Gemini-3-Flash), and report the results in \tabref{\tabablation}.

\begin{table}[h]
    \centering
    \small
    \caption{%
        \textbf{Ablation study across difficulty levels.}
        Success rates and token consumption (on medium) for brick insertion.
        Direct retrieval validates the need for principle abstraction.
        Component importance scales with task complexity.
    }
    \label{\tabablation}
    \setlength{\tabcolsep}{4pt} 
    \renewcommand{\arraystretch}{1.2} 

    \begin{tabularx}{\columnwidth}{l | Y Y Y | Y}
        \toprule
        & \multicolumn{3}{c|}{Success Rate} & \\
        \cmidrule(lr){2-4}
        Configuration & Easy & Med. & Hard & Tokens \\
        \midrule
        \method (Full)
        & \cellcolor{cellgreen2}89\%
        & \cellcolor{cellgreen2}76\%
        & \cellcolor{cellgreen3}\textbf{39\%}
        & \cellcolor{cellgreen2}1.0$\times$ \\
        \midrule
        Direct Retrieval
        & \cellcolor{cellred2}48\%
        & \cellcolor{cellred4}23\%
        & \cellcolor{cellred4}8\%
        & \cellcolor{cellgreen3}\textbf{0.55$\times$} \\
        \midrule
        w/o Resonance
        & \cellcolor{cellred1}81\%
        & \cellcolor{cellred3}58\%
        & \cellcolor{cellred4}21\%
        & \cellcolor{cellred2}1.3$\times$ \\

        w/o Verification
        & \cellcolor{cellgreen1}85\%
        & \cellcolor{cellred2}64\%
        & \cellcolor{cellred2}27\%
        & \cellcolor{cellgreen1}0.85$\times$ \\

        w/o Forgetting
        & \cellcolor{cellgreen3}\textbf{91\%}
        & \cellcolor{cellgreen3}\textbf{78\%}
        & \cellcolor{cellgreen1}36\%
        & \cellcolor{cellred4}3.4$\times$ \\

        w/o Working Mem.
        & \cellcolor{cellred1}84\%
        & \cellcolor{cellred1}69\%
        & \cellcolor{cellred3}28\%
        & \cellcolor{cellgreen1}0.75$\times$ \\
        \bottomrule
    \end{tabularx}
    \vspace{-1em}
\end{table}

Principle abstraction is the load-bearing component. Direct retrieval degrades from $48\%$ on easy to $23\%$ on medium and $8\%$ on hard, since state matching becomes fragile with complex dependency graphs while principled abstraction holds up.

Component importance scales with difficulty. Resonance filtering prevents conflicting principles from degrading decisions (-8\% on easy, -18\% on hard). Verification ensures hypothesis quality before promotion (-4\%$\to$-12\%). Working memory enables hypothesis exploration, with increasing importance on harder tasks (-5\%$\to$-11\%).

Forgetting trades accuracy against efficiency. Keeping all experiences gives a small bump on easy and medium ($+2\%$) but hurts performance on hard ($-3\%$) because accumulated noise interferes with decisions, and at $3.4\times$ the token cost the trade-off favors forgetting. Full ablations appear in \appendref{app:ablations}.

\section{Discussion \& Limitation}
\label{sec:discussion}

\paragraph{\textbf{Observation Space}}
Physical learning depends on observation modalities that capture properties vision alone cannot. Our experiments use visual observations and outcome feedback, but the physical world offers richer signals (tactile, force, and audio) that could accelerate principle discovery. Recent work shows that VLMs
contain latent physical knowledge that tactile grounding can
activate~\citep{huang2025tactilevla}, and audio sensing reveals material properties
inaccessible to vision~\citep{liu2024sonicsense}. Active perception
strategies~\citep{sripada2024apvlm} could further enable targeted exploration.
\method's architecture naturally extends to such inputs: principles could emerge
from tactile failure patterns or acoustic contact signatures.

\paragraph{\textbf{Reasoning Representation}}
Text-based principles excel at discrete rules but struggle with continuous dynamics like trajectories or
force profiles. Language cannot fully capture the continuous nature of physical
interaction~\citep{balazadeh2025physics, tian2025uniclothdiff}. Visual chain-of-thought
reasoning~\citep{zhao2025cot} achieves 17\% improvements over text-based
approaches by predicting future frames, and reasoning through continuous visual
tokens~\citep{qin2025chain} or latent world models~\citep{bi2025motus} suggests
future systems may bypass language entirely for physics prediction. Our
verification loop could operate on visual state predictions rather than text:
world models that imagine outcomes~\citep{hafner2023dreamerv3,bruce2024genie, bruce2025genie}
naturally complement the scientific loop by providing richer hypothesis testing.

\paragraph{\textbf{Why Abstraction Beats Retrieval}}
The direct-experience-retrieval ablation makes the point cleanly: raw episodic replay fails because situations never repeat exactly, which lines up with classical findings in episodic memory research~\citep{blundell2016episodic,pritzel2017neural}.
Principles work because they abstract over specific instances.
The transfer experiments show the flip side: abstraction has limits, and irrelevant principles hurt performance once the world drifts. Where retrieval-augmented approaches~\citep{lewis2020retrieval,gao2023rag} apply experience without checking whether it still holds, \method runs that check, and the ability to revise beliefs is what makes the loop a learning system rather than a memorizing one.

\paragraph{\textbf{Limitations}}
Our work focuses on high-level planning rather than low-level control. Integrating learned principles into end-to-end VLA execution is still open~\citep{pi2025pi06,gemini2025robotics}, and current VLAs struggle when task descriptions include novel physical constraints; closing that gap is a natural next step.
For continuous real-world deployment, environment reset is a practical bottleneck: when objects fall off tables or parts break, the system needs human intervention, and autonomous recovery and repair would be required for truly lifelong learning~\citep{liu2021lifelong}.


\section{Conclusion}
\label{sec:conclusion}

We presented \method, a test-time memory framework that lets a VLM robot planner learn physical principles through interaction. The scientific loop of hypothesis generation, verification, and promotion produces knowledge that is both effective and interpretable. Across three real-world tasks and simulation benchmarks, principled abstraction outperforms raw retrieval, and the ability to revise beliefs is what makes the system robust under shifting physical conditions. A natural next step is to let the same verification step run against world-model rollouts, so that text principles and learned dynamics can share one loop.


\bibliographystyle{plainnat}
\bibliography{references}

@inproceedings{ahn2022saycan,
  title     = {Do As I Can, Not As I Say: Grounding Language in Robotic Affordances},
  author    = {Ahn, Michael and Brohan, Anthony and Brown, Noah and Chebotar, Yevgen and Cortes, Omar and David, Byron and Finn, Chelsea and Fu, Chuyuan and Gober, Keerthana and Gopalakrishnan, Karol and others},
  booktitle = {Proceedings of the Conference on Robot Learning (CoRL)},
  pages     = {287--318},
  year      = {2022}
}

@article{driess2023palme,
  title   = {{PaLM-E}: An Embodied Multimodal Language Model},
  author = {Driess, Danny and Xia, Fei and Sajjadi, Mehdi SM and Lynch, Corey and Chowdhery, Aakanksha and Wahid, Ayzaan and Tompson, Jonathan and Vuong, Quan and Yu, Tianhe and Huang, Wenlong and others},
  journal = {arXiv preprint arXiv:2303.03378},
  year    = {2023}
}

@article{brohan2022rt1,
  title={Rt-1: Robotics transformer for real-world control at scale},
  author={Brohan, Anthony and Brown, Noah and Carbajal, Justice and Chebotar, Yevgen and Dabis, Joseph and Finn, Chelsea and Gopalakrishnan, Keerthana and Hausman, Karol and Herzog, Alex and Hsu, Jasmine and others},
  journal={arXiv preprint arXiv:2212.06817},
  year={2022}
}

@article{brohan2023rt2,
  title   = {{RT-2}: Vision-Language-Action Models Transfer Web Knowledge to Robotic Control},
  author  = {Brohan, Anthony and Brown, Noah and Carbajal, Justice and Chebotar, Yevgen and Chen, Xi and Choromanski, Krzysztof and Ding, Tianli and Driess, Danny and Dubey, Avinava and Finn, Chelsea and others},
  journal = {arXiv preprint arXiv:2307.15818},
  year    = {2023}
}

@inproceedings{huang2022inner,
  title     = {Inner Monologue: Embodied Reasoning through Planning with Language Models},
  author    = {Huang, Wenlong and Xia, Fei and Xiao, Ted and Chan, Harris and Liang, Jacky and Florence, Pete and Zeng, Andy and Tompson, Jonathan and Mordatch, Igor and Chebotar, Yevgen and others},
  booktitle = {Proceedings of the Conference on Robot Learning (CoRL)},
  pages     = {1769--1782},
  year      = {2022}
}

@inproceedings{liang2023code,
  title     = {Code as Policies: Language Model Programs for Embodied Control},
  author    = {Liang, Jacky and Huang, Wenlong and Xia, Fei and Xu, Peng and Hausman, Karol and Ichter, Brian and Florence, Pete and Zeng, Andy},
  booktitle = {IEEE International Conference on Robotics and Automation (ICRA)},
  pages     = {9493--9500},
  year      = {2023}
}

@misc{open_x_embodiment2023,
title={Open {X-E}mbodiment: Robotic Learning Datasets and {RT-X} Models},
author = {Open X-Embodiment Collaboration and Abby O'Neill and Abdul Rehman and Abhinav Gupta and Abhiram Maddukuri and Abhishek Gupta and Abhishek Padalkar and Abraham Lee and Acorn Pooley and Agrim Gupta and Ajay Mandlekar and Ajinkya Jain and Albert Tung and Alex Bewley and Alex Herzog and Alex Irpan and Alexander Khazatsky and Anant Rai and Anchit Gupta and Andrew Wang and Andrey Kolobov and Anikait Singh and Animesh Garg and Aniruddha Kembhavi and Annie Xie and Anthony Brohan and Antonin Raffin and Archit Sharma and Arefeh Yavary and Arhan Jain and Ashwin Balakrishna and Ayzaan Wahid and Ben Burgess-Limerick and Beomjoon Kim and Bernhard Schölkopf and Blake Wulfe and Brian Ichter and Cewu Lu and Charles Xu and Charlotte Le and Chelsea Finn and Chen Wang and Chenfeng Xu and Cheng Chi and Chenguang Huang and Christine Chan and Christopher Agia and Chuer Pan and Chuyuan Fu and Coline Devin and Danfei Xu and Daniel Morton and Danny Driess and Daphne Chen and Deepak Pathak and Dhruv Shah and Dieter Büchler and Dinesh Jayaraman and Dmitry Kalashnikov and Dorsa Sadigh and Edward Johns and Ethan Foster and Fangchen Liu and Federico Ceola and Fei Xia and Feiyu Zhao and Felipe Vieira Frujeri and Freek Stulp and Gaoyue Zhou and Gaurav S. Sukhatme and Gautam Salhotra and Ge Yan and Gilbert Feng and Giulio Schiavi and Glen Berseth and Gregory Kahn and Guangwen Yang and Guanzhi Wang and Hao Su and Hao-Shu Fang and Haochen Shi and Henghui Bao and Heni Ben Amor and Henrik I Christensen and Hiroki Furuta and Homanga Bharadhwaj and Homer Walke and Hongjie Fang and Huy Ha and Igor Mordatch and Ilija Radosavovic and Isabel Leal and Jacky Liang and Jad Abou-Chakra and Jaehyung Kim and Jaimyn Drake and Jan Peters and Jan Schneider and Jasmine Hsu and Jay Vakil and Jeannette Bohg and Jeffrey Bingham and Jeffrey Wu and Jensen Gao and Jiaheng Hu and Jiajun Wu and Jialin Wu and Jiankai Sun and Jianlan Luo and Jiayuan Gu and Jie Tan and Jihoon Oh and Jimmy Wu and Jingpei Lu and Jingyun Yang and Jitendra Malik and João Silvério and Joey Hejna and Jonathan Booher and Jonathan Tompson and Jonathan Yang and Jordi Salvador and Joseph J. Lim and Junhyek Han and Kaiyuan Wang and Kanishka Rao and Karl Pertsch and Karol Hausman and Keegan Go and Keerthana Gopalakrishnan and Ken Goldberg and Kendra Byrne and Kenneth Oslund and Kento Kawaharazuka and Kevin Black and Kevin Lin and Kevin Zhang and Kiana Ehsani and Kiran Lekkala and Kirsty Ellis and Krishan Rana and Krishnan Srinivasan and Kuan Fang and Kunal Pratap Singh and Kuo-Hao Zeng and Kyle Hatch and Kyle Hsu and Laurent Itti and Lawrence Yunliang Chen and Lerrel Pinto and Li Fei-Fei and Liam Tan and Linxi "Jim" Fan and Lionel Ott and Lisa Lee and Luca Weihs and Magnum Chen and Marion Lepert and Marius Memmel and Masayoshi Tomizuka and Masha Itkina and Mateo Guaman Castro and Max Spero and Maximilian Du and Michael Ahn and Michael C. Yip and Mingtong Zhang and Mingyu Ding and Minho Heo and Mohan Kumar Srirama and Mohit Sharma and Moo Jin Kim and Muhammad Zubair Irshad and Naoaki Kanazawa and Nicklas Hansen and Nicolas Heess and Nikhil J Joshi and Niko Suenderhauf and Ning Liu and Norman Di Palo and Nur Muhammad Mahi Shafiullah and Oier Mees and Oliver Kroemer and Osbert Bastani and Pannag R Sanketi and Patrick "Tree" Miller and Patrick Yin and Paul Wohlhart and Peng Xu and Peter David Fagan and Peter Mitrano and Pierre Sermanet and Pieter Abbeel and Priya Sundaresan and Qiuyu Chen and Quan Vuong and Rafael Rafailov and Ran Tian and Ria Doshi and Roberto Mart{'i}n-Mart{'i}n and Rohan Baijal and Rosario Scalise and Rose Hendrix and Roy Lin and Runjia Qian and Ruohan Zhang and Russell Mendonca and Rutav Shah and Ryan Hoque and Ryan Julian and Samuel Bustamante and Sean Kirmani and Sergey Levine and Shan Lin and Sherry Moore and Shikhar Bahl and Shivin Dass and Shubham Sonawani and Shubham Tulsiani and Shuran Song and Sichun Xu and Siddhant Haldar and Siddharth Karamcheti and Simeon Adebola and Simon Guist and Soroush Nasiriany and Stefan Schaal and Stefan Welker and Stephen Tian and Subramanian Ramamoorthy and Sudeep Dasari and Suneel Belkhale and Sungjae Park and Suraj Nair and Suvir Mirchandani and Takayuki Osa and Tanmay Gupta and Tatsuya Harada and Tatsuya Matsushima and Ted Xiao and Thomas Kollar and Tianhe Yu and Tianli Ding and Todor Davchev and Tony Z. Zhao and Travis Armstrong and Trevor Darrell and Trinity Chung and Vidhi Jain and Vikash Kumar and Vincent Vanhoucke and Vitor Guizilini and Wei Zhan and Wenxuan Zhou and Wolfram Burgard and Xi Chen and Xiangyu Chen and Xiaolong Wang and Xinghao Zhu and Xinyang Geng and Xiyuan Liu and Xu Liangwei and Xuanlin Li and Yansong Pang and Yao Lu and Yecheng Jason Ma and Yejin Kim and Yevgen Chebotar and Yifan Zhou and Yifeng Zhu and Yilin Wu and Ying Xu and Yixuan Wang and Yonatan Bisk and Yongqiang Dou and Yoonyoung Cho and Youngwoon Lee and Yuchen Cui and Yue Cao and Yueh-Hua Wu and Yujin Tang and Yuke Zhu and Yunchu Zhang and Yunfan Jiang and Yunshuang Li and Yunzhu Li and Yusuke Iwasawa and Yutaka Matsuo and Zehan Ma and Zhuo Xu and Zichen Jeff Cui and Zichen Zhang and Zipeng Fu and Zipeng Lin},
howpublished  = {\url{https://arxiv.org/abs/2310.08864}},
year = {2023},
}

@inproceedings{team2024octo,
  title     = {Octo: An Open-Source Generalist Robot Policy},
  author    = {{Octo Model Team} and Ghosh, Dibya and Walke, Homer and Pertsch, Karl and Black, Kevin and Mees, Oier and Dasari, Sudeep and Hejna, Joey and Kreber, Tobias and Finn, Chelsea and Levine, Sergey},
  booktitle = {Proceedings of Robotics: Science and Systems (RSS)},
  year      = {2024}
}

@inproceedings{zawalski2024ecot,
  title     = {Robotic Control via Embodied Chain-of-Thought Reasoning},
  author    = {Zawalski, Micha{\l} and Chen, William and Pertsch, Karl and Mees, Oier and Finn, Chelsea and Levine, Sergey},
  booktitle = {Proceedings of the Conference on Robot Learning (CoRL)},
  year      = {2024}
}

@article{gemini2025robotics,
  title   = {Gemini Robotics 1.5: Pushing the Frontier of Generalist Robots with Advanced Embodied Reasoning, Thinking, and Motion Transfer},
  author  = {{Gemini Robotics Team} and Abdolmaleki, Abbas and Brohan, Anthony and Brown, Noah and Bousmalis, Konstantinos and Finn, Chelsea and Hausman, Karol and Levine, Sergey and others},
  journal = {arXiv preprint arXiv:2510.03342},
  year    = {2025}
}

@article{helix2025figure,
  title   = {Helix: A Vision-Language-Action Model for Generalist Humanoid Control},
  author  = {{Figure AI Team}},
  journal = {Technical Report},
  year    = {2025}
}

@article{zhang2026vlm4vla,
  title={VLM4VLA: Revisiting Vision-Language-Models in Vision-Language-Action Models},
  author={Zhang, Jianke and Chen, Xiaoyu and Wang, Qiuyue and Li, Mingsheng and Guo, Yanjiang and Hu, Yucheng and Zhang, Jiajun and Bai, Shuai and Lin, Junyang and Chen, Jianyu},
  journal={arXiv preprint arXiv:2601.03309},
  year={2026}
}

@article{li2024robovlms,
  title   = {Towards Generalist Robot Policies: What Matters in Building Vision-Language-Action Models},
  author  = {Li, Xinghang and Li, Peiyan and Liu, Minghuan and Wang, Dong and Liu, Jirong and Kang, Bingyi and Ma, Xiao and Kong, Tao and Zhang, Hanbo and Liu, Huaping},
  journal = {arXiv preprint arXiv:2412.14058},
  year    = {2024}
}

@inproceedings{kim2024openvla,
  title     = {{OpenVLA}: An Open-Source Vision-Language-Action Model},
  author    = {Kim, Moo Jin and Pertsch, Karl and Karamcheti, Siddharth and Xiao, Ted and Balakrishna, Ashwin and Nair, Suraj and Finn, Chelsea and Levine, Sergey and Liang, Percy},
  booktitle = {Proceedings of the Conference on Robot Learning (CoRL)},
  year      = {2024}
}

@article{black2024pi0,
  title={$\pi$0: A vision-language-action flow model for general robot control. CoRR, abs/2410.24164, 2024. doi: 10.48550},
  author={Black, Kevin and Brown, Noah and Driess, Danny and Esmail, Adnan and Equi, Michael and Finn, Chelsea and Fusai, Niccolo and Groom, Lachy and Hausman, Karol and Ichter, Brian and others},
  journal={arXiv preprint ARXIV.2410.24164}
}

@article{sutton1999options,
  title   = {Between {MDPs} and Semi-{MDPs}: A Framework for Temporal Abstraction in Reinforcement Learning},
  author  = {Sutton, Richard S. and Precup, Doina and Singh, Satinder},
  journal = {Artificial Intelligence},
  volume  = {112},
  number  = {1-2},
  pages   = {181--211},
  year    = {1999}
}

@inproceedings{finn2017maml,
  title     = {Model-Agnostic Meta-Learning for Fast Adaptation of Deep Networks},
  author    = {Finn, Chelsea and Abbeel, Pieter and Levine, Sergey},
  booktitle = {International Conference on Machine Learning (ICML)},
  pages     = {1126--1135},
  year      = {2017}
}

@inproceedings{finn2017oneshot,
  title     = {One-Shot Visual Imitation Learning via Meta-Learning},
  author    = {Finn, Chelsea and Yu, Tianhe and Zhang, Tianhao and Abbeel, Pieter and Levine, Sergey},
  booktitle = {Proceedings of the Conference on Robot Learning (CoRL)},
  pages     = {357--368},
  year      = {2017}
}

@inproceedings{tobin2017domain,
  title     = {Domain Randomization for Transferring Deep Neural Networks from Simulation to the Real World},
  author    = {Tobin, Josh and Fong, Rachel and Ray, Alex and Schneider, Jonas and Zaremba, Wojciech and Abbeel, Pieter},
  booktitle = {IEEE/RSJ International Conference on Intelligent Robots and Systems (IROS)},
  pages     = {23--30},
  year      = {2017}
}

@article{luo2024rlvla,
  title   = {Improving Vision-Language-Action Model with Online Reinforcement Learning},
  author  = {Luo, Yanjiang and Wang, Zhecheng and Zhang, Xiaoyu and Xu, Zhixuan and Lu, Zhengrong and Qu, Yanjie and Xu, Huazhe},
  journal = {arXiv preprint arXiv:2501.01734},
  year    = {2025}
}

@inproceedings{laskin2022incontext,
  title     = {In-Context Reinforcement Learning with Algorithm Distillation},
  author    = {Laskin, Michael and Wang, Luyu and Oh, Junhyuk and Parisotto, Emilio and Spencer, Stephen and Steiber, Richie and Strouse, DJ and Hansen, Steven and Fiez, Angelos and Simchowitz, Max and others},
  booktitle = {International Conference on Learning Representations (ICLR)},
  year      = {2023}
}

@article{pi2025pi06,
  title   = {$\pi^{*}_{0.6}$: A {VLA} That Learns From Experience},
  author  = {{Physical Intelligence Team} and Black, Kevin and Brown, Noah and Finn, Chelsea and Hausman, Karol and Ichter, Brian and Levine, Sergey and Pertsch, Karl and Shi, Lucy Xiaoyang and others},
  journal = {arXiv preprint arXiv:2511.14759},
  year    = {2025}
}

@inproceedings{shi2024yell,
  title     = {Yell At Your Robot: Improving On-the-Fly from Language Corrections},
  author    = {Shi, Lucy Xiaoyang and Hu, Zheyuan and Zhao, Tony Z. and Sharma, Archit and Pertsch, Karl and Luo, Jianlan and Levine, Sergey and Finn, Chelsea},
  booktitle = {Proceedings of Robotics: Science and Systems (RSS)},
  year      = {2024}
}

@inproceedings{chen2021decision,
  title     = {Decision Transformer: Reinforcement Learning via Sequence Modeling},
  author    = {Chen, Lili and Lu, Kevin and Rajeswaran, Aravind and Lee, Kimin and Grover, Aditya and Laskin, Michael and Abbeel, Pieter and Srinivas, Aravind and Mordatch, Igor},
  booktitle = {Advances in Neural Information Processing Systems (NeurIPS)},
  volume    = {34},
  year      = {2021}
}

@article{andrychowicz2020dactyl,
  title   = {Learning Dexterous In-Hand Manipulation},
  author  = {Andrychowicz, Marcin and Baker, Bowen and Chociej, Maciek and J{\'o}zefowicz, Rafa{\l} and McGrew, Bob and Pachocki, Jakub and Petron, Arthur and Plappert, Matthias and Powell, Glenn and Ray, Alex and others},
  journal = {International Journal of Robotics Research},
  volume  = {39},
  number  = {1},
  pages   = {3--20},
  year    = {2020}
}

@inproceedings{sun2020ttt,
  title={Test-time training with self-supervision for generalization under distribution shifts},
  author={Sun, Yu and Wang, Xiaolong and Liu, Zhuang and Miller, John and Efros, Alexei and Hardt, Moritz},
  booktitle = {International Conference on Machine Learning (ICML)},
  pages={9229--9248},
  year={2020},
  organization={PMLR}
}

@article{blundell2016episodic,
  title={Model-free episodic control},
  author={Blundell, Charles and Uria, Benigno and Pritzel, Alexander and Li, Yazhe and Ruderman, Avraham and Leibo, Joel Z and Rae, Jack and Wierstra, Daan and Hassabis, Demis},
  journal={arXiv preprint arXiv:1606.04460},
  year={2016}
}

@inproceedings{pritzel2017neural,
  title     = {Neural Episodic Control},
  author    = {Pritzel, Alexander and Uria, Benigno and Srinivasan, Sriram and Puigdom{\`e}nech, Adri{\`a} and Vinyals, Oriol and Hassabis, Demis and Wierstra, Daan and Blundell, Charles},
  booktitle = {International Conference on Machine Learning (ICML)},
  pages     = {2827--2836},
  year      = {2017}
}

@article{shinn2023reflexion,
  title   = {Reflexion: Language Agents with Verbal Reinforcement Learning},
  author  = {Shinn, Noah and Cassano, Federico and Gopinath, Ashwin and Narasimhan, Karthik and Yao, Shunyu},
  journal = {Advances in Neural Information Processing Systems (NeurIPS)},
  volume  = {36},
  year    = {2023}
}

@article{madaan2023selfrefine,
  title   = {Self-Refine: Iterative Refinement with Self-Feedback},
  author  = {Madaan, Aman and Tandon, Niket and Gupta, Prakhar and Hallinan, Skyler and Gao, Luyu and Wiegreffe, Sarah and Alon, Uri and Dziri, Nouha and Prabhumoye, Shrimai and Yang, Yiming and others},
  journal = {Advances in Neural Information Processing Systems (NeurIPS)},
  volume  = {36},
  year    = {2023}
}

@article{gao2023rag,
  title   = {Retrieval-Augmented Generation for Large Language Models: A Survey},
  author  = {Gao, Yunfan and Xiong, Yun and Gao, Xinyu and Jia, Kangxiang and Pan, Jinliu and Bi, Yuxi and Dai, Yi and Sun, Jiawei and Wang, Meng and Wang, Haofen},
  journal = {arXiv preprint arXiv:2312.10997},
  year    = {2023}
}

@inproceedings{lewis2020retrieval,
  title     = {Retrieval-Augmented Generation for Knowledge-Intensive {NLP} Tasks},
  author    = {Lewis, Patrick and Perez, Ethan and Piktus, Aleksandra and Petroni, Fabio and Karpukhin, Vladimir and Goyal, Naman and K{\"u}ttler, Heinrich and Lewis, Mike and Yih, Wen-tau and Rockt{\"a}schel, Tim and others},
  booktitle = {Advances in Neural Information Processing Systems (NeurIPS)},
  volume    = {33},
  pages     = {9459--9474},
  year      = {2020}
}

@inproceedings{fang2019scene,
  title     = {Scene Memory Transformer for Embodied Agents in Long-Horizon Tasks},
  author    = {Fang, Kuan and Toshev, Alexander and Fei-Fei, Li and Savarese, Silvio},
  booktitle = {IEEE Conference on Computer Vision and Pattern Recognition (CVPR)},
  pages     = {538--547},
  year      = {2019}
}

@inproceedings{li2024memonav,
  title     = {{MemoNav}: Working Memory Model for Visual Navigation},
  author    = {Li, Hongxin and Wang, Zeyu and Yang, Xu and Yang, Yuran and Mei, Shuqi and Zhang, Zhaoxiang},
  booktitle = {IEEE Conference on Computer Vision and Pattern Recognition (CVPR)},
  pages     = {17913--17922},
  year      = {2024}
}

@article{sridhar2025memer,
  title={MemER: Scaling Up Memory for Robot Control via Experience Retrieval},
  author={Sridhar, Ajay and Pan, Jennifer and Sharma, Satvik and Finn, Chelsea},
  journal={arXiv preprint arXiv:2510.20328},
  year={2025}
}

@article{shi2025memoryvla,
  title={Memoryvla: Perceptual-cognitive memory in vision-language-action models for robotic manipulation},
  author={Shi, Hao and Xie, Bin and Liu, Yingfei and Sun, Lin and Liu, Fengrong and Wang, Tiancai and Zhou, Erjin and Fan, Haoqiang and Zhang, Xiangyu and Huang, Gao},
  journal={arXiv preprint arXiv:2508.19236},
  year={2025}
}

@article{fang2025sam2act,
  title   = {{SAM2Act}: Integrating Visual Foundation Model with A Memory Architecture for Robotic Manipulation},
  author  = {Fang, Haoquan and Grotz, Markus and Pumacay, Wilbert and Wang, Yi Ru and Fox, Dieter and Krishna, Ranjay and Duan, Jiafei},
  journal = {arXiv preprint arXiv:2501.18564},
  year    = {2025}
}

@article{hafner2023dreamerv3,
  title   = {Mastering Diverse Domains through World Models},
  author  = {Hafner, Danijar and Pasukonis, Jurgis and Ba, Jimmy and Lillicrap, Timothy},
  journal = {arXiv preprint arXiv:2301.04104},
  year    = {2023}
}

@article{bruce2025genie,
  title  = {Genie 3: A New Frontier for World Models},
  author = {Philip J. Ball and Jakob Bauer and Frank Belletti and Bethanie Brownfield and Ariel Ephrat and Shlomi Fruchter and Agrim Gupta and Kristian Holsheimer and Aleksander Holynski and Jiri Hron and Christos Kaplanis and Marjorie Limont and Matt McGill and Yanko Oliveira and Jack Parker-Holder and Frank Perbet and Guy Scully and Jeremy Shar and Stephen Spencer and Omer Tov and Ruben Villegas and Emma Wang and Jessica Yung and Cip Baetu and Jordi Berbel and David Bridson and Jake Bruce and Gavin Buttimore and Sarah Chakera and Bilva Chandra and Paul Collins and Alex Cullum and Bogdan Damoc and Vibha Dasagi and Maxime Gazeau and Charles Gbadamosi and Woohyun Han and Ed Hirst and Ashyana Kachra and Lucie Kerley and Kristian Kjems and Eva Knoepfel and Vika Koriakin and Jessica Lo and Cong Lu and Zeb Mehring and Alex Moufarek and Henna Nandwani and Valeria Oliveira and Fabio Pardo and Jane Park and Andrew Pierson and Ben Poole and Helen Ran and Tim Salimans and Manuel Sanchez and Igor Saprykin and Amy Shen and Sailesh Sidhwani and Duncan Smith and Joe Stanton and Hamish Tomlinson and Dimple Vijaykumar and Luyu Wang and Piers Wingfield and Nat Wong and Keyang Xu and Christopher Yew and Nick Young and Vadim Zubov and Douglas Eck and Dumitru Erhan and Koray Kavukcuoglu and Demis Hassabis and Zoubin Gharamani and Raia Hadsell and A{\"a}ron van den Oord and Inbar Mosseri and Adrian Bolton and Satinder Singh and Tim Rockt{\"a}schel},
  year   = {2025},
  url    = {}
}

@inproceedings{bruce2024genie,
  title={Genie: Generative interactive environments},
  author={Bruce, Jake and Dennis, Michael D and Edwards, Ashley and Parker-Holder, Jack and Shi, Yuge and Hughes, Edward and Lai, Matthew and Mavalankar, Aditi and Steigerwald, Richie and Apps, Chris and others},
  booktitle={Forty-first International Conference on Machine Learning},
  year={2024}
}

@article{liu2021lifelong,
  title   = {A Lifelong Learning Approach to Mobile Robot Navigation},
  author  = {Liu, Bo and Xiao, Xuesu and Stone, Peter},
  journal = {IEEE Robotics and Automation Letters},
  volume  = {6},
  number  = {2},
  pages   = {1090--1097},
  year    = {2021}
}

@article{meng2025legion,
  title   = {Preserving and Combining Knowledge in Robotic Lifelong Reinforcement Learning},
  author  = {Meng, Yuan and Bing, Zhenshan and Yao, Xiangtong and Chen, Kejia and Huang, Kai and Gao, Yang and Sun, Fuchun and Knoll, Alois},
  journal = {Nature Machine Intelligence},
  year    = {2025}
}

@inproceedings{zhang2024extract,
  title     = {{EXTRACT}: Efficient Policy Learning by Extracting Transferable Robot Skills from Offline Data},
  author    = {Zhang, Jesse and Heo, Minho and Liu, Zuxin and Biyik, Erdem and Lim, Joseph J. and Liu, Yao and Fakoor, Rasool},
  booktitle = {Proceedings of the Conference on Robot Learning (CoRL)},
  year      = {2024}
}

@book{popper1959logic,
  title     = {The Logic of Scientific Discovery},
  author    = {Popper, Karl},
  publisher = {Routledge},
  year      = {1959}
}

@article{kirillov2023segment,
  title={Segment Anything},
  author={Kirillov, Alexander and Mintun, Eric and Ravi, Nikhila and Mao, Hanzi and Rolland, Chloe and Gustafson, Laura and Xiao, Tete and Whitehead, Spencer and Berg, Alexander C. and Lo, Wan-Yen and Doll{\'a}r, Piotr and Girshick, Ross},
  journal={arXiv:2304.02643},
  year={2023}
}

@misc{feng2025reflective,
  title={Reflective Planning: Vision-Language Models for Multi-Stage Long-Horizon Robotic Manipulation}, 
  author={Yunhai Feng and Jiaming Han and Zhuoran Yang and Xiangyu Yue and Sergey Levine and Jianlan Luo},
  year={2025},
  eprint={2502.16707},
  archivePrefix={arXiv},
  primaryClass={cs.RO},
  url={https://arxiv.org/abs/2502.16707}, 
}

@article{tian2025uniclothdiff,
  author    = {Tian, Tongxuan and Li, Haoyang and Ai, Bo and Yuan, Xiaodi and Huang, Zhiao and Su, Hao},
  title     = {Diffusion Dynamics Models with Generative State Estimation for Cloth Manipulation},
  journal   = {Conference on Robot Learning (CoRL)},
  year      = {2025},
}

@article{qin2025chain,
  title={Chain-of-Visual-Thought: Teaching VLMs to See and Think Better with Continuous Visual Tokens},
  author={Qin, Yiming and Wei, Bomin and Ge, Jiaxin and Kallidromitis, Konstantinos and Fu, Stephanie and Darrell, Trevor and Wang, Xudong},
  journal={arXiv preprint arXiv:2511.19418},
  year={2025}
}

@inproceedings{zhao2025cot,
  title={Cot-vla: Visual chain-of-thought reasoning for vision-language-action models},
  author={Zhao, Qingqing and Lu, Yao and Kim, Moo Jin and Fu, Zipeng and Zhang, Zhuoyang and Wu, Yecheng and Li, Zhaoshuo and Ma, Qianli and Han, Song and Finn, Chelsea and others},
  booktitle={Proceedings of the Computer Vision and Pattern Recognition Conference},
  pages={1702--1713},
  year={2025}
}

@article{agarwal2025cosmos,
  title   = {Cosmos World Foundation Model Platform for Physical {AI}},
  author  = {Agarwal, Niket and Ali, Arslan and Bala, Maciej and Balaji, Yogesh and Barker, Erik and Cai, Tiffany and Chattopadhyay, Prithvijit and Chen, Yongxin and Cui, Yin and Ding, Yifan and others},
  journal = {arXiv preprint arXiv:2501.03575},
  year    = {2025}
}

@misc{google2025gemini3flash,
  author = {Google DeepMind},
  title = {Gemini 3 Flash: Frontier Intelligence Built for Speed},
  year = {2025},
  url = {https://blog.google/products-and-platforms/products/gemini/gemini-3-flash/},
  note = {Accessed: 2026-01-21}
}

@article{qwen3vl2025,
  title={Qwen3-VL Technical Report},
  author={Bai, Shuai and Cai, Yuxuan and Chen, Ruizhe and Chen, Keqin and Cheng, Zesen and Deng, Lianghao and Ding, Wei and Gao, Chang and Ge, Chunjiang and Ge, Wenbin and others},
  journal={arXiv preprint arXiv:2511.21631},
  year={2025},
  url={https://arxiv.org/abs/2511.21631}
}

@misc{openai2025gpt5,
  author = {OpenAI},
  title = {{GPT-5.1}: Advanced Multimodal Reasoning Model},
  year = {2025},
  url = {https://openai.com/gpt-5},
}

@misc{google2025geminirobotics,
  author = {Google DeepMind},
  title = {Gemini Embodied Reasoning 1.5: Multi-Step Reasoning for Robotic Planning},
  year = {2025},
  url = {https://deepmind.google/technologies/gemini/},
  note = {Accessed: 2026-01-21}
}

@article{team2025sima2,
  title   = {{SIMA} 2: A Generalist Embodied Agent for Virtual Worlds},
  author  = {{SIMA Team} and Bolton, Adrian and Lerchner, Alexander and others},
  journal = {arXiv preprint arXiv:2512.04797},
  year    = {2025}
}

@article{bjorck2025groot,
  title={{GR00T N1}: An open foundation model for generalist humanoid robots},
  author={Nvidia, J Bjorck and Castaneda, Fernando and Cherniadev, N and Da, X and Ding, R and Fan, L and Fang, Y and Fox, D and Hu, F and Huang, S and others},
  journal={arXiv preprint arXiv:2503.14734},
  year={2025}
}

@article{kawaharazuka2025vla,
  title   = {Vision-Language-Action Models for Robotics: A Review Towards Real-World Applications},
  author  = {Kawaharazuka, Kento and Oh, Jihoon and Yamada, Jun and Posner, Ingmar and Zhu, Yuke},
  journal = {IEEE Access},
  volume  = {13},
  pages   = {162467--162504},
  year    = {2025}
}

@article{li2025mapvla,
  title   = {{MAP-VLA}: Memory-Augmented Prompting for Vision-Language-Action Model in Robotic Manipulation},
  author  = {Li, Runhao and Guo, Wenkai and Wu, Zhenyu and Xu, Huazhe and others},
  journal = {arXiv preprint arXiv:2511.09516},
  year    = {2025}
}

@inproceedings{zhao2025cotvla,
  title     = {{CoT-VLA}: Visual Chain-of-Thought Reasoning for Vision-Language-Action Models},
  author    = {Zhao, Qingqing and Lu, Yao and Kim, Moo Jin and Fu, Zipeng and Zhang, Zhuoyang and Wu, Yecheng and Li, Zhaoshuo and Ma, Qianli and Han, Song and Finn, Chelsea and others},
  booktitle = {IEEE/CVF Conference on Computer Vision and Pattern Recognition (CVPR)},
  year      = {2025}
}

@inproceedings{kim2025feedtta,
  title     = {Test-Time Adaptation for Online Vision-Language Navigation with Feedback-based Reinforcement Learning},
  author    = {Kim, Sungjune and Oh, Gyeongrok and Ko, Heeju and Ji, Daehyun and Lee, Dongwook and Lee, Byung-Jun and Jang, Sujin and Kim, Sangpil},
  booktitle = {International Conference on Machine Learning (ICML)},
  year      = {2025}
}

@inproceedings{xu2025tarl,
  title={Test-time Adapted Reinforcement Learning with Action Entropy Regularization},
  author={Xu, Shoukai and Tan, Mingkui and Liu, Liu and Zhang, Zhong and Zhao, Peilin and others},
  booktitle={Forty-second International Conference on Machine Learning}
}

@inproceedings{yoo2025wormi,
  title     = {World Model Implanting for Test-time Adaptation of Embodied Agents},
  author    = {Yoo, Minjong and Jang, Jinwoo and Yoon, Sihyung and Woo, Honguk},
  booktitle = {International Conference on Machine Learning (ICML)},
  year      = {2025}
}

@article{huang2025tactilevla,
  title   = {Tactile-{VLA}: Unlocking Vision-Language-Action Model's Physical Knowledge for Tactile Generalization},
  author  = {Huang, Jialei and Wang, Shuo and Lin, Fanqi and Hu, Yihang and Wen, Chuan and Gao, Yang},
  journal = {arXiv preprint arXiv:2507.09160},
  year    = {2025}
}

@inproceedings{liu2024sonicsense,
  title     = {{SonicSense}: Object Perception from In-Hand Acoustic Vibration},
  author    = {Liu, Jiaxun and Chen, Boyuan},
  booktitle = {Proceedings of the Conference on Robot Learning (CoRL)},
  year      = {2024}
}

@article{sripada2024apvlm,
  title   = {{AP-VLM}: Active Perception Enabled by Vision-Language Models},
  author  = {Sripada, Venkatesh and Carter, Samuel and Guerin, Frank and Ghalamzan, Amir},
  journal = {arXiv preprint arXiv:2409.17641},
  year    = {2024}
}

@article{bi2025motus,
  title   = {Motus: A Unified Latent Action World Model},
  author  = {Bi, Hongzhe and Tan, Hengkai and Xie, Shenghao and Wang, Zeyuan and Huang, Shuhe and Liu, Haitian and Zhao, Ruowen and Feng, Yao and Xiang, Chendong and Rong, Yinze and Zhao, Hongyan and Liu, Hanyu and Su, Zhizhong and Ma, Lei and Su, Hang and Zhu, Jun},
  journal = {arXiv preprint arXiv:2512.13030},
  year    = {2025}
}

@inproceedings{balazadeh2025physics,
  title={Physics context builders: A modular framework for physical reasoning in vision-language models},
  author={Balazadeh, Vahid and Ataei, Mohammadmehdi and Cheong, Hyunmin and Khasahmadi, Amir Hosein and Krishnan, Rahul G},
  booktitle={Proceedings of the IEEE/CVF International Conference on Computer Vision},
  pages={7318--7328},
  year={2025}
}

@article{torne2026mem,
  title={MEM: Multi-Scale Embodied Memory for Vision Language Action Models},
  author={Torne, Marcel and Pertsch, Karl and Walke, Homer and Vedder, Kyle and Nair, Suraj and Ichter, Brian and Ren, Allen Z. and Wang, Haohuan and Tang, Jiaming and Stachowicz, Kyle and Dhabalia, Karan and Equi, Michael and Vuong, Quan and Springenberg, Jost Tobias and Levine, Sergey and Finn, Chelsea and Driess, Danny},
  journal={arXiv preprint arXiv:2603.03596},
  year={2026}
}

\clearpage

\appendices
\section{Method Implementation Details}
\label{app:method-details}

This appendix provides complete algorithmic details for \method that were
omitted from the main paper due to space constraints.

\begin{figure*}[b]
    \centering
    \includegraphics[width=\textwidth]{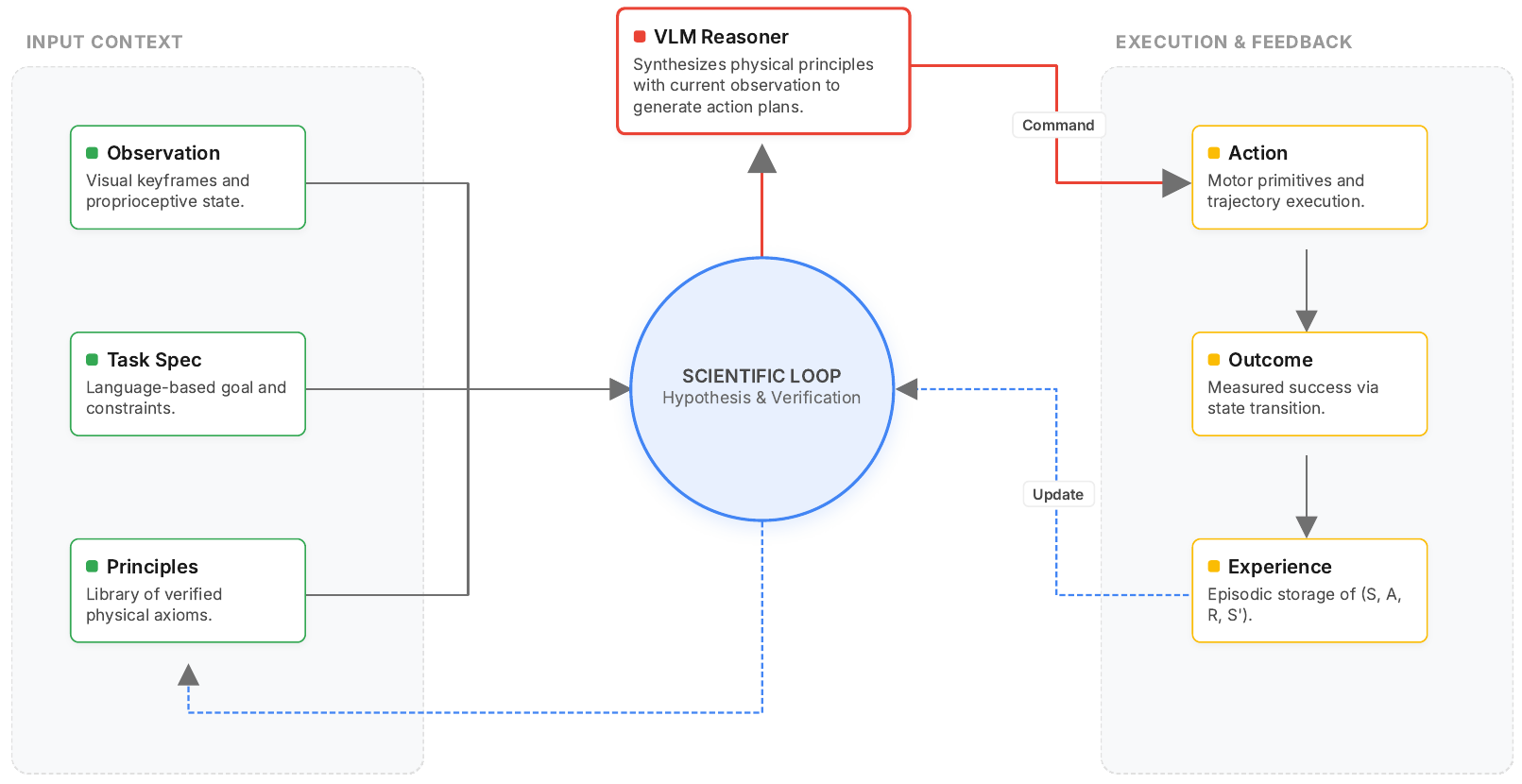}
    \caption{%
        \textbf{PhysMem overall pipeline.} PhysMem enables VLM robot planners to learn physical principles through interaction. The scientific loop transforms raw experiences into verified knowledge: collecting observations, generating hypotheses, verifying through action-level attribution, and promoting validated principles to guide future decisions.
    }
    \label{fig:assembly-demo}
\end{figure*}

\subsection{Experience Data Structure}
\label{app:experience-structure}

Each experience $e$ in episodic memory contains the following fields:

\begin{compactitem}
    \item \textbf{eid}: Unique identifier
    \item \textbf{observation}: Visual observation (image or keyframes)
    \item \textbf{action}: Selected option/action
    \item \textbf{outcome}: Success (1) or failure (0)
    \item \textbf{context}: Task description, subtask identifier
    \item \textbf{symbolic\_state}: Discrete features for filtering:
    \begin{compactitem}
        \item \texttt{action\_type}: e.g., ``pick'', ``insert'', ``push''
        \item \texttt{object\_properties}: shape, size, material
        \item \texttt{dependencies\_satisfied}: boolean
        \item \texttt{progress}: task completion fraction
    \end{compactitem}
    \item \textbf{resonance\_score}: Alignment with active principles
    \item \textbf{fail\_tag}: Failure category (if applicable)
    \item \textbf{timestamp}: Episode and step indices
\end{compactitem}


\subsection{Hypothesis Generation}
\label{app:hypothesis-generation}

\textbf{Prompt Template.}
The reflection model receives:
\begin{compactenum}
    \item 3--5 sample experiences from the cluster
    \item Extracted patterns (action types, object properties, outcomes)
    \item Existing principles and hypotheses (to avoid duplication)
    \item Task context
\end{compactenum}

\textbf{Output Format.}
Each hypothesis includes:
\begin{compactitem}
    \item \textbf{Type}: \prinAvoid, \prinPrefer, \prinSequence, etc.
    \item \textbf{Statement}: Natural language rule
    \item \textbf{Applicable Actions}: When this rule applies
    \item \textbf{Trigger Conditions}: Contextual requirements
\end{compactitem}


\subsection{Experience Clustering}
\label{app:clustering}

We use hierarchical agglomerative clustering on symbolic state features:

\begin{compactenum}
    \item Extract text embedding from symbolic state
    \item Compute pairwise cosine similarity
    \item Apply hierarchical clustering with threshold $\tau = 0.6$
    \item Retain clusters with $\geq n_{\min} = 2$ experiences
\end{compactenum}

\newpage
\subsection{Confidence Update Rules}
\label{app:confidence}

The action-level attribution updates hypothesis confidence as follows:

\begin{equation}
    \text{conf}(h) \gets
    \begin{cases}
        \min(1.0, \text{conf}(h) + 0.1 \cdot r_a) & \text{if } r_a \geq 0.7 \\
        \max(0.0, \text{conf}(h) - 0.1 \cdot (1 - r_a)) & \text{if } r_a \leq 0.3 \\
        \text{conf}(h) \pm 0.02 & \text{otherwise}
    \end{cases}
\end{equation}

where $r_a$ is the action-level success rate for actions matching the hypothesis.

\textbf{Promotion Criteria.}
\begin{compactitem}
    \item Confidence $\geq 0.8$
    \item Supporting episodes $\geq 3$
    \item Accuracy $\geq 85\%$ (supporting / total)
\end{compactitem}

\textbf{Refutation Criteria.}
\begin{compactitem}
    \item Confidence $\leq 0.3$
    \item Contradicting episodes $\geq 2$
\end{compactitem}


\subsection{Memory Management}
\label{app:memory-management}

\textbf{Capacity and Forgetting.}
Episodic memory is bounded at $N_{\max} = 3000$ experiences.
When capacity is reached, garbage collection removes:
\begin{compactenum}
    \item Folded experiences older than TTL (100 episodes)
    \item Oldest experiences by priority (folded $>$ old failures $>$ old successes)
\end{compactenum}

\textbf{Principle Decay.}
Principles decay following an Ebbinghaus forgetting curve:
\begin{equation}
    \text{score}_{t+1} = \text{score}_t \cdot \gamma, \quad \gamma = 0.995
\end{equation}
This yields approximately 50\% retention after 138 episodes without reinforcement.


\subsection{Hyperparameter Summary}
\label{app:hyperparameters}

\tabref{tab:hyperparams} lists all hyperparameters used in our experiments.

\begin{table}[h]
    \centering
    \caption{\textbf{Hyperparameter settings.}}
    \label{tab:hyperparams}
    \begin{tabular}{lcc}
        \toprule
        Parameter & Value & Description \\
        \midrule
        \multicolumn{3}{l}{\textit{Memory Capacity}} \\
        $N_{\max}$ & 3000 & Max episodic experiences \\
        Folded TTL & 100 & Episodes before folded exp. removal \\
        \midrule
        \multicolumn{3}{l}{\textit{Consolidation}} \\
        Interval & 50 & Episodes between consolidation \\
        Min cluster size & 2 & Experiences per cluster \\
        Similarity threshold & 0.6 & Clustering threshold \\
        Max hypotheses/cluster & 3 & Hypothesis generation cap \\
        \midrule
        \multicolumn{3}{l}{\textit{Principle Management}} \\
        Promotion threshold & 0.8 & Confidence for promotion \\
        Refutation threshold & 0.3 & Confidence for refutation \\
        Min supporting & 3 & Episodes for promotion \\
        Max in prompt & 5 & Principles injected \\
        Decay factor $\gamma$ & 0.995 & Ebbinghaus decay \\
        \bottomrule
    \end{tabular}
\end{table}

\subsection{Examples of Refuted Hypotheses}
\label{app:refuted-hypotheses}

The examples below are drawn from candidate hypotheses that triggered the refutation criterion ($\text{conf}(h) \leq \tau_r{=}0.3$ with $|\E_{\text{contradict}}| \geq 2$) and were therefore removed before they could condition the planner. Each entry was plausible at generation time---it was generated from a cluster of two or more experiences sharing a symbolic-state pattern---but proved empirically wrong in the deployment context once verification accumulated contradicting evidence. These cases show that verification prunes plausible-but-incorrect hypotheses rather than only catching obvious errors.


\textbf{Parts Organization.}
Two refuted hypotheses from real-world deployment:

\begin{promptlisting}
\### [REFUTED] Hypothesis (from 3 experiences)

Type:        PREFER
Statement:   PREFER placing q-shaped parts at rotation=0; rotation=90 leads to
             wasted cells.
Generated:   episode 4
Cluster:     3 experiences, all q-shaped parts placed at rotation=0 with
             successful packing
Initial conf: 0.62
Final conf:   0.18   (refuted at episode 9)
Lifespan:     5 episodes

Why plausible at generation time:
Three early successful episodes happened to use rotation=0 for white-q and red-q.
The cluster shared symbolic state (action=place, shape=q, rotation=0,
outcome=+1), and the reflection model abstracted "rotation=0 is safer for
q-shapes."

Contradicting evidence (sample):
- Experience #34: red-q at rotation=0 collided with adjacent white-q's [a,c]
  region (-1). The right-aligned internal edge of red-q overlapped with
  white-q's left edge.
- Experience #41: white-q at rotation=180 succeeded by inverting its [c]
  region toward an empty cell, enabling tighter packing than rotation=0
  would have allowed.

Why wrong in context:
Optimal q-rotation depends on the neighbor's edge alignment, not on a global
rotation preference. The early cluster confounded "rotation=0 worked" with
"the neighbors happened to be empty cells." The correct rule (later promoted)
conditions on the adjacent part's [a,c]/[b,d] occupancy.
\end{promptlisting}

\clearpage

\begin{promptlisting}
\### [REFUTED] Hypothesis (from 2 experiences)

Type:        SEQUENCE
Statement:   SEQUENCE: place black-I before any U-shaped part to reserve corner
             cells.
Generated:   episode 3
Initial conf: 0.55
Final conf:   0.22   (refuted at episode 8)
Lifespan:     5 episodes

Why plausible at generation time:
Two early high-scoring episodes happened to place black-I first; the small
contact footprint left more flexibility for later U-shape placement.

Contradicting evidence (sample):
- Experience #29: Placing black-I first forced white-U into the center,
  blocking red-q from interlocking on the boundary. Final score: -0.4
  (sub-optimal packing).
- Experience #37: Placing white-U first (rotation=180) and slotting black-I
  into the U's empty [a,c] region produced score 9.2, the best of the run.

Why wrong in context:
The two early successes were a small-sample artifact; the real constraint is
that 4-cell parts (U-shapes) have less placement flexibility than 2-cell parts
(I-shapes), so U-shapes should be placed first. The refuted hypothesis
inverted the correct size-constrained ordering. The contradicting evidence
directly informed the later SEQUENCE principle "place 4-cell parts before
2-cell parts."
\end{promptlisting}

\newpage
\textbf{Ball Navigation.}
Two refuted hypotheses from real-world deployment:

\begin{promptlisting}
\### [REFUTED] Hypothesis (from 2 experiences)

Type:        PREFER
Statement:   PREFER high speed when ball is in Zone A (initial), regardless of
             distance to OBS1.
Generated:   episode 2
Initial conf: 0.60
Final conf:   0.15   (refuted at episode 7)
Lifespan:     5 episodes

Why plausible at generation time:
Two early Zone A pushes used high speed and successfully reached the archway
region. The cluster (action=push, zone=A, speed=high, outcome=+3) suggested
a simple zone-to-speed mapping.

Contradicting evidence (sample):
- Experience #18: Ball at (y=350, x=420), ~150px from OBS1 archway. High-speed
  push caused the ball to overshoot through the archway and land on top of
  OBS3 (y=680). Stuck state, episode terminated. Score: -1.
- Experience #24: Ball at (y=380, x=380), ~120px from archway. High-speed
  diagonal push rebounded off OBS1 wall (missed bridge hole by ~30px).
  Score: -2.

Why wrong in context:
Zone alone is not the right conditioning variable; distance to OBS1 archway
matters. The two early successes happened at >200px distance where high speed
is acceptable. The correct rule (later promoted as a verified principle) is
distance-conditional: >200px -> high; 100-200px -> medium; <100px -> low.
The refuted hypothesis would have caused systematic OBS3-stuck failures had
it been promoted.
\end{promptlisting}

\newpage

\begin{promptlisting}
\### [REFUTED] Hypothesis (from 3 experiences)

Type:        AVOID
Statement:   AVOID end_direction containing "bottom" after passing through the
             archway.
Generated:   episode 5
Initial conf: 0.58
Final conf:   0.20   (refuted at episode 11)
Lifespan:     6 episodes

Why plausible at generation time:
Three post-archway pushes with "bottom" or "bottom-left" end directions
resulted in the ball drifting toward the OBS3 danger zone (y > 550). The
reflection model generalized "bottom = danger after archway."

Contradicting evidence (sample):
- Experience #44: Post-archway push with end_direction="bottom-right" at LOW
  speed successfully descended to target without entering the OBS3 zone.
  Score: +5.
- Experience #51: end_direction="bottom" at LOW speed from (y=430, x=520)
  reached the target in one step. Score: +9 (target reached with 2 steps
  remaining out of 6).

Why wrong in context:
The three original experiences that supported the hypothesis all used
HIGH/MEDIUM speed, confounding direction with speed. Once speed is held
fixed, "bottom" directions at LOW speed are in fact preferred for target
approach (as the contradicting cases above show). The refuted hypothesis
would have pushed the planner toward inefficient lateral pushes; the correct
rule conditions on speed, not on direction alone.
\end{promptlisting}

\newpage


\textbf{Balanced Stacking.}
Two refuted hypotheses from real-world deployment:

\begin{promptlisting}
\### [REFUTED] Hypothesis (from 3 experiences)

Type:        PREFER
Statement:   PREFER selecting the largest stone as the base, regardless of
             surface material.
Generated:   episode 2
Initial conf: 0.65
Final conf:   0.18   (refuted at episode 8)
Lifespan:     6 episodes

Why plausible at generation time:
Three early towers used the largest available stone as the base and reached
at least 3 layers before any collapse. The cluster (step=1, size=large,
outcome=+1) suggested size alone determines base suitability.

Contradicting evidence (sample):
- Experience #21: PAINT_SQR_L_C (largest stone, painted/smooth surface)
  selected as base. Second stone (3M_SQR_M_C) slid off during placement.
  Tower collapsed at step 2. Score: -2.
- Experience #28: WOOD_DMND_L_H (large but horizontal-bar geometry) selected
  as base. Stable for one stone, collapsed when third stone was placed
  off-center. Score: -4.

Why wrong in context:
Size is correlated with stability but not causal -- friction class and contact
area matter more. The three early successes coincidentally had high-friction
surfaces (3M, BLK) on the largest stones. The correct rule (later promoted)
is conjunctive: "largest stone WITH highest friction class
(3M > BLK > WHT > WOOD > PAINT) AND flat compact orientation." The refuted
hypothesis would have chosen PAINT_SQR_L_C or WOOD_DMND_L_H as the base,
both of which are high collapse risk.
\end{promptlisting}

\newpage

\begin{promptlisting}
\### [REFUTED] Hypothesis (from 2 experiences)

Type:        AVOID
Statement:   AVOID placing HEX-shaped stones at any layer; their small contact
             area is unstable.
Generated:   episode 4
Initial conf: 0.50
Final conf:   0.25   (refuted at episode 9)
Lifespan:     5 episodes

Why plausible at generation time:
Two episodes had HEX stones placed mid-tower (step 2 or 3) and collapsed.
The reflection model attributed collapse to the hexagonal cross-section's
reduced contact area.

Contradicting evidence (sample):
- Experience #33: 3M_HEX_L_C placed at step 4 (top layer) on a stable 3-layer
  base. Tower remained stable; final score +4.
- Experience #39: BLK_HEX_M_C placed at step 5 (top) on a 4-layer tower.
  No collapse; score +5.

Why wrong in context:
HEX instability is layer-dependent, not absolute. At lower layers HEX
cross-sections cannot bear weight from above; at top layers there is no
weight to bear, so the small contact is fine. The correct rule (later
promoted) is a SEQUENCE constraint: "place HEX stones at steps 4-5, not
earlier." The blanket AVOID would have eliminated viable top-layer
placements.
\end{promptlisting}

\newpage


\textbf{Brick Insertion (simulation).}
Two refuted hypotheses from the Reflect-VLM brick insertion benchmark:

\begin{promptlisting}
\### [REFUTED] Hypothesis (from 4 experiences, hard difficulty)

Type:        AVOID
Statement:   AVOID inserting blue before red.
Generated:   episode 6   (difficulty: hard)
Initial conf: 0.70
Final conf:   0.20   (refuted at episode 14)
Lifespan:     8 episodes

Why plausible at generation time:
On hard difficulty, four early episodes had insert(blue) attempts fail when
red was still on the table. The cluster (action=insert, target=blue,
red_remaining=true, outcome=fail) suggested a direct order constraint
between blue and red.

Contradicting evidence (sample):
- Experience #67: Inserted blue successfully while red was still on the
  table; the blocker was actually green, which had been relocated in the
  prior step. blue-before-red worked once green was out of the way.
- Experience #74: insert(red) BEFORE blue also failed, because green still
  blocked red's slot. The blue/red order had no causal role in either
  direction.

Why wrong in context:
The four early failures all coincidentally had green in a position that
blocked blue's slot. Symbolic-state clustering matched on (target=blue,
red_remaining=true) but missed the true blocker (green). The correct rule
(later promoted) is "AVOID insert(X) when target_blocked_by includes any
color Y," conditioning on the actual blocking variable rather than the
spurious blue-red correlation. This is a textbook confounding-variable
failure that verification correctly caught.
\end{promptlisting}

\newpage

\begin{promptlisting}
\### [REFUTED] Hypothesis (from 2 experiences, medium difficulty)

Type:        PREFER
Statement:   PREFER reorient before every insert action for elongated bricks.
Generated:   episode 3   (difficulty: medium)
Initial conf: 0.55
Final conf:   0.22   (refuted at episode 10)
Lifespan:     7 episodes

Why plausible at generation time:
Two early episodes had failed insert attempts for elongated bricks that
succeeded after a reorient action. The reflection model abstracted
"elongated bricks always need reorientation."

Contradicting evidence (sample):
- Experience #45: Elongated yellow brick picked up in correct orientation
  directly from the table; insert succeeded without reorient. Adding a
  reorient would have consumed an action step unnecessarily.
- Experience #52: After reorient on an already-correctly-oriented brick,
  the brick ended up MIS-oriented (rotated 90 degrees away from target).
  Subsequent insert failed; required a second reorient to recover.

Why wrong in context:
Reorientation is conditional on the brick's pickup orientation, not on its
shape. Mandatory reorient introduces failures when the brick was already
correct. The correct rule (promoted later) conditions on a quaternion-error
check between the held brick's orientation and the target slot orientation,
not on shape category. This refuted hypothesis was recovered by the
failure-mode category "over-generalization" (Section \ref{app:failure-modes}):
a narrow-context observation expanded into a universal rule.
\end{promptlisting}

\textbf{Aggregate refutation statistics.} Across the four task suites, verification refuted approximately $31\%$ of generated hypotheses ($87$ of $281$ across all real-world and simulation runs) before they could condition the planner. Refuted hypotheses had a median lifespan of $6$ episodes (IQR $4$--$9$) and a median initial confidence of $0.58$, indicating that they were not obviously bad at generation time---they accumulated contradicting evidence only after targeted interaction tested their predictions. Without verification (\tabref{\tabablation}, \emph{w/o Verification}: $-12\%$ on medium), these plausible-but-incorrect hypotheses would have been promoted to principles and degraded the planner's decisions; their refutation is the load-bearing function of the verification mechanism.

\newpage

\clearpage
\section{Complete Ablation Studies}
\label{app:ablations}

We conducted comprehensive ablations to validate each component of \method
across all three difficulty levels. Each configuration runs 100 episodes
using Gemini-3-Flash with thinking mode. We report both success rate and
relative token consumption (normalized to the full system at each difficulty).

\subsection{Full Results}

\tabref{tab:full-ablations} presents complete ablation results with token
consumption analysis.

\begin{table*}[t]
    \centering
    \caption{\textbf{Complete ablation results across difficulty levels.}
    Success rates and token consumption on simulation benchmark (100 episodes each).
    Token consumption is normalized relative to the full system at each difficulty level.
    $\Delta$ indicates absolute change from the full system. Best accuracy in \textbf{bold}.}
    \label{tab:full-ablations}
    \begin{tabular}{@{}l|ccc|ccc|ccc@{}}
        \toprule
        & \multicolumn{3}{c|}{Easy (2--3 bricks)} & \multicolumn{3}{c|}{Medium (4--5 bricks)} & \multicolumn{3}{c}{Hard (6--8 bricks)} \\
        \cmidrule(lr){2-4} \cmidrule(lr){5-7} \cmidrule(lr){8-10}
        Configuration & Success & $\Delta$ & Tokens & Success & $\Delta$ & Tokens & Success & $\Delta$ & Tokens \\
        \midrule
        \method (Full) & 89\% & --- & 1.0$\times$ & 76\% & --- & 1.0$\times$ & \textbf{39\%} & --- & 1.0$\times$ \\
        \midrule
        \multicolumn{10}{l}{\textit{Memory Architecture Ablations}} \\
        w/o Episodic Memory & 54\% & $-35$ & 0.3$\times$ & 37\% & $-39$ & 0.25$\times$ & 14\% & $-25$ & 0.2$\times$ \\
        w/o Working Memory & 84\% & $-5$ & 0.85$\times$ & 69\% & $-7$ & 0.8$\times$ & 28\% & $-11$ & 0.75$\times$ \\
        w/o Long-term Memory & 81\% & $-8$ & 1.15$\times$ & 64\% & $-12$ & 1.25$\times$ & 26\% & $-13$ & 1.35$\times$ \\
        \midrule
        \multicolumn{10}{l}{\textit{Mechanism Ablations}} \\
        w/o Resonance Filtering & 81\% & $-8$ & 1.15$\times$ & 58\% & $-18$ & 1.3$\times$ & 21\% & $-18$ & 1.45$\times$ \\
        w/o Verification & 85\% & $-4$ & 0.9$\times$ & 64\% & $-12$ & 0.85$\times$ & 27\% & $-12$ & 0.8$\times$ \\
        w/o Forgetting & \textbf{91\%} & $+2$ & 1.8$\times$ & \textbf{78\%} & $+2$ & 3.4$\times$ & 36\% & $-3$ & 4.8$\times$ \\
        w/o Folding & 88\% & $-1$ & 1.4$\times$ & 74\% & $-2$ & 2.1$\times$ & 37\% & $-2$ & 2.8$\times$ \\
        \midrule
        \multicolumn{10}{l}{\textit{Simplified Baselines}} \\
        Direct Retrieval & 48\% & $-41$ & 0.45$\times$ & 23\% & $-53$ & 0.55$\times$ & 8\% & $-31$ & 0.50$\times$ \\
        Only Episodic & 52\% & $-37$ & 0.6$\times$ & 28\% & $-48$ & 0.7$\times$ & 10\% & $-29$ & 0.65$\times$ \\
        Only Principles & 67\% & $-22$ & 0.4$\times$ & 51\% & $-25$ & 0.35$\times$ & 21\% & $-18$ & 0.3$\times$ \\
        \bottomrule
    \end{tabular}
\end{table*}

\subsection{Analysis by Component}

\subsubsection{Memory Architecture Ablations}

\textbf{Episodic memory} is the load-bearing tier. Without it, raw experiences are not stored at all and the system has nothing to draft hypotheses from; performance collapses to $54\%$, $37\%$, and $14\%$ across the three difficulties (about $25$ to $39$ points below the full system). Token usage drops to $0.2$--$0.3\times$ for the trivial reason that nothing is being consolidated.

\textbf{Working memory} holds hypotheses while they are being tested, and its value grows with task complexity ($-5\%$ on easy, $-7\%$ on medium, $-11\%$ on hard). On easy tasks the right move is usually obvious enough that hypothesis exploration adds little; on hard tasks the planner needs to actively test patterns like ``does brown block pink?'' across several episodes, which is exactly what working memory makes possible.

\textbf{Long-term memory} is what lets verified hypotheses persist as principles between episodes. Without it the system rediscovers the same patterns every run, which costs $-8\%$ on easy and grows to $-12\%$ to $-13\%$ on medium and hard, where complex dependency chains depend on stable principles. Tokens climb to $1.15$--$1.35\times$ because the model is generating hypotheses repeatedly that ought to have been settled already.

\subsubsection{Mechanism Ablations}

\textbf{Resonance filtering} retrieves principles based on prediction--outcome alignment rather than raw similarity, and like working memory it grows in importance with difficulty ($-8\%$ on easy, $-18\%$ on medium and hard). On harder tasks more principles accumulate (especially \prinAvoid constraints from failures), and without filtering the planner receives a steady stream of conflicting guidance.

\textbf{Verification} checks each hypothesis against later experiences before letting it shape the prompt, and it matters most when hypotheses get harder to vet by inspection ($-4\%$ on easy, $-12\%$ on medium and hard). Easy hypotheses tend to be self-evidently right or wrong from a single rollout; medium and hard hypotheses are about partial orderings and dependency chains, where guessing wrong propagates downstream.

\textbf{Forgetting} (Ebbinghaus-style decay of old principles) reveals an accuracy--efficiency trade-off that depends on difficulty. Disabling it gives a small bump on easy and medium ($+2\%$ each) but at $1.8\times$ to $3.4\times$ the tokens. On hard tasks the accumulated noise actually \emph{hurts} accuracy ($-3\%$ at $4.8\times$ tokens), because stale principles begin competing with newer, more accurate ones.

\textbf{Folding} compresses redundant experiences into representative clusters. Removing it costs almost nothing in accuracy ($-1\%$ to $-2\%$) but multiplies token cost by $1.4\times$ to $2.8\times$ across difficulties, so its value is efficiency rather than accuracy.

\subsubsection{Simplified Baselines}

\textbf{Direct Retrieval} stores raw experiences and pulls the nearest one by state-embedding distance. It fails across the board ($48\%$, $23\%$, $8\%$ across easy/medium/hard), and the failure mode is the same in every case: small state differences make the retrieved action a bad fit for the new scene, which is exactly the slot abstraction is meant to fill.

\textbf{Only Episodic.} Keeping episodic memory but skipping hypotheses and principles is only marginally better than Direct Retrieval ($52\%$, $28\%$, $10\%$). Richer experience records help a little; the gap from the full system ($-37$ to $-48$) is again the size of what the symbolic abstraction layer adds.

\textbf{Only Principles.} Pre-loading a fixed principle set without any test-time adaptation reaches moderate accuracy ($67\%$, $51\%$, $21\%$) on tasks where the prior generalizes well, but it cannot recover from a deployment that drifts away from the prior. The $18$- to $25$-point gap to the full system is the value of adaptation on top of a good prior.

\subsection{Summary of the Ablation}

Five patterns come out of the table together. The three-tier memory is the only configuration that holds up across difficulties: removing any tier (episodic, working, or long-term) costs anywhere from $5$ to $39$ points, with episodic memory carrying the largest weight ($-35$ to $-39$) and working and long-term memory growing in importance on harder tasks. Principle abstraction does most of the lifting; the two baselines that fall back on raw experience (Direct Retrieval and Only Episodic) drop $37$--$53$ points on medium and hard, which is the size of what the abstraction layer adds. Within the mechanism, resonance filtering, verification, and working memory each grow more important as tasks get harder, and the relative ordering is stable across difficulties. Forgetting trades a small accuracy cost for a $3$--$5\times$ token saving, and on hard tasks it actually helps because stale principles begin to interfere with newer ones. Finally, a fixed prior alone (Only Principles) covers a fair amount of ground but cannot recover when the deployment drifts from the prior, leaving an $18$- to $25$-point gap that test-time adaptation closes.


\section{Robustness, Failure Modes, and Runtime}
\label{app:robustness}

\subsection{Sensitivity Analysis}
\label{app:sensitivity}

\method's design includes three threshold parameters: a promotion threshold $\tau_p$ (default $0.8$), a clustering threshold $\tau_c$ (default $0.7$), and a decay rate $\gamma$ (default $0.995$). On medium-difficulty brick insertion (Gemini-3-Flash, 100 episodes per condition):
\begin{compactitem}
    \item $\tau_p \in \{0.6, 0.7, 0.8, 0.9\}$: $71\%, 74\%, 76\%, 73\%$. Optimum is broad; $[0.7, 0.9]$ stays within $3\%$ of default.
    \item $\tau_c \in \{0.5, 0.6, 0.7, 0.8\}$: $72\%, 75\%, 76\%, 74\%$. Insensitive within realistic range.
    \item $\gamma \in \{0.99, 0.995, 0.999, 1.0\}$: $74\%, 76\%, 77\%, 73\%$. Mild decay helps; no-decay underperforms because stale principles compete with newer ones.
\end{compactitem}
Across $4{\times}4{\times}4 = 64$ threshold configurations, $59$ land within $5\%$ of the default ($76\%$), and even the worst ($66\%$) still beats every non-\method baseline in the main paradigm comparison, so the qualitative findings do not depend on a particular choice of thresholds.

\subsection{Failure Mode Analysis}
\label{app:failure-modes}

We manually categorize the $24\%$ of medium-difficulty failures ($24/100$ episodes) into three modes.

\textbf{Over-generalization ($40\%$).} A principle that was committed from a narrow context gets re-applied where it no longer holds. For example, an \prinAvoid principle learned for thin bricks (``do not place piece B before piece A'') is retrieved for a wide-brick scene where the order constraint is reversed. Stronger context conditioning during retrieval would help here; the current symbolic-state matcher catches object type but not always size.

\textbf{Delayed correction ($35\%$).} A principle that has stopped being true (a friction estimate from earlier episodes, say) is only refuted after several contradicting experiences accumulate, a $2$--$4$ episode lag on average. Lowering the refutation threshold $\tau_r$ or weighting contradicting evidence more heavily would shorten the lag, at the cost of over-correcting.

\textbf{Planner--principle mismatch ($25\%$).} The principle is correct but the VLM planner ignores or misapplies it, which is a prompt-formatting issue rather than a memory issue. Tighter principle templates in the planner prompt or a structured-output planner head are the natural fixes.

The first two categories ($75\%$ together) are addressable by refining the memory mechanism. The third sits at the limit of what frozen-base prompting can do and would need execution-layer integration to fix.

\subsection{Runtime Overhead}
\label{app:runtime}

\method adds memory operations on two timescales. Per-step retrieval (symbolic filtering, semantic ranking, prompt injection) adds about $180$~ms per planning step on a single workstation, dominated by the VLM-based relevance judgment. Asynchronous consolidation (clustering, hypothesis drafting, action-conditioned confidence updates) runs in a background thread every $15$~s with an average wall-clock cost of $1.4$~s per cycle and does not block the planner. Total wall-clock overhead averages around $8\%$ of base VLM inference time on the brick-insertion benchmark and tracks the same range on the real-world tasks. Token overhead is reported in the main paradigm comparison ($1.0\times$ for \method versus $0.55\times$ for Direct Retrieval and $2.5\times$ for MemoryVLA-style).

\newpage
\section{Hypothesis and Principle Quantity Analysis}
\label{app:quantity}

We investigated the optimal number of hypotheses to generate per cluster
and principles to include in prompts.

\begin{table}[h]
    \centering
    \caption{\textbf{Quantity configuration results.} H/C = hypotheses per cluster,
    P/P = principles in prompt. Evaluated on medium difficulty (100 episodes each).}
    \label{tab:quantity}
    \begin{tabular}{lccccc}
        \toprule
        Config & H/C & P/P & Success & Principles & Tokens \\
        \midrule
        Sparse & 1 & 3 & 71\% & 12 & 0.7$\times$ \\
        Balanced & 2 & 5 & \textbf{76\%} & 18 & 1.0$\times$ \\
        Dense & 3 & 10 & 74\% & 31 & 1.6$\times$ \\
        Hypo-Heavy & 3 & 3 & 73\% & 26 & 1.2$\times$ \\
        Principle-Heavy & 1 & 10 & 72\% & 14 & 1.1$\times$ \\
        \bottomrule
    \end{tabular}
\end{table}

\textbf{Analysis.}
The results show moderate variance across configurations (71--76\%), with the
``balanced'' configuration (2 hypotheses per cluster, 5 principles in prompt)
achieving the best performance. Sparse configurations under-explore the hypothesis
space, while dense configurations introduce noise from unverified hypotheses.
The balanced setting provides a favorable trade-off between exploration coverage
and computational cost (1.0$\times$ baseline tokens).

\newpage
\section{Simulation Experiment Details}
\label{app:simulation}

This appendix provides comprehensive details for our simulation-based experiments
using the Brick Assembly environment. The simulation enables rapid, reproducible
evaluation of \method across many episodes with ground-truth feedback.


\subsection{Environment Overview}
\label{app:sim-overview}

We use the Reflect-VLM simulation environment~\citep{feng2025reflective}, built on
MuJoCo physics simulation with a Franka Emika Panda robot arm.

\textbf{Simulator Configuration.}
\begin{compactitem}
    \item \textbf{Physics Engine}: MuJoCo 3.x with realistic contact dynamics
    \item \textbf{Robot}: Franka Panda 7-DOF arm with parallel-jaw gripper
    \item \textbf{Control}: Differential Inverse Kinematics with nullspace control
    \item \textbf{Timestep}: 1ms simulation, 20ms control frequency (frame skip = 20)
    \item \textbf{Rendering}: 1280$\times$720 RGB images, offscreen rendering
\end{compactitem}


\subsection{\taskBrick: Brick Assembly}
\label{app:task-brick}

The brick assembly task requires the robot to insert multiple colored bricks
into a board fixture, testing dependency reasoning and sequential planning.

\subsubsection{Task Description}
\label{app:brick-description}

\textbf{Setup.}
A board fixture with insertion slots is placed on the table, surrounded by
2--5 colored bricks with varying shapes. Each brick must be inserted into
its designated slot on the board. Bricks are procedurally generated with:
\begin{compactitem}
    \item \textbf{Shape variation}: Rectangular blocks, elongated beams,
          L-shapes, and nail-like pegs
    \item \textbf{Size variation}: 3--25 voxel units in length, 3--8 in width
    \item \textbf{Color coding}: 9 distinct colors (red, blue, green, yellow,
          orange, purple, brown, pink, gray)
    \item \textbf{Slot features}: Through-slots requiring specific orientations
\end{compactitem}

\textbf{Objective.}
Insert all bricks into their designated board slots. Success requires each
brick to be within 5mm of its target position and correctly oriented
(quaternion error $< 0.02$ radians).

\textbf{Challenge.}
The task is challenging because:
(1) bricks have \textbf{dependency constraints}, where some bricks physically
    block insertion of others until removed,
(2) bricks may land in incorrect orientations requiring \textbf{reorientation},
(3) the VLM must infer the \textbf{correct insertion sequence} from visual
    observation of spatial relationships, and
(4) failed insertions provide only sparse error signals without explicit
    explanation of the blocking relationship.

\textbf{Dependency Graph.}
Dependencies are computed from voxel spatial intersections: if brick $A$'s
insertion path intersects brick $B$'s current position, then $B$ must be
removed before $A$ can be inserted. This creates a directed acyclic graph
of assembly constraints that the VLM must implicitly discover through
trial and error.

\begin{figure*}[t]
    \centering
    \includegraphics[width=\textwidth]{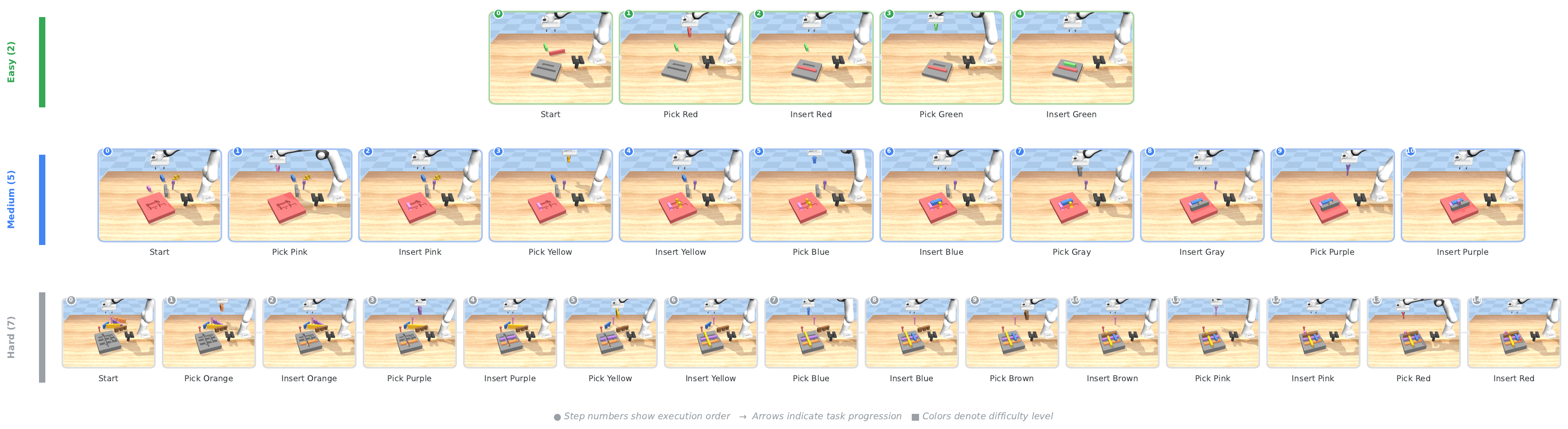}
    \caption{%
        \textbf{Assembly task demonstration across difficulty levels.}
        Each row shows the oracle planner's execution sequence from initial
        state to completion. Step numbers indicate execution order.
        \textbf{Easy} (green, 2 bricks): 5 steps total, minimal dependencies.
        \textbf{Medium} (blue, 5 bricks): 11 steps with moderate dependency chains.
        \textbf{Hard} (gray, 7 bricks): 15 steps requiring complex ordering to
        avoid blocking. The VLM must learn correct insertion sequences through
        trial and error, as inserting bricks in the wrong order causes failures
        that provide only sparse error signals.
    }
    \label{fig:assembly-demo}
\end{figure*}

\textbf{Difficulty Levels.}
We evaluate across three difficulty levels based on brick count and dependency
complexity (\figref{fig:assembly-demo}):
\begin{compactitem}
    \item \textbf{Easy (2--3 bricks)}: Minimal dependencies, typically requiring
          5--7 pick-and-place steps. Most orderings succeed.
    \item \textbf{Medium (4--5 bricks)}: Moderate dependency chains where 2--3
          bricks may block others. Requires 9--11 steps and partial ordering
          constraints.
    \item \textbf{Hard (6--8 bricks)}: Complex dependency graphs with multiple
          blocking relationships. Requires 13--17 steps; incorrect orderings
          lead to repeated failures until the VLM learns the constraints.
\end{compactitem}

\subsubsection{VLM Interface}
\label{app:brick-vlm}

\textbf{Visual Input.}
The VLM receives two images per decision step:
\begin{compactenum}
    \item \textbf{Goal Image}: Target assembly configuration showing all bricks
          correctly inserted in the board (reference for completion)
    \item \textbf{Current Image}: Current workspace state captured from a
          fixed ``table\_back'' camera showing:
          \begin{compactitem}
              \item Board fixture with any already-inserted bricks
              \item Remaining bricks on the table surface
              \item Robot gripper (potentially holding a brick)
          \end{compactitem}
\end{compactenum}

Both images are resized to 512$\times$512 pixels and encoded as base64 PNG
for transmission to the VLM API.

\textbf{Action Output.}
The VLM outputs one of five action primitives:
\begin{compactitem}
    \item \texttt{pick up [color]}: Grasp a brick from the table or board
    \item \texttt{put down [color]}: Release held brick onto the table
    \item \texttt{reorient [color]}: Rotate held brick using a fixture surface
          to achieve correct insertion orientation
    \item \texttt{insert [color]}: Insert held brick into its board slot
    \item \texttt{done}: Declare task completion
\end{compactitem}

The \texttt{[color]} argument must match one of the available brick colors
(e.g., ``red'', ``blue'', ``green'').

\subsubsection{Scoring Mechanism}
\label{app:brick-scoring}

Each action returns an error code indicating success or failure:
\begin{compactitem}
    \item \textbf{err = 0}: Action executed successfully
    \item \textbf{err = -1}: Invalid precondition (e.g., ``pick up'' while
          already holding an object, ``insert'' without holding anything)
    \item \textbf{err = -2}: Object already in specified state (e.g.,
          ``pick up'' an object already in hand)
\end{compactitem}

\textbf{No-Progress Detection.}
Beyond error codes, we detect semantic failures:
\begin{compactitem}
    \item \textbf{Insert no-progress}: Action succeeds (err=0) but gripper
          still holds the brick, indicating the slot is blocked
    \item \textbf{Pick no-progress}: Action succeeds but holding state unchanged
\end{compactitem}

\textbf{Episode Success.}
An episode succeeds when \texttt{env.is\_success()} returns true, verified by:
\begin{compactitem}
    \item All bricks within 5mm of target position
    \item All bricks correctly oriented (quaternion error $< 0.02$)
    \item Special case: nail-type bricks must be upright (z-axis alignment $> 0.9$)
\end{compactitem}

\textbf{Auto-Completion.}
The system checks \texttt{is\_success()} after every action. If the goal
is achieved mid-episode (before the VLM outputs ``done''), the episode
automatically terminates as successful.

\subsubsection{Low-Level Executor}
\label{app:brick-executor}

Given a VLM action command, the low-level controller executes:

\textbf{Pick Up.}
\begin{compactenum}
    \item Identify target brick by color name
    \item Compute grasp pose based on brick geometry
    \item Move gripper to approach position above brick
    \item Lower gripper and close fingers to grasp
    \item Lift brick to safe transport height
\end{compactenum}

\textbf{Insert.}
\begin{compactenum}
    \item Verify brick is in hand and correctly oriented
    \item Align brick's insertion site with board's target hole
    \item Lower brick along insertion trajectory
    \item Release gripper once contact is detected
    \item Return to home position
\end{compactenum}

\textbf{Reorient.}
\begin{compactenum}
    \item Place brick on fixture surface
    \item Release and re-grasp from side orientation
    \item Rotate to achieve upright configuration
    \item Verify orientation before proceeding
\end{compactenum}

All motions use differential IK with collision avoidance and velocity limits.

\subsubsection{Symbolic State Extraction}
\label{app:brick-symbolic}

For memory-based learning, we extract symbolic state features from each
experience:
\begin{compactitem}
    \item \textbf{action\_type}: pick, insert, reorient, or putdown
    \item \textbf{target\_signature}: Shape description of target brick
          (e.g., ``elongated rectangular block'')
    \item \textbf{holding\_signature}: Shape of currently held brick (if any)
    \item \textbf{dependencies\_satisfied}: Boolean indicating if all
          prerequisite bricks are inserted
    \item \textbf{target\_blocked\_by\_colors}: List of brick colors currently
          blocking the target slot
    \item \textbf{remaining\_pieces}: Colors of bricks not yet inserted
    \item \textbf{progress}: Fraction of bricks successfully inserted (0--1)
\end{compactitem}

These features enable the consolidation engine to cluster similar experiences
and generate hypotheses about blocking relationships without requiring
explicit dependency graph access.


\subsection{Example Learned Principles}
\label{app:brick-principles}

Example principles discovered by \method during simulation experiments:

\begin{compactitem}
    \item \prinSequence: ``When inserting elongated blocks, ensure all
          blocking pieces in the target slot are removed first''
    \item \prinAvoid: ``Don't attempt to insert a brick when dependencies
          are not satisfied; the slot will be physically blocked''
    \item \prinPrefer: ``If an insert action fails repeatedly, put down
          the current brick and remove the blocking piece first''
    \item \prinAvoid: ``Don't reorient the same brick more than twice. If
          orientation still seems wrong, the issue may be slot blocking''
    \item \prinSequence: ``For L-shaped bricks, insert the base portion
          before attempting to insert adjacent pieces''
\end{compactitem}


\subsection{Simulation Scaling Experiments}
\label{app:sim-experiments}

The simulation scaling experiments (VLM difficulty scaling and principle scaling)
are presented in the main text (\secref{sec:exp:difficulty} and \secref{sec:exp:scaling}).
This section provides additional implementation details.

\textbf{Evaluation Protocol.}
For each VLM and difficulty level combination, we run 100 independent episodes
with fresh random seeds. Success is defined as correctly inserting all bricks
within the maximum step limit. We report mean success rate with standard deviation
computed across 3 random seeds for variance estimation.

\textbf{Principle Counting.}
The principle count at each checkpoint is determined by the number of verified
principles in long-term memory. We measure performance at powers of 2 (1, 2, 4,
8, 16, 32, 64, 128 principles) by saving memory snapshots during a 500-episode
training run and evaluating each snapshot on a held-out test set of 50 episodes.

\newpage
\section{Real-World Experiment Details}
\label{app:realworld}

This appendix provides comprehensive details for our three real-world manipulation
tasks. Each task is designed to test \method's ability to learn physics-grounded
principles from interaction experience in settings where pre-trained VLM knowledge
is insufficient.


\subsection{\textbf{Hardware Configuration}}
\label{app:hardware}

All experiments share a common hardware platform:

\begin{compactitem}
    \item \textbf{Robot Arm}: xArm6 6-DOF robotic arm
    \item \textbf{End Effector}: TPU-printed fin-ray effect soft grippers,
          providing compliant grasping for irregular objects
    \item \textbf{Camera System}:
    \begin{compactitem}
        \item \textbf{Top-down}: Intel RealSense D435, 1280$\times$720 @ 15fps,
              mounted above workspace for global scene observation
        \item \textbf{Wrist-mounted}: Intel RealSense D435, 1280$\times$720 @ 15fps,
              for close-up manipulation views
        \item \textbf{Base camera} (optional): Intel RealSense D435,
              1280$\times$720 @ 15fps, for side-angle views
        \item \textbf{External cameras}: For documentation and qualitative analysis
    \end{compactitem}
    \item \textbf{Compute}: NVIDIA RTX 4090 for VLM inference through the API.
\end{compactitem}

\textbf{Model Configuration.}
\begin{compactitem}
    \item \textbf{VLM Planner}: Gemini-3.0-Flash with thinking mode enabled,
          invoked at episode-level for high-level planning decisions
    \item \textbf{Reflection Model}: Qwen3-VL with thinking mode,
          running asynchronously every 15 seconds for hypothesis generation
          and consolidation
\end{compactitem}

\clearpage
\subsection{\textbf{\taskParts}}
\label{app:task-parts}

The parts organization task requires the robot to efficiently pack irregularly-shaped
3D-printed components onto a discrete grid, testing spatial reasoning and
learning optimal placement strategies through trial and error.

\subsubsection{Task Description}
\label{app:parts-description}

\textbf{Setup.}
Five distinct 3D-printed parts (labeled 001--005) are presented to the robot in fixed order.
Each part occupies 2--4 adjacent grid cells in various configurations (L-shapes, U-shapes,
I-shapes, T-shapes). The workspace contains a $3 \times 10$ grid (30 total cells)
where parts must be placed sequentially.

\textbf{Placement Constraints.}
To simplify the decision space, the selectable grid region is constrained to
\textbf{adjacent areas} relative to already-occupied cells. Specifically, the VLM
can only select grid cells within $\pm 3$ columns from the current occupied boundary.
This prevents the VLM from placing parts in isolated regions and encourages
compact packing strategies. Note that parts may share the same grid cell indices
when their 3D structures allow vertical overlap (\eg, one part's overhang above
another part's base).

\textbf{Objective.}
The robot must place all five parts onto the grid while minimizing the total number
of occupied grid cells. This requires learning efficient packing strategies,
including optimal rotation angles and placement positions that account for
part geometries and remaining space.

\textbf{Challenge.}
The task is challenging because:
(1) optimal placement depends on the sequence of previously placed parts,
(2) irregular part shapes create complex spatial constraints,
(3) the VLM must reason about rotation effects on cell occupancy,
(4) parts can structurally overlap in 3D while sharing grid indices, and
(5) the constrained selection region requires planning within local neighborhoods.

\subsubsection{VLM Interface}
\label{app:parts-vlm}

\textbf{Visual Input.}
The VLM receives three images per decision step:
\begin{compactenum}
    \item \textbf{Wrist Camera RGB}: Close-up view of the current part to be placed,
          showing detailed geometry and grasping orientation
    \item \textbf{Enhanced Third-View Camera}: Top-down view of the grid workspace
          with visual overlays showing:
          \begin{compactitem}
              \item Grid cell boundaries and numbering (1--30)
              \item Currently occupied cells (highlighted)
              \item Selectable placement region ($\pm 3$ columns from occupied boundary)
          \end{compactitem}
    \item \textbf{Part Template with Grid Watermark}: Reference image showing the
          current part type with a grid watermark overlay indicating which cells
          the part would occupy at different rotation angles. This helps the VLM
          understand the mapping between rotation and grid occupancy.
\end{compactenum}

\textbf{Action Output.}
The VLM outputs depend on the part type:
\begin{compactitem}
    \item \textbf{2-grid and 3-grid parts}: Output only the grid cell indices
          (\eg, \texttt{place(3, 13, 14)} for 3 cells)
    \item \textbf{4-grid parts}: Output grid cell indices \textbf{and} rotation angle
          (\eg, \texttt{place(6, 7, 16, 17, rotation=90)})
\end{compactitem}
The rotation angle is required for 4-grid parts because different rotations can
result in the same grid occupancy pattern but different physical orientations.

\subsubsection{Part Naming and Spatial Coordinate System}
\label{app:parts-coordinates}

To enable precise spatial reasoning and hypothesis formation about part interactions,
we introduce a standardized naming convention and a template-based coordinate system.

\textbf{Part Naming Convention.}
Each part is uniquely identified by \texttt{\{color\}-\{shape\}}:
\begin{compactitem}
    \item \textbf{Colors}: black, white, red
    \item \textbf{Shapes}: L-shape (3 cells), q-shape (3 cells), I-shape (2 cells), U-shape (4 cells)
\end{compactitem}
The six parts in this task are: \texttt{red-L}, \texttt{white-q}, \texttt{red-q}, \texttt{white-U}, \texttt{black-U}, \texttt{black-I}.

\textbf{Template-Based Coordinate System.}
Since all parts occupy at most 4 cells, we represent each part's local geometry
using a $2 \times 2$ template grid with cells labeled \texttt{[a,b]} (top row) and
\texttt{[c,d]} (bottom row):

\begin{center}
\begin{verbatim}
    +---+---+
    | a | b |
    +---+---+
    | c | d |
    +---+---+
\end{verbatim}
\end{center}

\noindent Each shape's default (0°) cell occupancy is defined as:
\begin{compactitem}
    \item \textbf{L-shape}: occupies \texttt{[a, c, d]} -- vertical bar on left, horizontal extension at bottom-right
    \item \textbf{q-shape}: occupies \texttt{[a, c, d]} -- similar to L but with different internal geometry
    \item \textbf{I-shape}: occupies \texttt{[a, c]} -- vertical bar on the left column
    \item \textbf{U-shape}: occupies \texttt{[a, b, c, d]} -- all four cells
\end{compactitem}

\textbf{Rotation Transformations.}
Counterclockwise rotation transforms the template coordinates as follows:
\begin{compactitem}
    \item \textbf{90° CCW}: \texttt{[[a,b],[c,d]]} $$\rightarrow$$ \texttt{[[b,d],[a,c]]}
          (original $a \to c$, $b \to a$, $c \to d$, $d \to b$)
    \item \textbf{180° CCW}: \texttt{[[a,b],[c,d]]} $$\rightarrow$$ \texttt{[[d,c],[b,a]]}
          (original $a \to d$, $b \to c$, $c \to b$, $d \to a$)
    \item \textbf{270° CCW}: \texttt{[[a,b],[c,d]]} $$\rightarrow$$ \texttt{[[c,a],[d,b]]}
          (original $a \to b$, $b \to d$, $c \to a$, $d \to c$)
\end{compactitem}

\noindent This allows consistent reference to part sub-regions regardless of rotation.
For example, the ``left vertical portion'' of a U-shape is always \texttt{[a,c]},
whether the part is at 0° or rotated.

\textbf{Part-Specific Offset Information.}
Even when occupying the same template cells, parts have internal geometric biases:
\begin{compactitem}
    \item \textbf{black-U / white-U}: At 0°, the physical mass is biased toward \texttt{[b,d]};
          cells \texttt{[a,c]} contain mostly empty space within the ``U'' opening.
    \item \textbf{white-q}: At 0°, the left edge aligns with the left boundary of \texttt{[a,c]}.
    \item \textbf{red-q}: At 0°, the right edge aligns with the right boundary of \texttt{[a,c]}.
    \item \textbf{black-I}: At 0°, the top and right edges align with the top-right of \texttt{[a,c]}.
\end{compactitem}
These offsets determine whether adjacent parts can physically interlock or overlap
when sharing grid cells.

\textbf{Using Coordinates for Spatial Reasoning.}
This coordinate system enables precise hypothesis formation about part interactions:
\begin{compactitem}
    \item ``PREFER placing the \texttt{[c]} region of \texttt{white-q} toward the right side''
    \item ``AVOID complete overlap between the \texttt{[a,c]} regions of two q-shaped parts''
    \item ``When \texttt{white-U} is rotated 180°, its \texttt{[a,c]} region can interlock with another U-shape's \texttt{[b,d]} region''
\end{compactitem}

\begin{figure}[t]
    \centering
    \includegraphics[width=\columnwidth]{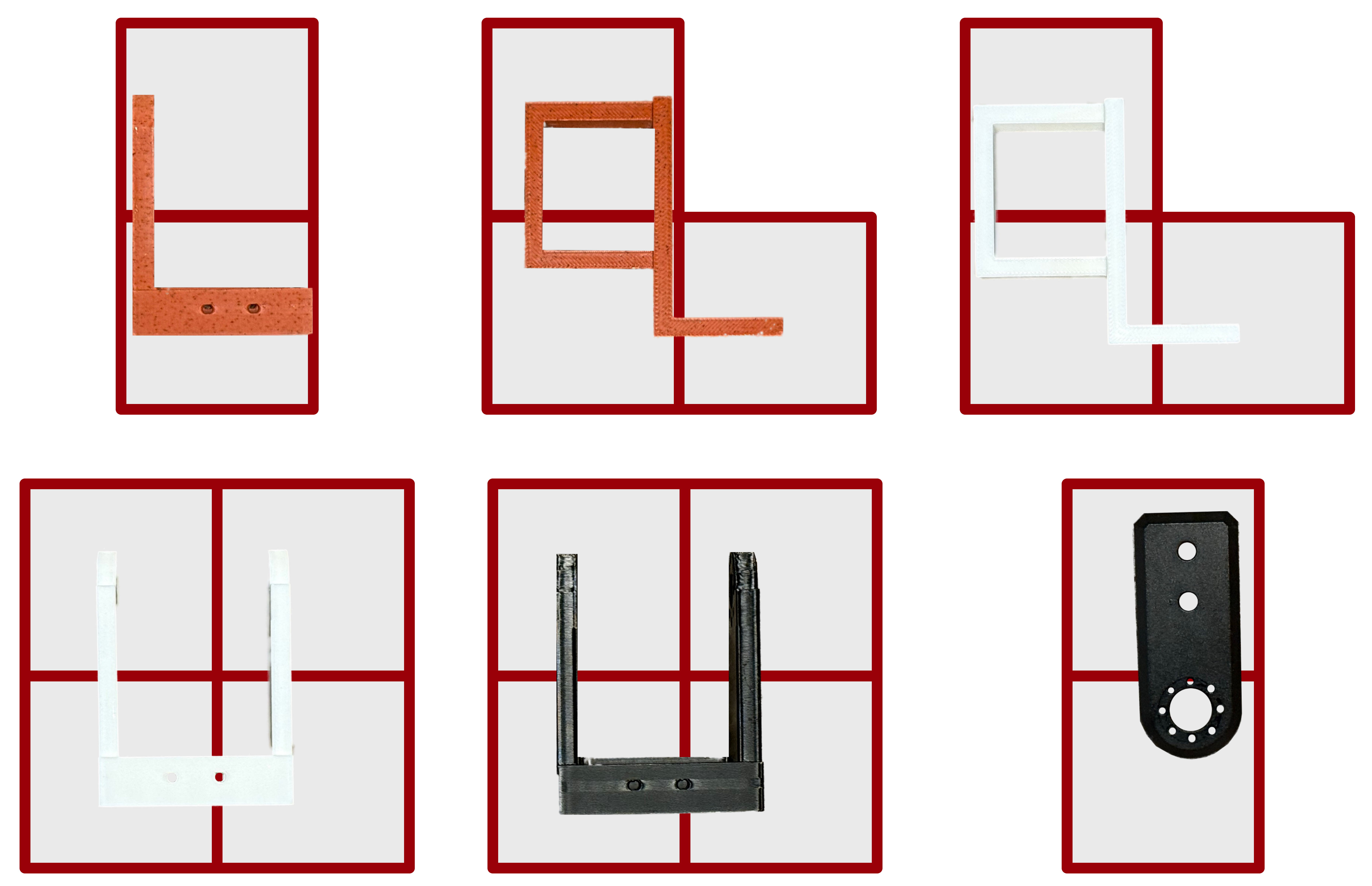}
    \caption{%
        \textbf{Parts Organization: Component shapes and grid occupancy.}
        Each panel shows a 3D-printed part overlaid on its $2\times 2$ template
        grid, illustrating which cells (\texttt{[a,b,c,d]}) each part occupies.
        \textbf{Top row}: \texttt{red-L} (2 cells, occupies \texttt{[a,c]}),
        \texttt{red-q} (3 cells, \texttt{[a,c,d]}),
        \texttt{white-q} with \texttt{red-L} interlocking example.
        \textbf{Bottom row}: \texttt{white-U} (4 cells, \texttt{[a,b,c,d]}),
        \texttt{black-U} (4 cells), \texttt{black-I} (2 cells, \texttt{[a,c]}).
        Red grid lines mark template boundaries. Note the internal geometric
        biases: U-shapes have empty space in \texttt{[a,c]} (the ``U'' opening),
        while q-shapes have opposite edge alignments (white-q left-aligned,
        red-q right-aligned).
    }
    \label{fig:parts-components}
\end{figure}

\figref{fig:parts-components} visualizes all six components with their grid cell
occupancy patterns, illustrating the template-based coordinate system in practice.

\subsubsection{Scoring Mechanism}
\label{app:parts-scoring}

\textbf{Step-Level Feedback.}
Each placement action receives immediate feedback:
\begin{compactitem}
    \item \textbf{+1}: Successful placement in a safe configuration
    \item \textbf{-1}: Collision occurs during placement, but the plan was theoretically
          valid (indicates execution error or unforeseen physical interaction)
    \item \textbf{-2}: Invalid or unreasonable plan (\eg, out-of-bounds grid number,
          selecting non-adjacent cells, impossible rotation angle)
\end{compactitem}

\textbf{Error Limits.}
To prevent excessive failures and ensure episode progression:
\begin{compactitem}
    \item \textbf{Per-part limit}: Maximum 3 failed attempts per part. After 3 failures,
          the system skips to the next part.
    \item \textbf{Cumulative limit}: Maximum 5 total errors per episode. Exceeding this
          terminates the episode early.
\end{compactitem}

\textbf{Episode-Level Score.}
Final episode success is measured by the total grid cell occupancy after all five
parts are placed. The normalized score is computed as:
\begin{equation}
    \text{Score} = 5 - 5 \times \text{clip}\left(\frac{s - s_{\min}}{s_{\max} - s_{\min}}, 0, 1\right)
\end{equation}
where $s$ is the actual grid occupancy, $s_{\min}$ is the theoretical minimum
occupancy (optimal packing), and $s_{\max}$ is the maximum possible occupancy.
A score of 5 indicates optimal packing, while 0 indicates worst-case packing.

\subsubsection{Safety Mechanism}
\label{app:parts-safety}

Before executing each placement, the system performs collision checking using
Segment Anything Model (SAM)~\citep{kirillov2023segment}:
\begin{compactenum}
    \item Generate segmentation mask for the part to be placed
    \item Generate masks for all currently placed parts
    \item Compute mask overlap between proposed placement and existing parts
    \item Reject placement if overlap exceeds threshold; return collision penalty
\end{compactenum}

This SAM-based collision detection prevents physical damage while providing
informative negative feedback for learning.

\subsubsection{Low-Level Executor}
\label{app:parts-executor}

Given a valid VLM plan (grid cell indices and optional rotation), the low-level
controller executes a heuristic pick-and-place sequence:
\begin{compactenum}
    \item \textbf{Grasp Pose}: Use a fixed pre-defined grasp pose for each part type,
          determined through offline calibration. This ensures consistent grasping
          regardless of VLM decisions.
    \item \textbf{Coordinate Transformation}: Convert target grid cell indices to
          3D world coordinates using camera-to-robot calibration matrices.
    \item \textbf{Placement Execution}: Execute pick-and-place motion with the
          gripper oriented according to the specified rotation angle.
    \item \textbf{Release and Retract}: Open gripper and retract to a safe
          observation pose.
\end{compactenum}

\subsubsection{Evidence Recording}
\label{app:parts-evidence}

For memory formation and learning, the system records keyframe evidence at
critical moments:
\begin{compactitem}
    \item \textbf{Step Start}: Capture visual state before any action is taken
          (current grid occupancy, part to be placed)
    \item \textbf{Post-Stabilization}: Capture visual state after the robot
          retracts and the scene stabilizes (placement outcome, new occupancy)
\end{compactitem}
These keyframes, along with the VLM's action and resulting score, form the
episodic memory entries used for principle extraction.

\newpage
\subsubsection{Memory Content Examples}
\label{app:parts-memory-examples}

We provide concrete examples of how the coordinate system enables structured
hypothesis formation and principle extraction.

\textbf{Episodic Memory (Experience Records).}
Each experience captures the context, action, outcome, and spatial reasoning:
\begin{promptlisting}
Experience #47:
- Part: white-U (4 cells, [a,b,c,d])
- Action: place(5, 6, 15, 16, rotation=180)
- Outcome: SUCCESS (+1)
- Context: black-U already at cells (3,4,13,14) with rotation=0
- Observation: white-U's [a,c] region (now bottom-left after 180 deg)
  interlocked with black-U's [b,d] region without collision

Experience #52:
- Part: red-q (3 cells, [a,c,d])
- Action: place(7, 17, 18)
- Outcome: COLLISION (-1)
- Context: white-q at cells (6, 16, 17) with rotation=0
- Observation: red-q's [a,c] region overlapped with white-q's [a,c]
  region; internal geometries conflicted despite same template cells
\end{promptlisting}

\textbf{Working Memory (Hypotheses Under Testing).}
Hypotheses are generated from clustered experiences, using template coordinates:
\begin{promptlisting}
\### Hypotheses (Under Testing)

1. [TESTING] (from 3 experiences) When placing U-shaped parts:
   - At 180 deg rotation, the [a,c] region becomes physically
     accessible for interlocking with another U's [b,d] region
   - Confidence: 67% (2/3 successes with this pattern)

2. [TESTING] (from 4 experiences) For q-shaped part pairs:
   - AVOID having both parts' [a,c] regions overlap on same cells
   - The [d] region of one q-shape CAN overlap with [c] of another
   - Confidence: 75% (3/4 successes following this rule)

3. [TESTING] (from 2 experiences) Black-I placement strategy:
   - The internal bias (top-right of [a,c]) allows black-I to fit
     into the empty [a,c] space of U-shaped parts
   - Requires U-shape at 0 deg (opening faces left)
\end{promptlisting}

\newpage
\textbf{Long-Term Memory (Verified Principles).}
Principles are promoted after sufficient verification (confidence $\geq 0.8$):
\begin{promptlisting}
\### Verified Principles

1. [92%] [PREFER] U-shape interlocking strategy:
   When two U-shapes must share adjacent cells, place the second
   at 180 deg rotation so its [a,c] region interlocks with the
   first U's [b,d] region. The complementary internal biases
   (physical mass in [b,d], empty space in [a,c]) enable overlap.

2. [88%] [AVOID] Q-shape [a,c] conflict:
   Never place two q-shaped parts with their [a,c] regions on
   overlapping grid cells. Despite occupying the same template
   positions, their internal geometries (left-aligned vs right-
   aligned) create physical collisions.

3. [85%] [PREFER] Q-shape complementary placement:
   Place white-q's [c] region adjacent to (or sharing cells with)
   red-q's [d] region. Their opposite internal biases create
   physical clearance for successful overlap.

4. [82%] [SEQUENCE] Size-constrained ordering:
   Place U-shaped parts (4 cells) before q-shaped parts (3 cells).
   U-shapes have less placement flexibility and benefit from
   early positioning when more grid space is available.
\end{promptlisting}

\textbf{Complete Specifications.}
The full coordinate system reference, spatial reasoning guidelines, and decision rules incorporated into the VLM prompt for \taskParts{} are elided in this release.


\clearpage
\subsection{\textbf{\taskBall}}
\label{app:task-ball}

The ball navigation task requires the robot to push a common soccer ball toy
through an obstacle course to reach a target location, testing the VLM's ability
to adapt to complex ball rolling dynamics and select appropriate push strategies.

\subsubsection{Task Description}
\label{app:ball-description}

\textbf{Setup.}
A soccer ball toy is placed on a tabletop workspace containing multiple obstacles.
The ball's starting position is randomly initialized within a fixed area and
manually placed. To facilitate smooth rolling interaction, the gripper holds a
cylindrical foam piece that contacts the ball during pushing.

\textbf{Task Stages.}
The ball must pass through at least three stages to reach the target:
\begin{compactenum}
    \item \textbf{Stage 1}: Pass through the first obstacle's ``bridge hole''
    \item \textbf{Stage 2}: Navigate around all obstacles and approach the target area
    \item \textbf{Stage 3}: Fine-tune position and reach the final target
\end{compactenum}
The goal is to complete all stages within a maximum of 6 push actions.

\textbf{Ball Dynamics.}
According to momentum principles and empirical testing, the ball's landing
position after each push depends on:
(1) gripper movement direction,
(2) gripper stopping point,
(3) gripper movement speed,
(4) the ball's intrinsic dynamics properties, and
(5) environmental factors.

\textbf{Evidence Recording.}
For each interaction step, we record keyframes from the top-down camera at
two time points: (1) the start of the step, and (2) after a fixed duration
when the ball has stabilized. These keyframes serve as \textbf{evidence} for
the experience stored in memory.

\textbf{Challenge.}
The task is challenging because:
(1) ball rolling dynamics are difficult to predict precisely,
(2) obstacle configurations require multi-step indirect pushing paths,
(3) different speed levels produce significantly different ball behaviors, and
(4) the VLM must learn ball-specific strategies through trial and error.

\subsubsection{VLM Interface}
\label{app:ball-vlm}

\textbf{Visual Input.}
The VLM receives two images per decision step:
\begin{compactenum}
    \item \textbf{Wrist Camera}: Close-up view of the ball and nearby obstacles,
          providing detailed local context for push planning
    \item \textbf{Top-Down Camera}: Bird's-eye view of the entire workspace showing:
          \begin{compactitem}
              \item Current ball position
              \item All obstacle locations and shapes
              \item Target region location
              \item Trajectory history from previous pushes (if any)
          \end{compactitem}
\end{compactenum}

\textbf{Action Output.}
The VLM outputs a push action combining visual grounding with text specification:
\begin{compactitem}
    \item \texttt{start\_direction}: Direction from ball center to gripper start position,
          from 8 options: \texttt{left}, \texttt{right}, \texttt{top}, \texttt{bottom},
          \texttt{top-left}, \texttt{top-right}, \texttt{bottom-left}, \texttt{bottom-right}
    \item \texttt{end\_direction}: Direction from ball center to gripper end position
          (same 8 options)
    \item \texttt{start\_y, start\_x}: Pixel coordinates where gripper starts the push
          (note: $y$ before $x$ following Gemini visual grounding convention)
    \item \texttt{end\_y, end\_x}: Pixel coordinates defining push endpoint
    \item \texttt{speed}: Push velocity level $\in \{\text{low}, \text{medium}, \text{high}\}$
\end{compactitem}

\textbf{Coordinate System.}
Pixel values are scaled to a normalized 0--1000 range regardless of original
image dimensions. The top-left corner is $(0, 0)$ and the bottom-right is
$(1000, 1000)$. These normalized coordinates are converted to absolute pixel
positions based on the camera frame's actual width and height.

\textbf{Direction Specification.}
The \texttt{start\_direction} and \texttt{end\_direction} fields explicitly
encode the positional relationship between the gripper points and the ball's
current position. This forces the VLM to reason about push geometry and
enhances its understanding of the push strategy, although these direction
labels are not directly passed to the low-level executor.

\textbf{Endpoint Constraint.}
To ensure physically reasonable push distances, the \textbf{end point must lie
within a circle of radius 150mm centered at the start point}. The low-level
executor clips any endpoint that exceeds this constraint.

Example: \texttt{push("left", "right", 300, 150, 300, 350, "medium")} specifies
a push where the gripper starts on the left side of the ball and pushes toward
the right at medium speed.

\subsubsection{Scoring Mechanism}
\label{app:ball-scoring}

The scoring system provides dense feedback for learning:
\begin{itemize}
    \item \textbf{$+1$}: Gripper successfully affects ball state (ball moves)
    \item \textbf{$-2$}: Gripper collides with any obstacle during interaction
    \item \textbf{$-2$}: VLM outputs invalid start/end point (in unsafe zone)
    \item \textbf{$+3$}: Ball successfully passes through obstacle 1's bridge hole
    \item \textbf{$+5$}: Ball reaches the final target at step $i$ ($i \leq 6$)
    \item \textbf{$+2 \times (6-i)$}: Early completion bonus for reaching target at step $i$
\end{itemize}

\textbf{Episode Termination.}
The episode terminates when:
(1) the ball reaches the target (success),
(2) the VLM outputs unsafe points more than 3 times consecutively,
(3) 6 steps are exhausted without reaching target (failure), or
(4) the ball exits the workspace boundaries (failure).
Upon termination, the ball is manually reset to a random starting position.

\subsubsection{Safety Mechanism}
\label{app:ball-safety}

To prevent gripper-obstacle collisions during push execution:
\begin{itemize}
    \item Segment all obstacles on the tabletop using the top-down camera
    \item Generate obstacle masks via SAM segmentation
    \item Apply morphological dilation to expand obstacle regions by a safety
          margin (similar to costmap inflation in ROS navigation stack)
    \item Create an ``inflated obstacle map'' representing unsafe zones
    \item Check if the planned start point or end point falls within unsafe zones
    \item Reject plans where either point enters the inflated obstacle region
\end{itemize}

This inflation-based approach provides conservative collision avoidance while
allowing the ball itself to pass close to obstacles. If the VLM outputs unsafe
points more than 3 consecutive times, the episode is terminated.

\subsubsection{Low-Level Executor}
\label{app:ball-executor}

Given a valid VLM push specification, the low-level controller implements a
heuristic push policy:
\begin{itemize}
    \item Convert normalized pixel coordinates (0--1000) to absolute pixels
          based on camera frame dimensions
    \item Convert 2D pixel coordinates to 3D world coordinates using camera
          calibration
    \item Move gripper to the start point position
    \item Rotate gripper to align with the push direction (start $\to$ end vector)
    \item Execute constant-velocity linear motion toward the end point:
          \begin{compactitem}
              \item Low: 150 mm/s
              \item Medium: 300 mm/s
              \item High: 450 mm/s
          \end{compactitem}
    \item \textbf{Vertical offset}: The end point is offset slightly upward along
          the z-axis compared to the start point, simulating the human behavior
          of pushing while gradually lifting. This facilitates ball rolling and
          adds complexity to the motion dynamics.
    \item \textbf{Endpoint clipping}: Final end point is clipped to ensure it
          remains within the 150mm radius circle centered at the start point
    \item Retract gripper after push completion
\end{itemize}

\subsubsection{Obstacle Layout and Spatial Zones}
\label{app:ball-obstacles}

The workspace contains three fixed obstacles and a target region:

\textbf{Obstacle Configuration.}
\begin{itemize}
    \item \textbf{Obstacle 1 (Blue)}: Archway block on the LEFT side of the workspace.
          Has a ``bridge hole'' at approximately $y \in [400, 480]$ that the ball
          \emph{must} pass through. The ball cannot go over or around this obstacle.
    \item \textbf{Obstacle 2 (Red)}: Rectangular cuboid in the MIDDLE of the workspace.
          Acts as a solid blocker that the ball must navigate around.
    \item \textbf{Obstacle 3 (Purple)}: Rectangular cuboid at the BOTTOM of the workspace.
          \textbf{CRITICAL}: If the ball lands on top of this obstacle ($y \approx 600$--$700$),
          the gripper cannot access the ball from below, resulting in a stuck state.
    \item \textbf{Target}: Circular region with cross pattern on the RIGHT side.
\end{itemize}

\textbf{Spatial Zones.}
For reasoning about ball position and strategy selection, we define five zones:
\begin{itemize}
    \item \textbf{Zone A} (Initial): $y < 400$, $x > 300$ --- Above archway level
    \item \textbf{Zone B} (Pre-archway): $y \in [400, 480]$, $x < 350$ --- Aligned with archway
    \item \textbf{Zone C} (Post-archway): $x > 350$, $y < 600$ --- Transit area after archway
    \item \textbf{Zone D} (Target approach): $x > 650$, $y < 550$ --- Near target
    \item \textbf{Danger Zone}: $y > 550$, $x > 520$ --- Robot arm motion limit exceeded
\end{itemize}

\newpage
\subsubsection{Memory Content Examples}
\label{app:ball-memory-examples}

We provide concrete examples of how spatial zones and obstacle references
enable structured hypothesis formation and principle extraction.

\textbf{Episodic Memory (Experience Records).}
Each experience captures ball position, zone, distances to obstacles, and outcome:
\begin{promptlisting}
Experience #23:
- Step: 2/6, Ball Zone: ZONE B (pre-archway, aligned with archway)
- Ball Position: (y=440, x=280)
- Distance to OBS1 archway: ~50 pixels
- Distance to OBS3: ~280 pixels
- Action: push("left", "right", 440, 200, 440, 450, "medium")
- Outcome: SUCCESS (+3, passed through archway)
- Ball Position (after): (y=430, x=480)
- Observation: MEDIUM speed successfully pushed ball through archway.
  Horizontal push (constant y) maintained archway height alignment.

Experience #15:
- Step: 1/6, Ball Zone: ZONE A (initial, close to OBS1)
- Ball Position: (y=350, x=420)
- Distance to OBS1 archway: ~150 pixels
- Action: push("top-right", "bottom-left", 280, 480, 450, 300, "high")
- Outcome: OVERSHOOT (-1)
- Ball Position (after): (y=680, x=250)
- Observation: HIGH speed too strong from close distance. Ball passed
  through archway but landed ON TOP of OBS3 (y~680). Now stuck.
\end{promptlisting}

\textbf{Working Memory (Hypotheses Under Testing).}
Hypotheses reference obstacles and zones explicitly:
\begin{promptlisting}
\### Hypotheses (Under Testing)

1. [TESTING] Post-archway LOW speed requirement (from 3 experiences):
   After passing through OBS1, MUST use LOW speed for next push.
   MEDIUM/HIGH causes ball to roll onto OBS3 top surface (y~600-700).
   Confidence: 67% (2/3)

2. [TESTING] 45-degree descent for archway alignment (from 4 experiences):
   When in ZONE A (above archway level), diagonal descent
   (end_dir="bottom-left") is more controllable than straight "bottom".
   Confidence: 75% (3/4)
\end{promptlisting}
\newpage
\textbf{Long-Term Memory (Verified Principles).}
Principles are promoted after sufficient verification:
\begin{promptlisting}
\### Verified Principles

1. [95%] [SPEED] Distance-based speed for OBS1 approach:
   Ball >200px from archway: HIGH speed acceptable
   Ball 100-200px: MEDIUM preferred; Ball <100px: LOW REQUIRED

2. [88%] [AVOID] OBS3 danger zone (y>550, x>520):
   Never push toward target from DANGER zone. Robot arm exceeds
   motion limit. MUST push upward (end_dir includes "top") first.
\end{promptlisting}

\textbf{Complete Specifications.}
The full obstacle layout, spatial zone definitions, and stage-specific decision rules incorporated into the VLM prompt for \taskBall{} are elided in this release.

\clearpage
\subsection{\textbf{\taskStack}}
\label{app:task-stack}

The balanced stacking task requires the robot to build a stable tower from
irregularly-shaped ``balance stones'' with varying physical properties, testing
physics reasoning about stability, friction, and center of gravity.

\subsubsection{Task Description}
\label{app:stack-description}

\textbf{Setup.}
Five balance stones with diverse properties are scattered on the workspace:
\begin{compactitem}
    \item \textbf{Size variation}: Small, medium, and large stones
    \item \textbf{Shape variation}: Flat, rounded, angular geometries
    \item \textbf{Surface properties}: Different friction coefficients
          (smooth, textured, rough)
    \item \textbf{Weight distribution}: Varying centers of gravity
\end{compactitem}

\textbf{Objective.}
Stack all five stones into a stable tower. The tower must remain standing
(without collapse) for a stability verification period after the final
stone is placed.

\textbf{Challenge.}
The task is challenging because:
(1) optimal stacking order depends on stone properties that are not visually obvious,
(2) stone shapes create complex contact geometries affecting stability,
(3) friction between stone pairs varies and affects stackability, and
(4) the VLM must learn which stone combinations are stable through trial and error.

\subsubsection{VLM Interface}
\label{app:stack-vlm}

\textbf{Visual Input.}
The VLM receives three images per decision step:
\begin{compactenum}
    \item \textbf{Wrist Camera}: Close-up view of the stone being grasped or
          the current tower top surface, showing detailed texture and contact geometry
    \item \textbf{Top-Down Camera}: Overhead view showing all available stones
          and the current tower state
    \item \textbf{Base Camera}: Side-angle view of the stacking area providing
          depth perception for tower height and stone orientations
\end{compactenum}

\textbf{Action Output.}
The VLM outputs a visual grounding specification:
\begin{compactitem}
    \item \texttt{stack(point\_y, point\_x)}: Pixel coordinates identifying which
          stone to pick up next (note: $y$ before $x$ following Gemini convention)
\end{compactitem}

Example: \texttt{stack(200, 150)} selects the stone at pixel position $y=200$, $x=150$.
The system uses this point to identify the target stone via SAM segmentation,
then autonomously determines grasp pose and placement location using adaptive
height control.

\subsubsection{Scoring Mechanism}
\label{app:stack-scoring}

The scoring system rewards successful stacking while penalizing collapses:
\begin{compactitem}
    \item \textbf{$+i$}: Successfully placing a stone on layer $i$ (where $i=1$
          is the base, $i=2$ is second layer, etc.)
    \item \textbf{$-2j$}: Penalty when $j$ stones fall during or after placement
\end{compactitem}

This progressive reward structure encourages building taller stable towers:
placing the 5th stone on a 4-layer tower yields $+5$, but causing 3 stones
to fall incurs $-6$. The optimal strategy balances ambition with stability.

Episode success is defined as completing a 5-stone tower that remains stable.

\subsubsection{Safety Mechanism}
\label{app:stack-safety}

Tower stability is monitored continuously:
\begin{compactenum}
    \item Visual tracking of stone positions during and after placement
    \item Automatic episode termination if catastrophic collapse detected
    \item Force/torque sensing during placement to detect instability
\end{compactenum}

\subsubsection{Low-Level Executor}
\label{app:stack-executor}

Given the VLM's target stone selection, the low-level controller performs:

\textbf{Grasp Planning.}
\begin{compactitem}
    \item Apply SAM segmentation to isolate the selected stone
    \item Analyze stone geometry to determine optimal grasp direction
    \item Compute antipodal grasp points based on stone shape
    \item Plan collision-free approach trajectory
\end{compactitem}

\textbf{Adaptive Placement.}
\begin{compactitem}
    \item Move stone above current tower with clearance margin
    \item \textbf{Adaptive height adjustment}: Incrementally lower stone while
          monitoring force feedback
    \item Detect contact with tower surface via force threshold
    \item Release gripper and retract with minimal disturbance
    \item Verify tower stability before proceeding
\end{compactitem}

The adaptive height adjustment is critical for handling the irregular stone
geometries and varying tower heights.

\subsubsection{Stone Naming System and Surface Properties}
\label{app:stack-stones}

To enable precise hypothesis formation about stone interactions, we use a
systematic naming convention: \texttt{\{Surface\}\_\{Shape\}\_\{Size\}\_\{Orientation\}}.

\textbf{Surface Types (Friction Ranking).}
Five surface materials with different friction coefficients:
\begin{compactenum}
    \item \textbf{3M} (3M Gripping Material TB641): Black velcro-like, \emph{highest} friction
    \item \textbf{BLK} (Black Duct Tape): Industrial grade, high friction
    \item \textbf{WHT} (White PVC Tape): Insulation tape, medium friction
    \item \textbf{WOOD} (Rough Wooden): Natural wood grain, medium-low friction
    \item \textbf{PAINT} (Painted Surface): Smooth lacquer, \emph{lowest} friction
\end{compactenum}

\textbf{Shape Types (Cross-Section Geometry).}
\begin{compactenum}
    \item \textbf{SQR} (Square): Flat 4-sided, excellent contact area
    \item \textbf{HEX} (Hexagonal): 6-sided, smaller contact, best for top layers
    \item \textbf{DMND} (Diamond): Can be horizontal (H) or vertical (V) bar
    \item \textbf{PENT} (Pentagon): 5-sided irregular
    \item \textbf{OVAL} (Egg): Rounded, challenging contact
    \item \textbf{TREE} (Branch): Irregular, must always be placed last
\end{compactenum}

\textbf{Orientation Types.}
\begin{compactitem}
    \item \textbf{H} (Horizontal): Long bar lying flat, width $\gg$ height
    \item \textbf{V} (Vertical): Long bar standing up, height $\gg$ width
    \item \textbf{C} (Compact): Roughly cubic, width $\approx$ height
\end{compactitem}

Example: \texttt{3M\_HEX\_L\_C} = 3M surface + Hexagonal cross-section + Large + Compact

\subsubsection{Memory Content Examples}
\label{app:stack-memory-examples}

We provide concrete examples of how surface and shape properties enable
structured hypothesis formation and principle extraction.
\newpage
\textbf{Episodic Memory (Experience Records).}
Each experience captures stone properties, tower state, and placement outcome:
\begin{promptlisting}
Experience #18:
- Step: 1/5, Tower State: Empty (first stone)
- Stone Selected: BLK_DMND_M_V (Black Diamond Medium Vertical)
- Contact Cross-Section: Diamond point, ~3cm2
- Aspect Ratio: 1:3.5 (vertical bar)
- Surface Friction: HIGH (Black Tape)
- Action: stack(280, 380)
- Outcome: COLLAPSE (-2), Stone fell immediately
- Observation: Vertical bar (V orientation) as base stone FAILS.
  Small contact area insufficient. AVOID V-orientation at Step 1.

Experience #31:
- Step: 3/5, Tower State: 2 layers
- Current Top: 3M_SQR_M_C (3M Square)
- Stone Selected: WOOD_SQR_M_C (Wooden Square)
- Contact Compatibility: SQR -> SQR (excellent)
- Friction Pairing: 3M -> WOOD (good)
- Action: stack(400, 380)
- Outcome: SUCCESS (+3)
- Observation: SQR on SQR with 3M->WOOD friction is stable.
\end{promptlisting}

\textbf{Working Memory (Hypotheses Under Testing).}
Hypotheses reference surface and shape properties:
\begin{promptlisting}
\### Hypotheses (Under Testing)

1. [TESTING] Tree-shaped (TREE) stones must be LAST (Step 5):
   Irregular surface cannot support any stone above.
   Confidence: 100% (3/3)

2. [TESTING] SQR on DMND_H incompatibility:
   Square cross-section cannot stack on horizontal bar.
   Contact geometries do not align. 100% collapse rate.
   Confidence: 100% (3/3)
\end{promptlisting}

\textbf{Long-Term Memory (Verified Principles).}
Principles use the naming convention for precise specification:
\begin{promptlisting}
\### Verified Principles

1. [95%] [REQUIRE] Step 1 base stone criteria:
   Select LARGEST stone + HIGHEST friction (3M>BLK>WHT>WOOD>PAINT)
   + LARGEST contact area. NEVER select V-orientation.

2. [90%] [SEQUENCE] HEX stones at Steps 4-5:
   Hexagonal contact too small for lower layers.
   Place after Step 3 (toward top of tower).
\end{promptlisting}

\textbf{Complete Specifications.}
The full stone naming convention, friction compatibility matrix, and step-by-step selection rules incorporated into the VLM prompt for \taskStack{} are elided in this release.

\clearpage
\subsection{Example Learned Principles}
\label{app:principles-examples}

This section presents example principles learned by \method during real-world
experiments. These demonstrate the system's ability to discover task-specific
physics knowledge through interaction.

\textbf{\taskParts.}
\begin{itemize}
    \item \prinPrefer: ``Place elongated parts along grid edges to maximize
          interior space for irregular shapes''
    \item \prinAvoid: ``Don't place large multi-cell parts in the center first;
          they block flexible arrangements''
    \item \prinSequence: ``Place the most constrained (largest) parts first,
          then fill gaps with smaller parts''
    \item \prinAvoid: ``Avoid complete overlap between the [a,c] regions of two
          q-shaped parts; their internal geometries conflict''
    \item \prinPrefer: ``When placing white-U at 180 degrees, its [a,c] region
          can interlock with another U-shape's [b,d] region''
    \item \prinSequence: ``Place I-shaped parts last; their narrow profile fits
          into remaining gaps more easily''
\end{itemize}

\textbf{\taskBall.}
\begin{itemize}
    \item \prinAvoid: ``High speed pushes near obstacles; ball rebounds unpredictably''
    \item \prinPrefer: ``Use medium speed for long-distance pushes toward open areas''
    \item \prinPrefer: ``Aim for indirect paths that use obstacle edges to guide ball''
    \item \prinAvoid: ``Don't push directly toward the bridge hole from far away;
          small angular errors cause the ball to miss''
    \item \prinSequence: ``First navigate past the archway, then reduce speed
          for the final approach to target''
    \item \prinPrefer: ``Use low speed (0.3--0.5) after passing through narrow
          gaps to maintain control''
    \item \prinAvoid: ``Avoid pushing at angles greater than 45 degrees; the ball
          tends to spin rather than roll straight''
\end{itemize}

\textbf{\taskStack.}
\begin{itemize}
    \item \prinSequence: ``Place the flattest, largest stone as base for maximum
          contact area''
    \item \prinAvoid: ``Don't stack smooth-surfaced stones directly on each other;
          they slide easily''
    \item \prinPrefer: ``Alternate between rough and smooth stones for better grip''
    \item \prinAvoid: ``Don't place heavy stones on small contact surfaces;
          high pressure causes instability''
    \item \prinPrefer: ``Select stones with natural concavities as middle layers;
          they cradle the stones above''
    \item \prinAvoid: ``Don't place the largest stone on top; the high center
          of mass causes toppling''
    \item \prinSequence: ``Build towers with decreasing stone size from bottom
          to top for optimal stability''
\end{itemize}

\newpage
\subsection{Out-of-Distribution Experimental Settings}
\label{app:ood-settings}

\begin{figure*}[t]
    \centering
    \includegraphics[width=\textwidth]{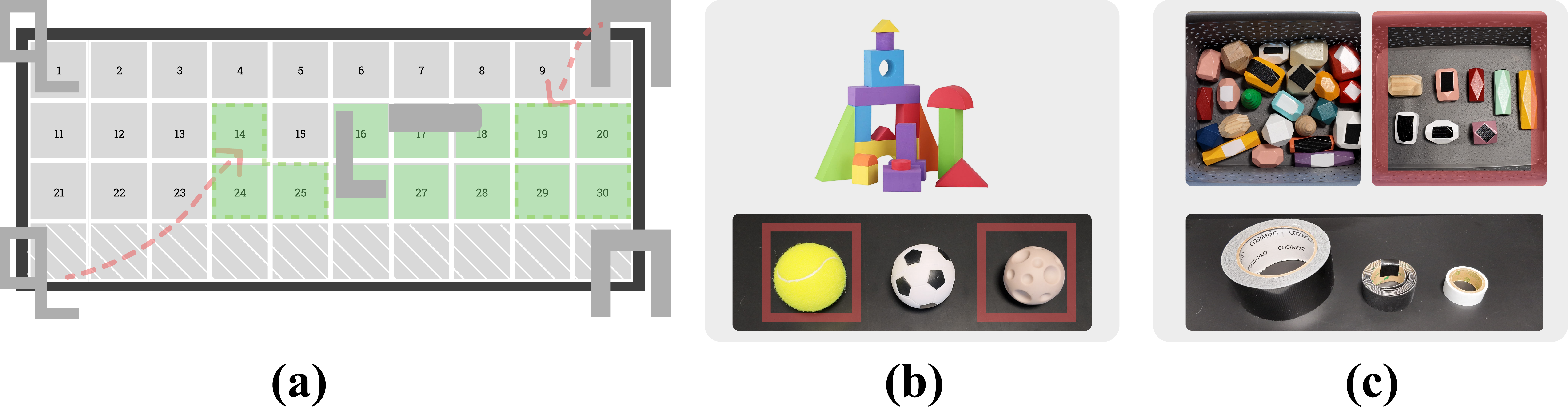}
    \caption{%
        \textbf{Out-of-distribution experimental settings and props.}
        (a)~\textbf{\taskParts~OOD}: Initial grid configuration is modified with
        parts pre-placed at different positions. Green cells (14--16, 23--25) are
        pre-occupied; hatched area indicates the constrained placement region
        ($\pm 3$ columns from occupied boundary). Dashed trajectory shows example
        placement paths.
        (b)~\textbf{\taskBall~OOD}: Top shows stacking blocks (illustrative only,
        not used in ball task). Bottom shows all balls: soccer ball (center) is
        the in-distribution training ball; tennis ball (yellow) and moon-textured
        plastic ball (red boxes) are OOD items with different friction and elasticity.
        (c)~\textbf{\taskStack~OOD}: Top rows show balance stones with varying
        surfaces---some treated with tapes (3M, black duct, white PVC; shown in
        bottom), others with natural WOOD or PAINT surfaces. Red-boxed stones
        (top-right) represent OOD combinations with novel shape/material/size
        configurations not seen during in-distribution training.
    }
    \label{fig:ood-props}
\end{figure*}

To evaluate memory transfer capabilities (\secref{sec:exp:transfer} in main text),
we designed out-of-distribution (OOD) variants for each task. \figref{fig:ood-props}
shows the OOD settings and experimental props.

\textbf{\taskParts~OOD Variant.}
We modify the initial grid configuration:
\begin{itemize}
    \item Parts are pre-placed at different positions (occupying cells 14--16, 23--25)
    \item Initial component count is increased, reducing available placement space
    \item The constrained placement region ($\pm 3$ columns from occupied boundary)
          creates more challenging packing scenarios
\end{itemize}
These modifications test whether learned spatial reasoning principles transfer
to novel initial configurations.

\textbf{\taskBall~OOD Variant.}
We introduce two new ball types with substantially different dynamics:
\begin{itemize}
    \item \textsc{Tennis ball}: Higher friction, less predictable bounce behavior
    \item \textsc{Moon-textured plastic ball}: Lower friction, different rolling dynamics
\end{itemize}
Prior principles about soccer ball dynamics (speed selection, push angle effects)
may not transfer directly, requiring test-time adaptation.

\textbf{\taskStack~OOD Variant.}
We test with five unseen stone combinations:
\begin{itemize}
    \item Novel weight distributions not encountered during training
    \item Surface textures created with different tape combinations
    \item Shape/material/size combinations that never appeared before
\end{itemize}
To ensure the fairness of the experiment, we used the same sequence of part combinations during the out-of-distribution (OOD) testing in different episodes. Stability principles (e.g., ``flat base first'') may transfer, but specific
friction pairing rules require re-learning. 


\clearpage
\section{Cross-Model Generalization Details}
\label{app:cross-model}

We tested \method across multiple VLM architectures to validate model-agnostic
benefits.

\begin{table}[h]
    \centering
    \caption{\textbf{Cross-model results on medium difficulty.} Success rates
    with and without \method across different VLM planners (100 episodes each).
    All models use thinking mode. Consolidation model fixed to Qwen3-VL.}
    \label{tab:cross-model-full}
    \begin{tabular}{lccc}
        \toprule
        Model & Baseline & +\method & $\Delta$ \\
        \midrule
        Gemini-3-Flash & 53\% & \textbf{76\%} & +23\% \\
        GPT-5.1 & 43\% & 57\% & +14\% \\
        Qwen3-VL-235B & 38\% & 50\% & +12\% \\
        Gemini-ER-1.5 & 29\% & 34\% & +5\% \\
        \bottomrule
    \end{tabular}
\end{table}

\textbf{Analysis.}
Test-time learning benefits scale with VLM capability. Gemini-3-Flash achieves
the largest improvement (+23\%), while weaker models show diminishing returns.
This suggests that effective hypothesis generation and verification require
sufficient base reasoning capability. The consolidation model (Qwen3-VL) remains
fixed across all conditions, isolating the effect of planner capability on
learning efficiency.


\onecolumn
\section{Detailed Task Specifications for VLM Planning}
\label{app:task-specs}
This section originally contained the complete task-specific reference materials directly incorporated into VLM planner prompts (coordinate systems, naming conventions, spatial reasoning frameworks, and decision rules) for \taskParts, \taskBall, and \taskStack. We have elided these task-specific specifications in this release. The general structure of how such specifications integrate with the memory system is described in the eight-section prompt outline below (\secref{app:prompts}); examples of the principles and hypotheses that emerge for each task are reported in \secref{app:parts-memory-examples}, \secref{app:ball-memory-examples}, and \secref{app:stack-memory-examples}.



\clearpage

\section{Prompt Templates: Structure and General Components}
\label{app:prompts}

This section describes the structure of the prompt templates used in our real-world experiments and shows the task-agnostic components (hypothesis generation and memory injection). Each task prompt follows an \textbf{eight-section structure} that integrates the memory system:

\begin{compactenum}
    \item \textbf{Task Description}: Core task specification and naming conventions
    \item \textbf{Action Definitions}: Available actions and their semantics
    \item \textbf{Strategic Guidance}: Planning heuristics and decision rules
    \item \textbf{Current State}: Robot's current situation
    \item \textbf{Repeated Action Warning}: Loop detection (if applicable)
    \item \textbf{Long-term Memory}: Verified principles (Tier 3)
    \item \textbf{Working Memory}: Hypotheses under test (Tier 2)
    \item \textbf{Action Request}: Output format specification
\end{compactenum}

Sections 6--7 are dynamically populated by the memory system at run time. The complete per-task prompt instances for \taskBrick, \taskParts, \taskBall, and \taskStack are elided in this release; we instead show the task-agnostic hypothesis generation template and the memory injection format used across all tasks.



\clearpage
\subsection{Hypothesis Generation Prompt Template}
\label{app:prompt-hypothesis}

The consolidation engine uses the following \textbf{general template} to generate
hypotheses from clustered experiences. This template is task-agnostic and can be
instantiated for any manipulation task by providing task-specific descriptions
and examples.

\begin{promptlisting}[title={\textbf{General template for hypothesis generation (task-agnostic).}}]
============================================================
HYPOTHESIS GENERATION FROM EXPERIENCE CLUSTERS
============================================================

You are analyzing robot manipulation experiences to identify
patterns and generate actionable hypotheses for improving
task performance.

============================================================
TASK CONTEXT
============================================================

Task Name: {task_name}
Task Description: {task_description}

Available Actions: {available_actions}
  (e.g., "pick, insert, reorient, put down" for assembly;
         "push" for ball navigation;
         "stack" for balanced stacking;
         "place" for parts organization)

Key Object Properties: {key_properties}
  (e.g., "brick color, shape, orientation" for assembly;
         "ball position, obstacle locations, speed" for navigation;
         "stone size, texture, weight distribution" for stacking;
         "part shape, cell occupancy, rotation" for organization)

============================================================
CLUSTER SUMMARY
============================================================

This cluster contains {n_experiences} similar experiences:

Action Type Distribution:
{action_type_distribution}

Outcome Statistics:
- Success rate: {success_rate}%
- Total successes: {n_successes}
- Total failures: {n_failures}
- Common failure reasons: {failure_tags}

Shared Context/Properties:
{shared_properties}

============================================================
SAMPLE EXPERIENCES FROM CLUSTER
============================================================

{formatted_experiences}

[Note: The system automatically formats 3-5 representative
experiences from the cluster, showing action, outcome, and
relevant context for each.]

============================================================
EXISTING KNOWLEDGE (avoid duplicating)
============================================================

Current Verified Principles (Long-term Memory):
{existing_principles}

Current Hypotheses Under Testing (Working Memory):
{existing_hypotheses}

IMPORTANT: Do NOT generate hypotheses that duplicate or
closely overlap with existing knowledge.

============================================================
YOUR TASK: GENERATE HYPOTHESES
============================================================

Analyze the patterns in this experience cluster and generate
1-3 NEW hypotheses that could explain the observed outcomes.

HYPOTHESIS TYPES:
  - AVOID: Identifies actions/conditions leading to failure
    "Don't do X when condition Y is true"
  - PREFER: Identifies actions/strategies leading to success
    "Do X when condition Y is true"
  - SEQUENCE: Identifies order-dependent rules
    "Do X before Y" or "After X, do Y"
  - COMPARE: Identifies comparative preferences
    "X works better than Y when condition Z"
  - GENERAL: General observations about the task domain

REQUIREMENTS FOR EACH HYPOTHESIS:
  1. Be SPECIFIC - include concrete conditions and actions
  2. Be ACTIONABLE - the agent can directly apply the rule
  3. Be GROUNDED - based on patterns in the experience cluster
  4. Reference relevant properties (objects, conditions, states)

============================================================
OUTPUT FORMAT
============================================================

For each hypothesis, provide:

HYPOTHESIS {N}:
  Type: {AVOID | PREFER | SEQUENCE | COMPARE | GENERAL}
  Statement: [Clear, actionable rule in natural language]
  Applicable_Actions: [List of action types this applies to]
  Trigger_Conditions: [When/where this hypothesis applies]
  Evidence: [Brief explanation of supporting experiences]

============================================================
EXAMPLE OUTPUTS BY TASK TYPE
============================================================

[Parts Organization Task]
HYPOTHESIS 1:
  Type: PREFER
  Statement: Place L-shaped parts in corner positions with the
    corner facing inward to maximize space utilization.
  Applicable_Actions: [place]
  Trigger_Conditions: When placing L-shaped parts and corner
    cells (1, 10, 21, 30) are available.
  Evidence: 4/5 successful placements used corner positions.

[Ball Navigation Task]
HYPOTHESIS 1:
  Type: AVOID
  Statement: Do not use high speed when the ball is within
    50 pixels of any obstacle.
  Applicable_Actions: [push]
  Trigger_Conditions: When ball-to-obstacle distance < 50 pixels.
  Evidence: 6/7 failures with high speed occurred near obstacles.

[Balanced Stacking Task]
HYPOTHESIS 1:
  Type: SEQUENCE
  Statement: Place the largest, flattest stone as the base
    before stacking any other stones.
  Applicable_Actions: [stack]
  Trigger_Conditions: When tower is empty (first placement).
  Evidence: All 5 successful towers started with large flat base.

[Brick Assembly Task]
HYPOTHESIS 1:
  Type: AVOID
  Statement: Do not attempt insert when adjacent slots are
    occupied by blocking bricks.
  Applicable_Actions: [insert]
  Trigger_Conditions: When dependencies are not satisfied.
  Evidence: 8/8 blocked inserts had adjacent occupied slots.
\end{promptlisting}

\clearpage

\subsection{Memory Injection Formatting}
\label{app:prompt-memory-format}

\textbf{Principle Formatting (Long-term Memory).}
Principles are formatted with confidence levels using a three-tier system:

\begin{promptlisting}
\#\# Learned Principles (Apply these!)

1. [95%] [SEQUENCE] Always place the largest stone as the base.
2. [88%] [AVOID] High-speed pushes near obstacles cause rebounds.
3. [82%] [PREFER] Place L-shaped parts in corners facing inward.
\end{promptlisting}

Confidence display thresholds:
\begin{compactitem}
    \item \textbf{HIGH} ($\geq 85\%$): Strong evidence, should be followed
    \item \textbf{MEDIUM} ($60\%$--$84\%$): Moderate evidence, consider carefully
    \item \textbf{LOW} ($< 60\%$): Weak evidence, use with caution
\end{compactitem}

\textbf{Hypothesis Formatting (Working Memory).}
Hypotheses are marked as under testing to indicate lower confidence:

\begin{promptlisting}
\#\# Hypotheses (Consider but verify)

1. [TESTING] Rough stones grip smooth stones better than vice versa.
2. [TESTING] Medium speed is optimal for most ball navigation.
\end{promptlisting}


\end{document}